\documentclass{article}

\PassOptionsToPackage{sectionbib}{natbib}
\usepackage{chapterbib}

\usepackage[final]{neurips_2023}

\usepackage[utf8]{inputenc} 
\usepackage[T1]{fontenc}    
\usepackage{hyperref}       
\usepackage{url}            
\usepackage{booktabs}       
\usepackage{amsfonts}       
\usepackage{nicefrac}       
\usepackage{microtype}      
\usepackage{xcolor}         

\usepackage{graphicx}
\usepackage{epstopdf}
\graphicspath{{figures/}}
\usepackage{amsmath}
\usepackage{amssymb}
\usepackage{subcaption}
\usepackage{wrapfig}
\usepackage{algorithm}

\usepackage[frozencache,cachedir=minted-cache]{minted}

\title{Implicit Contrastive Representation Learning with Guided Stop-gradient}

\author{
  Byeongchan Lee\thanks{These authors contributed equally to this work.}\\
  Gauss Labs\\
  Seoul, Korea \\
  \texttt{byeongchan.lee@gausslabs.ai} \\
  \And
  Sehyun Lee$^*$\\
  KAIST \\
  Daejeon, Korea \\
  \texttt{sehyun.lee@kaist.ac.kr} \\
}

\begin{document}

\maketitle

\begin{abstract}
  In self-supervised representation learning, Siamese networks are a natural architecture for learning transformation-invariance by bringing representations of positive pairs closer together. But it is prone to collapse into a degenerate solution. To address the issue, in contrastive learning, a contrastive loss is used to prevent collapse by moving representations of negative pairs away from each other. But it is known that algorithms with negative sampling are not robust to a reduction in the number of negative samples. So, on the other hand, there are algorithms that do not use negative pairs. Many positive-only algorithms adopt asymmetric network architecture consisting of source and target encoders as a key factor in coping with collapse. By exploiting the asymmetric architecture, we introduce a methodology to implicitly incorporate the idea of contrastive learning. As its implementation, we present a novel method guided stop-gradient. We apply our method to benchmark algorithms SimSiam and BYOL and show that our method stabilizes training and boosts performance. We also show that the algorithms with our method work well with small batch sizes and do not collapse even when there is no predictor. The code is available at \url{https://github.com/bych-lee/gsg}.
\end{abstract}

\section{Introduction}
\label{sec:introduction}

Representation learning has been a critical topic in machine learning. In visual representation learning, image representations containing high-level semantic information (e.g., visual concept) are learned for efficient training in downstream tasks. Because human annotation is labor-intensive and imperfect, learning representations without labels is getting more attention. In many cases, unsupervised or self-supervised learning (SSL) has surpassed its supervised counterpart.

In SSL \citep{jaiswal2020survey, jing2020self, liu2021self}, there are roughly two branches of algorithms. One branch is a set of algorithms trained on pretext tasks with pseudo labels. Examples of the pretext tasks are predicting relative positions \citep{doersch2015unsupervised}, solving jigsaw puzzles \citep{noroozi2016unsupervised}, colorization \citep{zhang2016colorful}, and identifying different rotations \citep{gidaris2018unsupervised}. However, relying on a specific pretext task can restrict the generality of the learned representations. The other branch is a set of algorithms trained by maximizing the agreement of representations of randomly augmented views from the same image. Many algorithms in this branch adopt Siamese networks with two parallel encoders as their architecture. Siamese networks are natural architecture with a minimal inductive bias to learn transformation-invariant representations. However, na\"ive use of Siamese networks can result in collapse, i.e., a constant representation for all images.  

\begin{figure}[t!]
\vskip 0.2in
     \centering
     \hspace*{\fill}%
     \begin{subfigure}[t]{0.42\linewidth}
         \centering
         \includegraphics[width=\columnwidth]{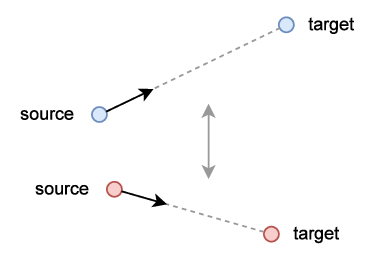}
         \caption{Implicit contrastive representation learning.}
         \label{fig:implicit_contrastive_learning}
     \end{subfigure}
     \hfill%
     \begin{subfigure}[t]{0.42\linewidth}
         \centering
         \includegraphics[width=\columnwidth]{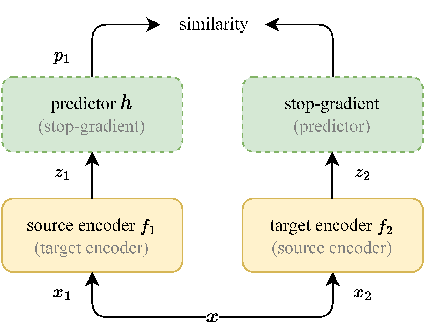}
         \caption{Symmetrized use of asymmetric networks.}
         \label{fig:asymmetric_symmetric}
     \end{subfigure}
     \hspace*{\fill}%
        \caption{(a) Dots of the same color are representations of a positive pair. Without contrastive loss, it aims for a repelling effect by carefully determining which to make the source representation and which to make the target representation. (b) In SimSiam and BYOL, a given image $x$ is randomly transformed into two views $x_1$ and $x_2$. The views are processed by encoders $f_1$ and $f_2$ to have projections $z_1$ and $z_2$. A predictor is applied on one side, and stop-gradient is applied on the other. Then, the similarity between the outputs from both sides is maximized. By using the predictor and stop-gradient alternately, a symmetric loss is constructed.}
        \label{fig:intuition}
\vskip -0.2in
\end{figure}

To tackle the collapse problem, there have been four strategies. The first strategy is contrastive learning \citep{bachman2019learning, ye2019unsupervised, chen2020simple, chen2020improved, he2020momentum, misra2020self, chen2021empirical}. Contrastive learning uses positive pairs (views of the same image) and negative pairs (views from different images). Minimizing contrastive loss encourages representations of negative pairs to push each other while representations of positive pairs pull each other. The second strategy is to use clustering \citep{caron2018deep, asano2019self, caron2019unsupervised, caron2020unsupervised, li2020prototypical}. It clusters the representations and then predicts the cluster assignment. Repeating this process keeps the cluster assignments consistent for positive pairs. The third strategy is to decorrelate between other variables of representations while maintaining the variance of each variable so that it does not get small \citep{bardes2021vicreg, zbontar2021barlow}. The fourth strategy is to break the symmetry of Siamese networks \citep{caron2020unsupervised, grill2020bootstrap, he2020momentum, caron2021emerging, chen2021exploring}. Many algorithms using this strategy make the representation from one encoder (called source encoder) follow the representation from the other encoder (called target encoder)\footnote{The source encoder is also called online, query, or student encoder, and the target encoder is also called key or teacher encoder in the literature depending on the context \citep{wang2022importance, tarvainen2017mean, grill2020bootstrap, he2020momentum, caron2021emerging}.}.

In this paper, we introduce a methodology to do contrastive learning implicitly by leveraging asymmetric relation between the source and target encoders. For a given positive pair, one of its representations becomes the source representation (from the source encoder), and the other becomes the target representation (from the target encoder). The target representation attracts the source representation. By investigating how representations are located in the embedding space, we carefully determine which representation will be the source (or target) representation so that representations of negative pairs repel each other (Figure \ref{fig:implicit_contrastive_learning}). There is no explicit contrastive part in our loss, thus the name implicit contrastive representation learning. The main idea of the methodology can be expressed as follows: \emph{Repel in the service of attracting}. We also present our guided stop-gradient method, an instance of the methodology. Our method can be applied to existing algorithms SimSiam \citep{chen2021exploring} and BYOL \citep{grill2020bootstrap}. We show that by applying our method, the performance of the original algorithms is boosted on various tasks and datasets.

The technical contributions of our work can be summarized as follows:
\begin{itemize}
\item We introduce a new methodology called implicit contrastive representation learning that exploits the asymmetry of network architecture for contrastive learning (Section \ref{sec:main_idea} and \ref{sec:method}).
\item We present new algorithms by applying our method to benchmark algorithms SimSiam and BYOL and show performance improvements in various tasks and datasets (Section \ref{sec:comparison}).
\item We demonstrate through empirical studies that our idea can be used to improve training stability to help prevent collapse, which is a fundamental problem of SSL (Section \ref{sec:empirical_study}).
\end{itemize}

\section{Related work}
\label{sec:related_work}

\textbf{Siamese networks} Siamese networks are symmetric in many respects. There are encoders of the same structure on both sides, and they share weights. The inputs to the two encoders have the same distribution, and the outputs are induced to be similar by the loss. On the one hand, this symmetric structure helps to learn transformation-invariance well, but on the other hand, all representations risk collapsing to a trivial solution. There have been several approaches to solving this collapse problem, and the approaches related to our work are contrastive learning and asymmetric learning. So, we introduce them in more detail below.

\textbf{Contrastive learning} Contrastive learning \citep{hadsell2006dimensionality, wang2020understanding, wu2018unsupervised, hjelm2018learning, tian2020contrastive} can be characterized by its contrastive loss \citep{le2020contrastive, chopra2005learning}. The basic idea is to design the loss so that representations of positive pairs pull together and representations of negative pairs push away. Through the loss, representations are formed where equilibrium is achieved between the pulling and pushing forces. Contrastive losses that have been used so far include noise contrastive estimation (NCE) loss \citep{gutmann2010noise}, triplet loss \citep{schroff2015facenet}, lifted structured loss \citep{oh2016deep}, multi-class $N$-pair loss \citep{sohn2016improved}, InfoNCE loss \citep{oord2018representation}, soft-nearest neighbors loss \citep{frosst2019analyzing}, and normalized-temperature cross-entropy loss \citep{chen2020simple}. All the losses mentioned above contain explicit terms in their formula that cause the negative pairs to repel each other. Under self-supervised learning scenarios, since we don't know the labels, a negative pair simply consists of views from different images. Then, in practice, there is a risk that the negative pair consists of views from images with the same label, and performance degradation occurs due to this sampling bias \citep{chuang2020debiased}. In addition, contrastive learning algorithms are sensitive to changes in the number of negative samples, so performance deteriorates when the number of negative samples is small.

\textbf{Asymmetric learning} Another approach to avoid collapse is to break the symmetry of Siamese networks and introduce asymmetry \citep{wang2022importance}. Asymmetry can be imposed on many aspects, including data, network architectures, weights, loss, and training methods. In MoCo \citep{he2020momentum}, the encoders do not simply share weights, and the weights of one encoder (key encoder) are a moving average of the weights of the other encoder (query encoder). This technique of slowly updating an encoder is called a momentum encoder. SwAV \citep{caron2020unsupervised} and DINO \citep{caron2021emerging} apply a multi-crop strategy when performing data augmentation to make the distribution of inputs into encoders different. Also, the loss applied to the outputs from the encoders is asymmetric. Furthermore, in DINO, stop-gradient is applied to the output of one encoder (teacher encoder) and not the other (student encoder). In SimSiam \citep{chen2021exploring} and BYOL \citep{grill2020bootstrap}, an additional module called predictor is stacked on one encoder (source encoder), and stop-gradient is applied to the output from the other encoder (target encoder). Compared to SimSiam, in BYOL, the target encoder is a momentum encoder.

\section{Main idea}
\label{sec:main_idea}

In this section, we explain our motivation and intuition behind our method using an example.

\textbf{Asymmetric architecture} In SimSiam and BYOL, for a given image $x$, two views $x_1$ and $x_2$ are generated by random transformations. The views are fed to encoders $f_1$ and $f_2$ to yield projections $z_1$ and $z_2$. An encoder is a backbone plus a projector\footnote{Ultimately, we use representations from the backbone, but it is common practice to compose the loss with projections from the projector, i.e., the encoder \citep{chen2020simple}.}. Then, a predictor $h$ is applied to one encoder (source encoder) to have a prediction, and stop-gradient\footnote{Applying stop-gradient to $z$ means treating $z$ as a constant. In actual implementation, $z$ is detached from the computational graph to prevent the propagation of the gradient.} is applied to the other encoder (target encoder). The algorithms maximize the similarity between the resulting prediction and projection. The difference between SimSiam and BYOL is in the weights of the encoders. In SimSiam, the source and target encoder share the weights. On the other hand, in BYOL, the target encoder is a momentum encoder. That is, the weights of the target encoder are an exponential moving average of the weights of the source encoder. Due to the existence of the predictor and stop-gradient (also momentum encoder in the case of BYOL), the algorithms have asymmetric architecture (Figure \ref{fig:asymmetric_symmetric}).

\textbf{Symmetric loss} Despite the asymmetry of the architecture, the losses in SimSiam and BYOL are symmetrized. After alternately applying a predictor and stop-gradient to the two encoders, the following loss is obtained by adding the resulting loss terms:
\begin{equation}
\mathcal{L} = \frac{1}{2}\mathcal{D}(p_1, \text{sg}(z_2)) + \frac{1}{2}\mathcal{D}(p_2, \text{sg}(z_1)),
\label{eq:loss_terms}
\end{equation}
where $\mathcal{D}(\cdot,\cdot)$ denotes negative cosine similarity, i.e., $\mathcal{D}(p,z)=-(p/\lVert p\rVert_2)\cdot (z/\lVert z\rVert_2)$, and $\text{sg}(\cdot)$ denotes the stop-gradient operator. The range of possible values for the loss is $[-1, 1]$. Minimizing the first term brings $p_1$ (closely related to $z_1$) closer to $z_2$, and minimizing the second term brings $p_2$ (closely related to $z_2$) closer to $z_1$. By minimizing the loss, we want to move $z_1$ in the direction of $z_2$ and $z_2$ in the direction of $z_1$.

\begin{wrapfigure}{r}{0.5\linewidth}
\vspace{-20pt}
    \begin{subfigure}[t]{0.49\linewidth}
        \centering
        \includegraphics[width=\linewidth]{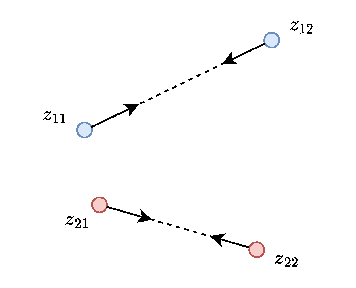}
        \caption{Loss terms for two imgs.}
        \label{fig:main_idea_1}
    \end{subfigure}
    \begin{subfigure}[t]{0.49\linewidth}
        \centering
        \includegraphics[width=\linewidth]{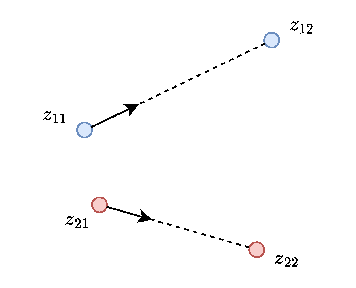}
        \caption{Selected loss terms.}
        \label{fig:main_idea_2}
    \end{subfigure}
    \caption{An example for two images. The dots represent four projections of the two images. The arrows represent the expected effect of the loss terms. We want dots of the same color to come close to each other. We select loss terms so that two closest dots with different colors will fall apart.}
    \label{fig:main_idea}
\vspace{-10pt}
\end{wrapfigure}

\textbf{Guided stop-gradient} The main idea of our guided stop-gradient method is to design an asymmetric loss to leverage the asymmetry of the architecture. In other words, we select one of the two loss terms in Equation (\ref{eq:loss_terms}) to help with training. However, it is known that randomly constructing an asymmetric loss is not beneficial \citep{chen2021exploring}. To do this systematically, we consider two different images $x_1$ and $x_2$ simultaneously and let each other be a reference. When trying to bring the representations of one image closer, we give a directional guide on which representation to apply stop-gradient to by utilizing the geometric relationship with the representations of the other reference image.

Specifically, let $z_{11}$ ($p_{11}$) and $z_{12}$ ($p_{12}$) be projections (predictions) from the image $x_1$, and $z_{21}$ ($p_{21}$) and $z_{22}$ ($p_{22}$) be projections (predictions) from the image $x_2$. Then, the symmetrized loss for two images will be as follows (Figure \ref{fig:main_idea_1}):

\begin{equation}
\mathcal{L} = \frac{1}{4}\mathcal{D}(p_{11}, \text{sg}(z_{12})) + \frac{1}{4}\mathcal{D}(p_{12}, \text{sg}(z_{11})) + \frac{1}{4}\mathcal{D}(p_{21}, \text{sg}(z_{22})) + \frac{1}{4}\mathcal{D}(p_{22}, \text{sg}(z_{21})).
\label{eq:four_loss_terms}
\end{equation}

Now, we select one term from the first two terms (terms to maximize agreement between $z_{11}$ and $z_{12}$) and another term from the remaining two terms (terms to maximize agreement between $z_{21}$ and $z_{22}$). One view of the image $x_1$ and one view of the image $x_2$ form a negative pair. So we want the projections of the image $x_1$ and the projections of the image $x_2$ not to be close to each other. In the example of Figure \ref{fig:main_idea}, since $z_{11}$ and $z_{21}$ are closest, we apply a predictor\footnote{In the case of SimSiam and BYOL, the presence of a predictor can make the interpretation tricky because when we move two close points $z$ and $z'$, we move them indirectly by moving $p=h(z)$ and $p'=h(z')$ through the predictor $h$. We assume that the predictor $h$ has a good regularity as a function, that is, if $\lVert z - z' \rVert_2$ is small, $\lVert h(z) - h(z') \rVert_2$ is also small. So, by trying to separate $p$ and $p'$, we separate $z$ and $z'$. Note that in practice, the predictor $h$ is usually an MLP with a few layers.} to $z_{11}$ and $z_{21}$ and apply stop-gradient to $z_{12}$ and $z_{22}$ to try to separate the projections. Then, the resulting loss will be as follows (Figure \ref{fig:main_idea_2}):
\begin{equation}
\mathcal{L} = \frac{1}{2}\mathcal{D}(p_{11}, \text{sg}(z_{12})) + \frac{1}{2}\mathcal{D}(p_{21}, \text{sg}(z_{22})).
\label{eq:selected_loss_terms}
\end{equation}
Selecting loss terms is equivalent to determining which of the two projections of each image to apply stop-gradient. Since we do this in a guided way by observing how the projections are located, we call the method Guided Stop-Gradient (GSG). In this way, by continuously moving toward representations that are not close together, representations of negative pairs are induced to spread well in the long run.

\textbf{Implicit contrastive representation learning} In our loss, there is no explicit part where the projections of a negative pair repulse each other. However, the loss we designed implicitly does it. We aim for a contrastive effect by making good use of the fact that the source projections go after the target projections in an asymmetric network architecture. Therefore, SimSiam and BYOL with GSG can also be viewed as a mixture of contrastive learning and asymmetric learning.

\begin{figure*}[t!]
\vskip 0.2in
\begin{center}
\centerline{\includegraphics[width=\linewidth]{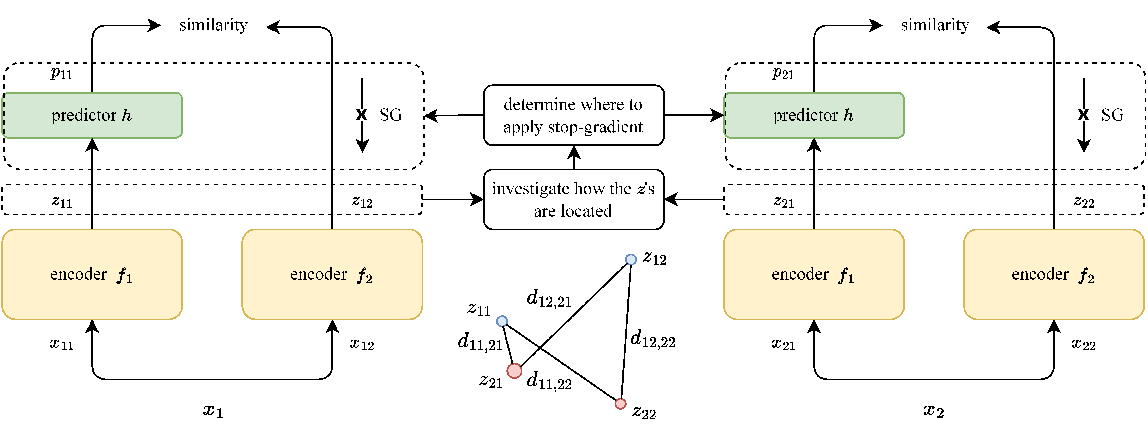}}
\caption{Overview of our guided stop-gradient method. (1) The encoders process two images $x_1$, $x_2$ that are reference to each other. (2) Investigate the distances $d_{11,21}$, $d_{11,22}$, $d_{12,21}$, and $d_{12,22}$ between the projections of negative pairs. (3) Determine which side to apply stop-gradient and which to apply a predictor.}
\label{fig:architecture}
\end{center}
\vskip -0.2in
\end{figure*}

\section{Method}
\label{sec:method}

In this section, we explain the specific application of GSG based on the main idea described above.

For a given batch of images for training, a batch of pairs of images is created by matching the images of the batch with the images of the shuffled batch one by one. Then, a pair of images in the resulting batch consists of two images, $x_1$ and $x_2$. Note that since the images are paired within a given batch, the batch size is the same as the existing algorithm. By applying random augmentation, we generate views $x_{11}$ and $x_{12}$ from the image $x_1$ and views $x_{21}$ and $x_{22}$ from the image $x_2$. The views are processed by an encoder $f$ to yield projections $z_{11}$ and $z_{12}$ of the image $x_1$ and projections $z_{21}$ and $z_{22}$ of the image $x_2$.

Let $d_{ij,kl}$ denote the Euclidean distance between projections $z_{ij}$ and $z_{kl}$, i.e.,
\begin{equation}
d_{ij,kl}=\lVert z_{ij} - z_{kl} \rVert_2,
\end{equation}
where $\lVert \cdot \rVert_2$ is $l_2$-norm.
We investigate distances $d_{11,21}$, $d_{11,22}$, $d_{12,21}$, and $d_{12,22}$. Note that there are four cases in total since we are looking at the distance between one of the two projections of $x_1$ and one of the two projections of $x_2$. Now, we find the minimum $m$ of the distances. That is,
\begin{equation}
m = \min \{d_{11,21}, d_{11,22}, d_{12,21}, d_{12,22}\}.
\end{equation}
We apply a predictor to the two projections corresponding to the smallest of the four distances and stop-gradient to the remaining two projections. Depending on which distance is the smallest, the loss is as follows:
\begin{equation}
\mathcal{L}=\begin{cases}
			\frac{1}{2}\mathcal{D}(p_{11}, \text{sg}(z_{12})) + \frac{1}{2}\mathcal{D}(p_{21}, \text{sg}(z_{22})), & \text{if $m=d_{11,21}$}\\
            \frac{1}{2}\mathcal{D}(p_{11}, \text{sg}(z_{12})) + \frac{1}{2}\mathcal{D}(p_{22}, \text{sg}(z_{21})), & \text{if $m=d_{11,22}$}\\
            \frac{1}{2}\mathcal{D}(p_{12}, \text{sg}(z_{11})) + \frac{1}{2}\mathcal{D}(p_{21}, \text{sg}(z_{22})), & \text{if $m=d_{12,21}$}\\
            \frac{1}{2}\mathcal{D}(p_{12}, \text{sg}(z_{11})) + \frac{1}{2}\mathcal{D}(p_{22}, \text{sg}(z_{21})), & \text{if $m=d_{12,22}$}.
		 \end{cases}
\label{eq:loss}
\end{equation}
For a better understanding, refer to Figure \ref{fig:architecture} and Appendix A. For simplicity, we present the overview and pseudocode for SimSiam with GSG, but they are analogous to BYOL with GSG.

\begin{table*}[t]
  \caption{Comparison of representation quality under standard evaluation protocols.}
  \label{tab:evaluate}
  \centering
  \begin{tabular}{lcccc}
    \toprule
    Algorithm &  \multicolumn{2}{c}{ImageNet} &  \multicolumn{2}{c}{CIFAR-10}\\
    \cmidrule(lr){2-3}
    \cmidrule(lr){4-5}
     & $k$-NN acc. ($\%$) & Linear acc. ($\%$) & $k$-NN acc. ($\%$) & Linear acc. ($\%$)\\
    \midrule
    SimSiam & 51.7$\pm$0.11 & 67.9$\pm$0.09 & 77.0$\pm$0.67 & 82.7$\pm$0.26 \\
    SimSiam w/ GSG & \textbf{58.4$\pm$0.17} & \textbf{69.4$\pm$0.02} & \textbf{82.2$\pm$0.48} & \textbf{86.4$\pm$0.28} \\
    \midrule
    BYOL & 56.5$\pm$0.16 & 69.9$\pm$0.02 & 85.4$\pm$0.24 & 88.0$\pm$0.09\\
    BYOL w/ GSG & \textbf{62.2$\pm$0.06} & \textbf{71.1$\pm$0.12} & \textbf{89.4$\pm$0.21} & \textbf{90.3$\pm$0.16}\\
    \bottomrule
  \end{tabular}
\end{table*}

\begin{table*}[t]
  \caption{Comparison in transfer learning for image recognition.}
  \label{tab:transfer_learning_image_recognition}
  \centerline{\begin{tabular}{lcccccccccc}
    \toprule
    Algorithm & CIFAR-10 & Aircraft & Caltech & Cars & DTD & Flowers & Food & Pets & SUN397 & VOC2007 \\
    \midrule
    SimSiam & 90.0 & 39.7 & 86.5 & 31.8 & 70.9 & 88.8 & 61.6 & 81.4 & 57.8 & 80.5 \\
    SimSiam w/ GSG & \textbf{92.3} & \textbf{47.0} & \textbf{89.6} & \textbf{41.2} & \textbf{73.0} & \textbf{89.5} & \textbf{64.3} & \textbf{85.1} & \textbf{59.8} & \textbf{82.6} \\
    \midrule
    BYOL & 91.0 & 42.5 & 88.9 & 39.3 & 71.7 & 89.1 & 64.0 & 85.3 & 60.4 & 82.2 \\
    BYOL w/ GSG & \textbf{93.6} & \textbf{47.1} & \textbf{89.9} & \textbf{46.9} & \textbf{72.6} & \textbf{89.5} & \textbf{67.1} & \textbf{89.1} & \textbf{61.6} & \textbf{82.7} \\
    \bottomrule
  \end{tabular}}
\end{table*}

\section{Comparison}
\label{sec:comparison}

In this section, we compare SimSiam and BYOL with GSG to the original SimSiam and BYOL on various datasets and tasks. For a fair comparison, we use the same experimental setup for all four algorithms on each dataset and task. For example, all algorithms perform the same number of gradient updates and exploit the same number of images in each update. We implement the algorithms with Pytorch \citep{paszke2019pytorch} and run all the experiments on 8 NVIDIA A100 GPUs. Our algorithms take about two days for ImageNet pre-training and 12 hours for linear evaluation.

\subsection{Pre-training}
\label{subsec:pre-training}

We first pre-train networks in an unsupervised manner. The trained networks will later be used in downstream tasks. We use ImageNet \citep{deng2009imagenet} and CIFAR-10 \citep{krizhevsky2009learning} as benchmark datasets. Refer to Appendix B for data augmentation details.

For ImageNet, we use the ResNet-50 backbone \citep{he2016deep}, a three-layered MLP projector, and a two-layered MLP predictor. We use a batch size of 512 and train the network for 100 epochs. We use the SGD optimizer with momentum of $0.9$, learning rate of $0.1$, and weight decay rate of $0.0001$. We use a cosine decay schedule \citep{chen2020simple, loshchilov2016sgdr} for the learning rate. 

For CIFAR-10, we use a CIFAR variant of the ResNet-18 backbone, a two-layered MLP projector, and a two-layered MLP predictor. We use a batch size of 512 and train the network for 200 epochs. We use the SGD optimizer with momentum of $0.9$, learning rate of $0.06$, and weight decay rate of $0.0005$. We do not use a learning rate schedule since no scaling shows better training stability. Refer to Appendix C for other implementation details. 

After pre-training, we obtain representations of the images with the trained backbone and then evaluate their quality. We use both $k$-nearest neighbors \citep{wu2018unsupervised} and linear evaluation, which are standard evaluation protocols.

\textbf{$k$-nearest neighbors} For a given test set image, we find its representation and obtain $k$ training set images with the closest representation. Then, we determine the predicted label of the image by majority voting of the labels of the $k$ images. We set $k=200$ for ImageNet and $k=1$ for CIFAR-10.

\textbf{Linear evaluation} We freeze the trained backbone, attach a linear classifier to the backbone, fit the classifier on the training set in a supervised manner for 90 epochs, and test the classifier on the test set. For ImageNet, we use a batch size of $4096$ and the LARS optimizer \citep{you2017large}, which can work well with a large batch size. For CIFAR-10, we use a batch size of $256$ and the SGD optimizer with momentum of $0.9$, learning rate of $30$, and a cosine decay schedule. We report top-$1$ accuracy for all cases.

\textbf{Results} Table \ref{tab:evaluate} shows that applying GSG consistently increases the performance. We report error bars (mean $\pm$ standard deviation) by running each algorithm three times independently and show that our method improves the performance reliably. The performance gain is obtained by keeping all other experimental setups the same and changing only the stop-gradient application method. This suggests that there is room for improvement in the symmetrized use of asymmetric networks in the existing algorithms.

\subsection{Transfer learning}
\label{subsec:transfer_learning}

An essential goal of representation learning is to obtain a pre-trained backbone that can be transferred to various downstream tasks. To evaluate whether our pre-trained backbones are transferable, we consider image recognition, object detection, and semantic segmentation tasks. We use the ResNet-50 backbones pre-trained on ImageNet. We follow the experimental setup in \citep{ericsson2021well}. More implementation details can be found in Appendix D.

\textbf{Image recognition} We carry out image recognition tasks on different datasets. For datasets, we adopt widely used benchmark datasets in transfer learning such as CIFAR-10, Aircraft \citep{maji2013fine}, Caltech \citep{fei2004learning}, Cars \citep{krause2013collecting}, DTD \citep{cimpoi2014describing}, Flowers \citep{nilsback2008automated}, Food \citep{bossard2014food}, Pets \citep{parkhi2012cats}, SUN397 \citep{xiao2010sun}, and VOC2007 \citep{everingham2010pascal}. These datasets vary in terms of the amount of data or the number of classes. We report the average precision AP at 11 recall levels $\{0,0.1,...,1\}$ on VOC2007, mean per-class accuracy on Aircraft, Pets, Caltech, and Flowers, and top-1 accuracy on the rest of the datasets. For the evaluation protocol, we perform the linear evaluation.

\begin{table}[t!]
  \caption{Comparison in transfer learning for object detection and semantic segmentation.}
  \label{tab:transfer_learning_object_detection}
  \centerline{\begin{tabular}{lcccccccc}
    \toprule
    Algorithm & \multicolumn{3}{c}{VOC detection} & \multicolumn{3}{c}{COCO detection} & \multicolumn{2}{c}{Semantic segmentation}\\
    \cmidrule(lr){2-4}
    \cmidrule(lr){5-7}
    \cmidrule(lr){8-9}
    & $\text{AP}_{50}$ & AP & $\text{AP}_{75}$ & $\text{AP}_{50}$ & AP & $\text{AP}_{75}$ & Mean IoU ($\%$) & Pixel acc. ($\%$) \\
    \midrule
    SimSiam & 77.0 & 48.8 & 52.2 & 50.7 & 31.2 & 32.8 & 0.2626 & 65.63 \\
    SimSiam w/ GSG & \textbf{79.8} & \textbf{51.2} & \textbf{55.1} & \textbf{53.2} & \textbf{33.4} & \textbf{35.6} & \textbf{0.3345} & \textbf{76.57} \\
    \midrule
    BYOL & 79.2 & 50.3 & 54.5 & 52.0 & 32.5 & 34.5 & 0.2615 & 62.79 \\
    BYOL w/ GSG & \textbf{80.5} & \textbf{52.0} & \textbf{56.4} & \textbf{53.7} & \textbf{33.8} & \textbf{36.1} & \textbf{0.2938} & \textbf{74.78} \\
    \bottomrule
  \end{tabular}}
\end{table}

\begin{table}[t!]
  \caption{Comparison between explicit and implicit contrastive learning algorithms.}
  \label{tab:the_number_of_negative_samples}
  \centering
  \begin{tabular}{llccc}
    \toprule
    Methodology & Algorithm & $b=1024$ & $b=512$ & $b=256$ \\
    \midrule
    Explicit contrastive & End-to-end \citep{he2020momentum} & 57.3 & 56.3 & 54.9 \\
    & InstDisc \citep{wu2018unsupervised} & 54.1 & 52.0 & 50.0 \\
    & MoCo \citep{he2020momentum} & 57.5 & 56.4 & 54.7 \\
    & SimCLR \citep{chen2020simple} & 62.8 & 60.7 & 57.5 \\
    \midrule
    Implicit contrastive & SimSiam w/ GSG & 70.1 & 69.4 & 69.9 \\
    & BYOL w/ GSG & 71.9 & 71.0 & 71.6 \\
    \bottomrule
  \end{tabular}
\end{table}

\textbf{Object detection} We perform object detection tasks on Pascal-VOC \citep{everingham2010pascal} and MS-COCO \citep{lin2014microsoft}. We use VOC2007 trainval as the training set and VOC2007 test as the test set. We report $\text{AP}_{50}$, AP, and $\text{AP}_{75}$. $\text{AP}_{50}$ and $\text{AP}_{75}$ are average precision with intersection over union (IoU) threshold 0.5 and 0.75, respectively. We freeze the pre-trained backbone except for the last residual block. We use a Feature Pyramid Network \citep{lin2017feature} to extract representations, and a Faster R-CNN \citep{ren2015faster} to predict. We do the experiments on the Detectron2 platform \citep{wu2019detectron2}.

\textbf{Semantic segmentation} We conduct semantic segmentation tasks on MIT ADE20K \citep{zhou2019semantic}. We use ResNet-50 as the encoder and use UPerNet \citep{xiao2018unified} (the implementation in the CSAIL semantic segmentation framework \citep{zhou2018semantic, zhou2017scene}) as the decoder. It is based on Feature Pyramid Network and Pyramid Pooling Module \citep{zhao2017pyramid}. We train for 30 epochs and test on the validation set. We report mean IoU and pixel accuracy. For the predicted results of each algorithm, refer to Appendix F.

\textbf{Results} Table \ref{tab:transfer_learning_image_recognition} and \ref{tab:transfer_learning_object_detection} show that applying GSG increases the performance consistently. We can conclude that the pre-trained backbones are transferable to different tasks, and our method helps to get better quality representations.

\subsection{The number of negative samples}
\label{subsec:the_number_of_negative_samples}

It is known that the performance of contrastive learning algorithms is vulnerable to reducing the number of negative samples \citep{tian2020contrastive, wu2018unsupervised, he2020momentum, chen2020simple}. In this respect, we compare our algorithms to benchmark contrastive learning algorithms (all with the ResNet-50 backbone). Table \ref{tab:the_number_of_negative_samples} reports linear evaluation accuracy on ImageNet (the performance of the benchmark algorithms is from \citep{he2020momentum} and \citep{chen2020simple}). End-to-end, SimCLR, and our algorithms use samples in the batch as negative samples, so for these algorithms, $b$ in the table denotes the batch size. InstDisc and MoCo maintain a separate memory bank or dictionary, so for these algorithms, $b$ in the table denotes the number of negative samples from the memory bank or dictionary. The table shows that our algorithms work well with small batch sizes.

\begin{figure*}[t!]
\vskip 0.2in
    \centering
    \begin{subfigure}[b]{0.35\linewidth}
        \centering
        \includegraphics[width=\linewidth]{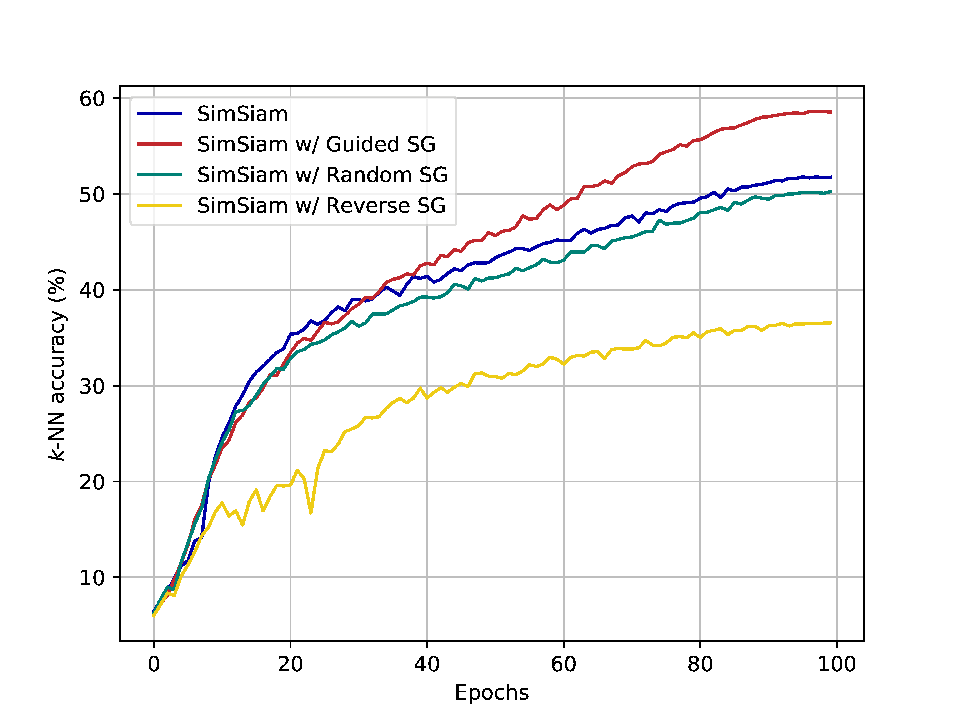}
        \caption{ImageNet, SimSiam}    
        \label{subfig:guide_imagenet_simsiam}
    \end{subfigure}
    \begin{subfigure}[b]{0.35\linewidth}  
        \centering 
        \includegraphics[width=\linewidth]{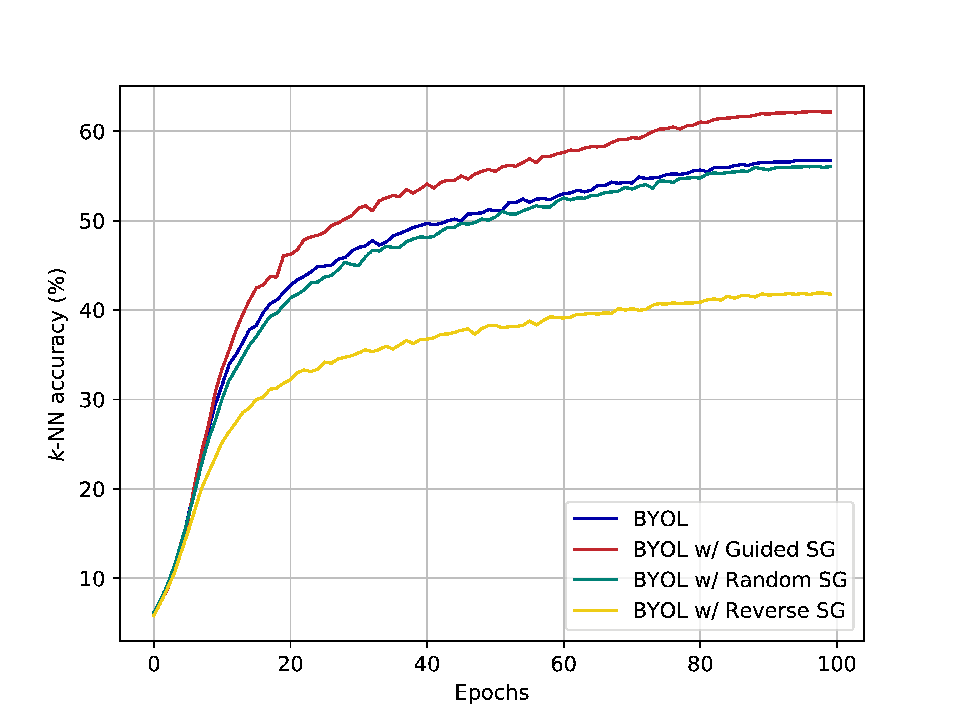}
        \caption{ImageNet, BYOL}    
        \label{subfig:guide_imagenet_byol}
    \end{subfigure}
    \begin{subfigure}[b]{0.35\linewidth}   
        \centering 
        \includegraphics[width=\linewidth]{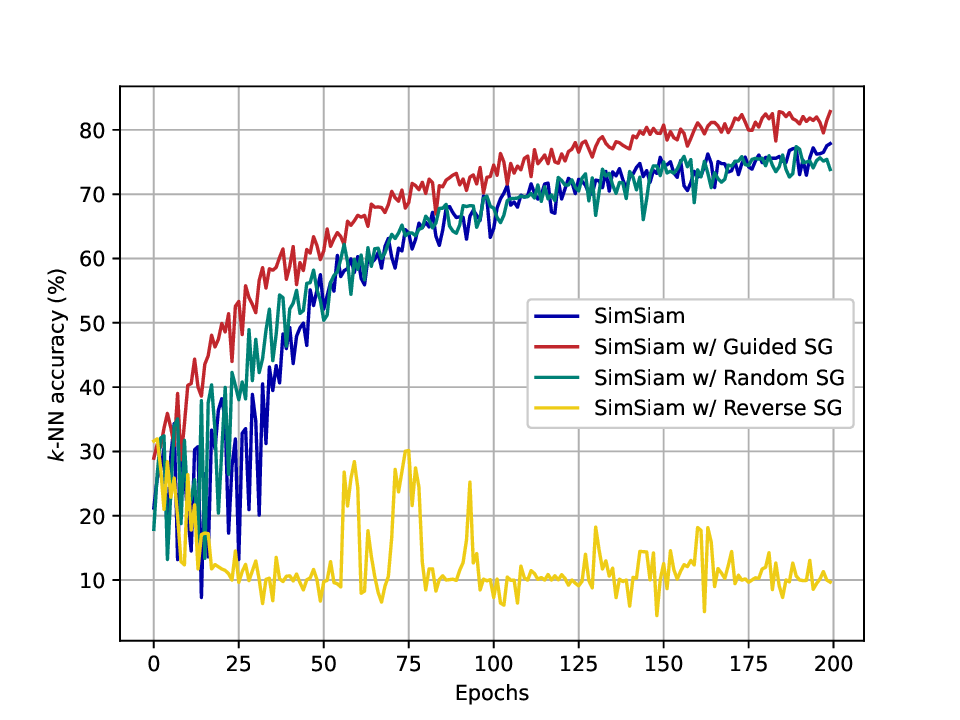}
        \caption{CIFAR-10, SimSiam}   
        \label{subfig:guide_cifar10_simsiam}
    \end{subfigure}
    \begin{subfigure}[b]{0.35\linewidth}   
        \centering 
        \includegraphics[width=\linewidth]{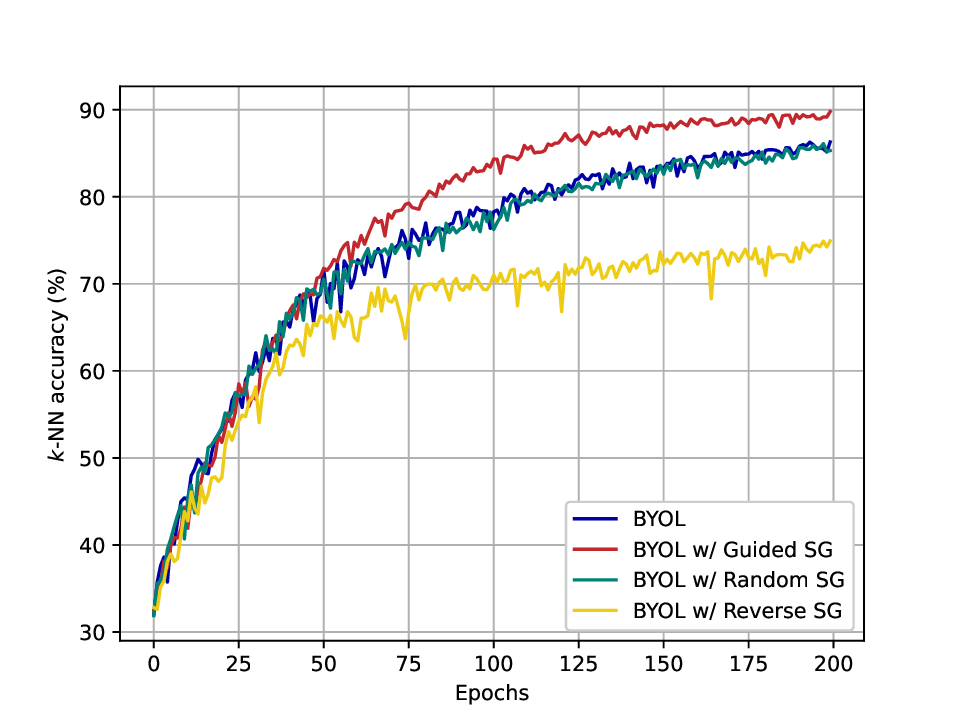}
        \caption{CIFAR-10, BYOL}    
        \label{subfig:guide_cifar10_byol}
    \end{subfigure}
    \caption{Importance of guiding. Depending on how stop-gradient is used, performance is significantly different. It shows the best performance when used along with GSG.} 
    \label{fig:guide}
\vskip -0.2in
\end{figure*}

\section{Empirical study}
\label{sec:empirical_study}

In this section, we broaden our understanding of stop-gradient, predictor, and momentum encoder, which are components of SimSiam and BYOL by comparing variants of the algorithms.

\subsection{Importance of guiding}
\label{subsec:importance_of_guiding}

We demonstrate that it is crucial to apply stop-gradient in a guided way by changing how stop-gradient is used. We compare our GSG with what we name random stop-gradient and reverse stop-gradient. In random stop-gradient, stop-gradient is randomly applied to one of the two encoders, and a predictor is applied to the other. That is, we randomly select one of the four equations in Equation (\ref{eq:loss}). On the other hand, in reverse stop-gradient, stop-gradient is applied opposite to the intuition in GSG. In other words, we select the remaining loss terms other than the selected loss terms when GSG is used. In Equation (\ref{eq:four_loss_terms}), two loss terms will be selected to form the loss as follows:
\begin{equation}
\mathcal{L} = \frac{1}{2}\mathcal{D}(p_{12}, \text{sg}(z_{11})) + \frac{1}{2}\mathcal{D}(p_{22}, \text{sg}(z_{21})).
\label{eq:reversely_selected_loss_terms}
\end{equation}
Therefore, the random stop-gradient is a baseline where stop-gradient is na\"ively applied, and the reverse stop-gradient is the worst-case scenario according to our intuition.

Figure \ref{fig:guide} shows the results of applying GSG, random stop-gradient, and reverse stop-gradient along with the existing algorithm for SimSiam and BYOL. We observe the $k$-NN accuracy at each epoch while training on ImageNet and CIFAR-10. First of all, in all cases, it can be seen that the algorithm applying our GSG outperforms algorithms using other methods. In addition, it can be seen that the performance of the existing algorithm and the algorithm to which random stop-gradient is applied are similar. When applying random stop-gradient, the number of loss terms is doubled, and half of them are randomly selected, so there is expected to be no significant difference from the existing algorithm. In the case of reverse stop-gradient, the performance drops significantly. This highlights the importance of using stop-gradient in a guided manner.

If we look at the case of CIFAR-10 (Figure \ref{subfig:guide_cifar10_simsiam} and Figure \ref{subfig:guide_cifar10_byol}), which has relatively much smaller data and is more challenging to train stably, we can obtain some more interesting results. First, when reverse stop-gradient is applied to SimSiam, it collapses, and the accuracy converges to 10$\%$, which is the chance-level accuracy of CIFAR-10. However, this was not the case for BYOL. This implies that the momentum encoder can help prevent collapse. Note that SimSiam and BYOL are identical in our experimental setup, except that BYOL has a momentum encoder. In addition, in the case of the existing SimSiam and SimSiam with random stop-gradient, the fluctuation of the accuracy at the beginning of training is severe. However, it is relatively less in the case of SimSiam with GSG. This suggests that GSG can help the stability of training.

\begin{figure*}[t!]
\vskip 0.2in
    \centering
    \begin{subfigure}[b]{0.24\linewidth}
        \centering
        \includegraphics[width=\linewidth]{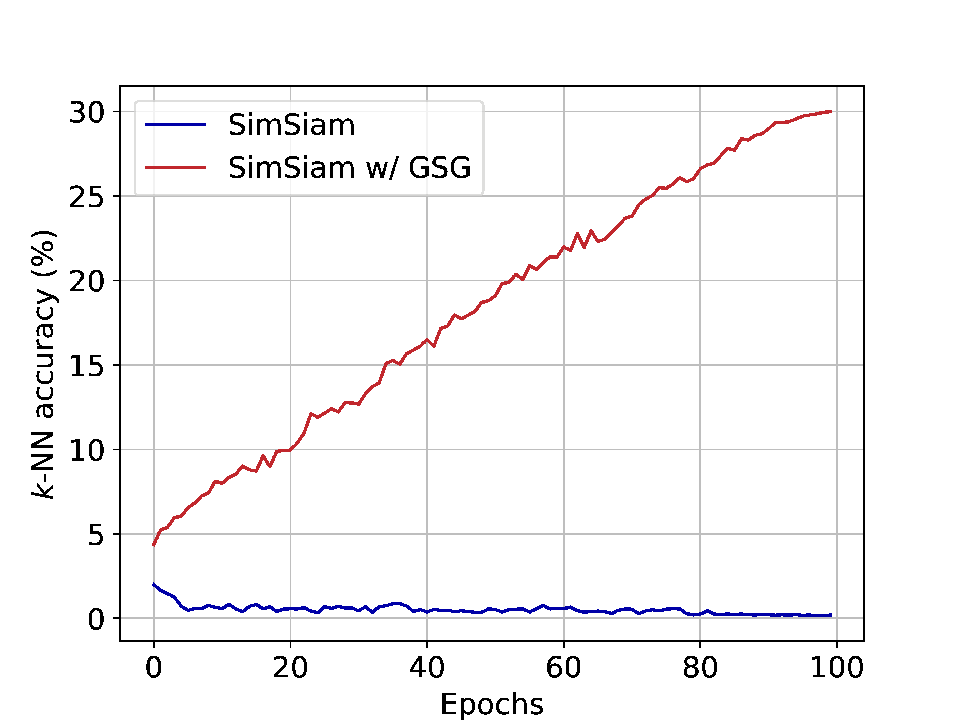}
        \caption{ImageNet, SimSiam}    
        \label{subfig:collapse_imagenet_simsiam}
    \end{subfigure}
    \begin{subfigure}[b]{0.24\linewidth}  
        \centering 
        \includegraphics[width=\linewidth]{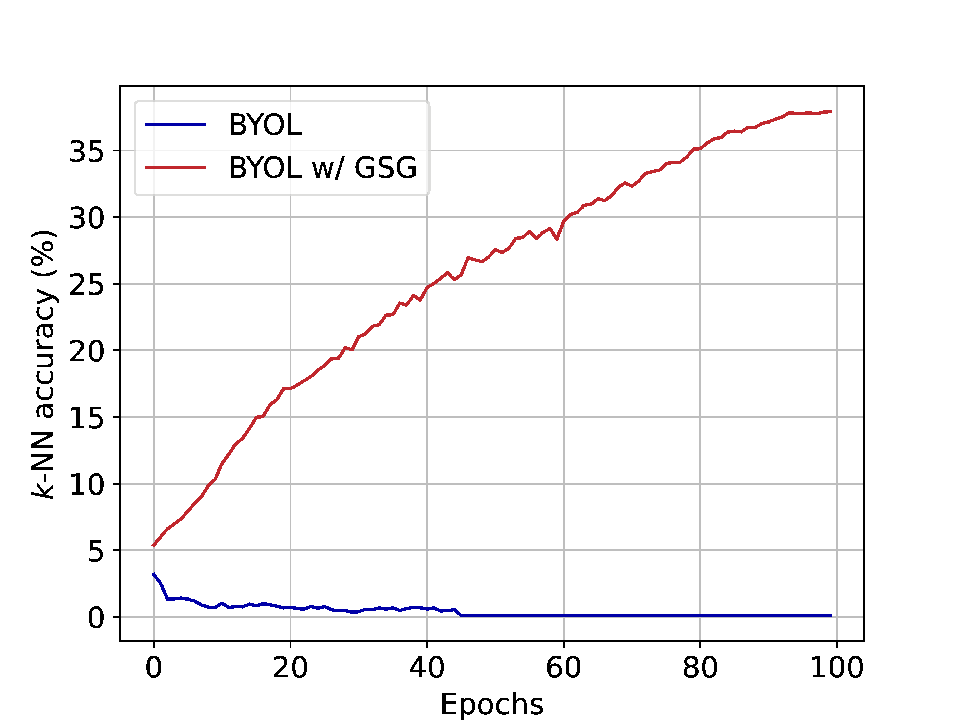}
        \caption{ImageNet, BYOL}    
        \label{subfig:collapse_imagenet_byol}
    \end{subfigure}
    \begin{subfigure}[b]{0.24\linewidth}   
        \centering 
        \includegraphics[width=\linewidth]{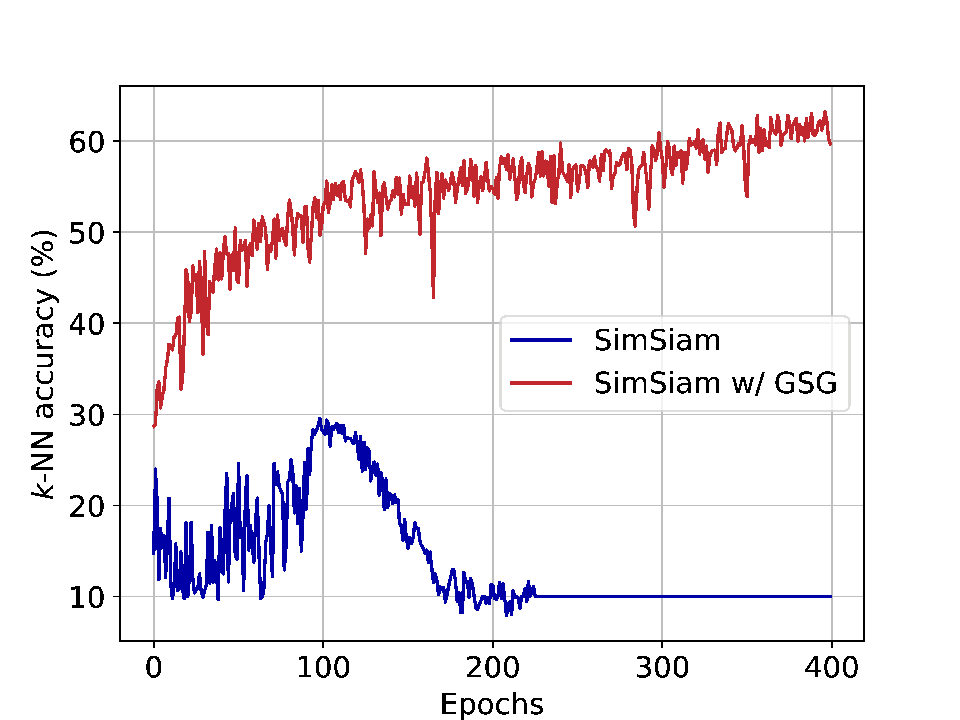}
        \caption{CIFAR-10, SimSiam}   
        \label{subfig:collapse_cifar10_simsiam}
    \end{subfigure}
    \begin{subfigure}[b]{0.24\linewidth}   
        \centering 
        \includegraphics[width=\linewidth]{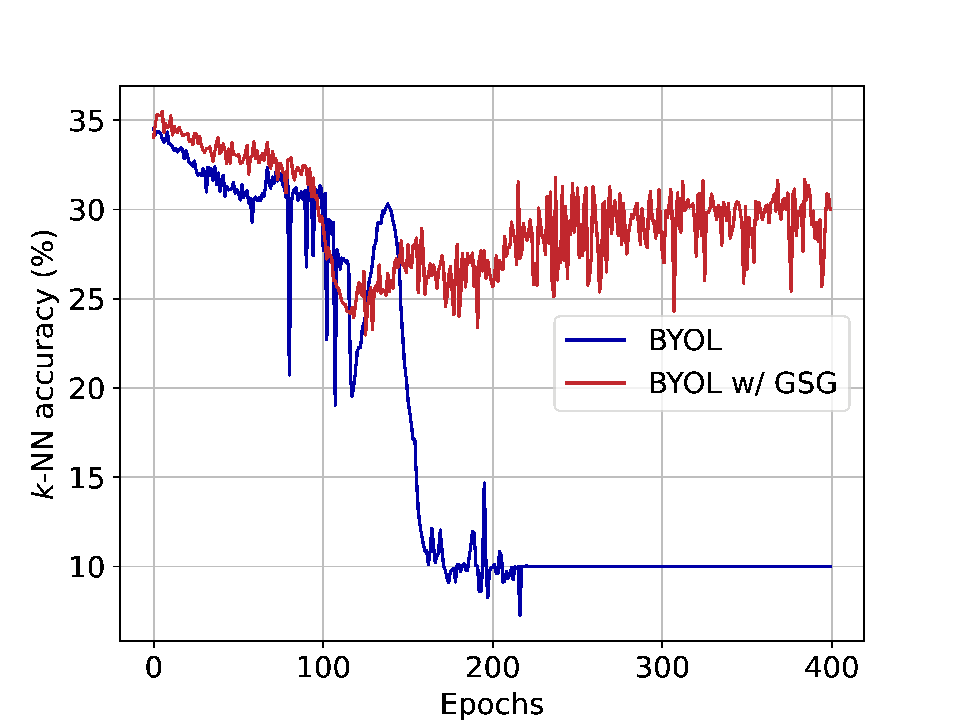}
        \caption{CIFAR-10, BYOL}    
        \label{subfig:collapse_cifar10_byol}
    \end{subfigure}
    \caption{Preventing collapse. Unlike existing algorithms, algorithms to which GSG is applied do not collapse even when the predictor is removed.}
    \label{fig:collapse}
\vskip -0.2in
\end{figure*}

\subsection{Preventing collapse}
\label{subsec:preventing_collapse}

One of the expected effects of GSG is to prevent collapse by implicitly repelling each other away from negative pairs. It is known that SimSiam and BYOL collapse without a predictor \citep{chen2021exploring, grill2020bootstrap}. Then, a natural question is whether SimSiam and BYOL will not collapse even if the predictor is removed when GSG is applied. Figure \ref{fig:collapse} reports the performance when the predictor is removed. In the case of CIFAR-10, we run up to 400 epochs to confirm complete collapse.

First, the existing algorithms collapse as expected, and the accuracy converges to the chance-level accuracy (0.1$\%$ for ImageNet and 10$\%$ for CIFAR-10). Interestingly, however, our algorithms do not collapse. This shows that GSG contributes to training stability, which is our method's intrinsic advantage. Nevertheless, the final accuracy is higher when there is a predictor. This indicates that the predictor in SimSiam and BYOL contributes to performance improvement.

\section{Conclusion}
\label{sec:conclusion}

We have proposed implicit contrastive representation learning for visual SSL. In this methodology, while representations of positive pairs attract each other, representations of negative pairs are promoted to repel each other. It exploits the asymmetry of network architectures with source and target encoders without contrastive loss. We have instantiated the methodology and presented our guided stop-gradient method, which can be applied to existing SSL algorithms such as SimSiam and BYOL. We have shown that our algorithms consistently perform better than the benchmark asymmetric learning algorithms for various tasks and datasets. We have also shown that our algorithms are more robust to reducing the number of negative samples than the benchmark contrastive learning algorithms. Our empirical study has shown that our method contributes to training stability. We hope our work leads the community to better leverage the asymmetry between source and target encoders and enjoy the advantages of both contrastive learning and asymmetric learning.

\bibliographystyle{plainnat}
\bibliography{wrapper}

\clearpage
\title{Implicit Contrastive Representation Learning with Guided Stop-gradient - Appendix}

\author{%
  Byeongchan Lee$^*$ \\
  Gauss Labs\\
  Seoul, Korea \\
  \texttt{byeongchan.lee@gausslabs.ai} \\
  \And
  Sehyun Lee$^*$ \\
  KAIST \\
  Daejeon, Korea \\
  \texttt{sehyun.lee@kaist.ac.kr} \\
}

\maketitleappendix

\setcounter{page}{1}
\renewcommand{\thesection}{\Alph{section}}
\setcounter{section}{0}  

\section{Algorithm}
\label{sec:algorithm}

We present a PyTorch-style pseudocode for SimSiam with GSG. It can be written analogously for BYOL with GSG. The pseudocode can be vectorized more, but we offer it in its current form for a better understanding.
\begin{algorithm}
\caption{SimSiam with GSG}
\begin{minted}[fontsize=\footnotesize]{python}
# pytorch style pseudocode

for batch in loader:
    loss = []
    for x1, x2 in zip(batch, shuffle(batch)):
        x11, x12 = aug(x1), aug(x1) # views
        x21, x22 = aug(x2), aug(x2) # aug: random augmentation
        
        z11, z12 = f(x11), f(x12) # projections
        z21, z22 = f(x21), f(x22) # f: encoder = backbone + projector (mlp)
        
        p11, p12 = h(z11), h(z12) # predictions
        p21, p22 = h(z21), h(z22) # h: predictor (mlp)
        
        d1121 = d(z11, z21) # Euclidean distances
        d1122 = d(z11, z22)
        d1221 = d(z12, z21)
        d1222 = d(z12, z22)
        m = min(d1121, d1122, d1221, d1222)
        
        # D: negative cosine similarity
        # sg: stop-gradient, sg(z) = z.detach()
        if d1121 == m:
            l = D(p11, sg(z12)) + D(p21, sg(z22))
        elif d1122 == m:
            l = D(p11, sg(z12)) + D(p22, sg(z21))
        elif d1221 == m:
            l = D(p12, sg(z11)) + D(p21, sg(z22))
        elif d1222 == m:
            l = D(p12, sg(z11)) + D(p22, sg(z21))
        loss.append(0.5 * l)

    loss = sum(loss) / len(loss) # mean
    loss.backward() # back-propagate
    update(f, h) # SGD update
\end{minted}
\label{alg:algorithm}
\end{algorithm}

\section{Data augmentation details}
\label{sec:data_augmentation_details}

\subsection{ImageNet}
\label{subsec:data_augmentation_details_imagenet}

The training set consists of 1,281,167 images, and the validation dataset consists of 50,000 images. Since the label of ImageNet's test set is not provided, we use the validation set for evaluation. The number of classes in ImageNet is 1,000. The images are variable-sized, so they are cropped to size $224 \times 224$ during the data augmentation process. We apply in turn the following data transformations in pre-training.

\begin{itemize}
\item \texttt{RandomResizedCrop}: Crop a random patch of the image with scale $(0.2, 1)$ and then resize it to a size of $(224, 224)$.
\item \texttt{ColorJitter}: With a probability of $0.8$, change the image's brightness, contrast, saturation, and hue with strength $(0.4, 0.4, 0.4, 0.1)$.
\item \texttt{RandomGrayscale}: With a probability of $0.2$, convert the image to grayscale.
\item \texttt{GaussianBlur}: With a probability of $0.5$, apply the Gaussian blur filter to the image with a radius uniformly sampled from $[0.1, 2]$.
\item \texttt{RandomHorizontalFlip}: With a probability of $0.5$, flop the image horizontally.
\item \texttt{Normalize}: Normalize the image with mean $(0.485, 0.456, 0.406)$ and standard deviation $(0.229, 0.224, 0.225)$.
\end{itemize}

\subsection{CIFAR-10}
\label{subsec:data_augmentation_details_cifar-10}

The training set consists of 60,000 images, and the test set consists of 10,000 images. The number of classes in CIFAR-10 is 10. The images have a fixed size of $32 \times 32$. We apply in turn the following data transformations in pre-training.

\begin{itemize}
\item \texttt{RandomResizedCrop}: Crop a random patch of the image with scale $(0.08, 1)$ and then resize it to a size of $(32, 32)$.
\item \texttt{RandomHorizontalFlip}: With a probability of $0.5$, flop the image horizontally.
\item \texttt{ColorJitter}: With a probability of $0.8$, change the image's brightness, contrast, saturation, and hue with strength $(0.4, 0.4, 0.4, 0.1)$.
\item \texttt{RandomGrayscale}: With a probability of $0.2$, convert the image to grayscale.
\item \texttt{Normalize}: Normalize the image with mean $(0.485, 0.456, 0.406)$ and standard deviation $(0.229, 0.224, 0.225)$.
\end{itemize}

\section{Pre-training implementation details}
\label{sec:pre-training_implementation_details}

\subsection{ImageNet}
\label{subsec:pre-training_implementation_details_imagenet}

For the ImageNet experiments, we follow the experimental setup in SimSiam's original paper \citep{chen2021exploring} and \url{https://github.com/facebookresearch/simsiam}. The encoder $f$ consists of a backbone and a projector. The backbone is ResNet-50 \citep{he2016deep} with about 23M parameters, and the projector consists of three fully-connected (FC) layers. Batch normalization (BN) and ReLU are applied to each layer except for the output layer, where only BN is used. The input and output dimensions of the layers are 2,048. The predictor $h$ consists of two FC layers. BN and ReLU are applied to the first layer, and nothing is applied to the second layer. The input and output dimensions of the first layer are 2,048 and 512, and the input and output dimensions of the second layer are 512 and 2,048 (bottleneck structure).

\subsection{CIFAR-10}
\label{subsec:pre-training_implementation_details_cifar-10}

For the CIFAR-10 experiments, we follow the default setup in \citep{susmelj2020lightly} and \url{https://github.com/lightly-ai/lightly}. The encoder $f$ consists of a backbone and a projector. The backbone is a variant of ResNet-18 for CIFAR-10 with about 11M parameters, and the projector consists of two FC layers. BN and ReLU are applied to the first layer, and only BN is applied to the second layer. The input and output dimensions of the first layer are 512 and 2,048, and the input and output dimensions of the second layer are 2,048. The predictor $h$ is the same as in the case of ImageNet.

\section{Transfer learning implementation details}
\label{sec:transfer_learning_implementation_details}

We present the implementation details for transfer learning. Other details can be found in \citep{ericsson2021well} and \url{https://github.com/linusericsson/ssl-transfer}. We minimally change the code under the same computational parameters.

\subsection{Image recognition}
\label{subsec:transfer_learning_implementation_details_image_recognition}

In the Caltech dataset, 30 images for each class are randomly selected to form the training set, and the remaining images form the test set. On the DTD and SUN397 datasets, the first dataset split provided by the authors is used. For validation sets, the pre-defined 
ones are used on the Aircraft, DTD, Flowers, and VOC2007 datasets, and on the rest of the datasets, $20\%$ of the training set are randomly selected to form the validation set. 

We resize the images so that the number of pixels of the shorter side of each image is $224$ using bicubic interpolation. We crop the images at the center with size $(224, 224)$. The best weight decay rate is selected on the validation set over $45$ evenly spaced values from $10^{-6}$ to $10^5$ on a logarithmic scale with base $10$. We find the best hyperparameters on the validation set, retrain the networks on the training and validation sets, and evaluate on the test set. Since VOC2007 is a multi-label dataset, we fit a binary classifier for each class. L-BFGS \citep{liu1989limited} is used as a solver when doing a logistic regression.

\subsection{Object detection}
\label{subsec:transfer_learning_implementation_details_object_detection}
We initially normalize the images with mean $(123.675, 116.280, 103.530)$ and standard deviation $(58.395, 57.120, 57.375)$.

\textbf{Pascal-VOC} The images are resized so that the number of pixels of the shorter side of each image is one of $\{480, 512, 544, 576, 608, 640, 672, 704, 736, 768, 800\}$ for training and $800$ for test.  We use the batch size of $2$. The base learning rate is chosen as $0.0025$, and it decays by tenths at the $96,000$th and $128,000$th iterations, respectively. We warm up the networks for $100$ iterations and train them for $144,000$ iterations.

\textbf{MS-COCO} The images are resized so that the number of pixels of the shorter side of each image is one of $\{640, 672, 704, 736, 768, 800\}$ for training and $800$ for test. We use the batch size of $16$. We choose the base learning rate as $0.02$, and it decays by tenths at the $210,000$th and $250,000$th iterations, respectively. We warm up the networks for $1000$ iterations and train them for $270,000$ iterations. 

\subsection{Semantic segmentation}
\label{subsec:semantic_segmentation}

We use the SGD optimizer with momentum of $0.9$, learning rate of $0.02$, weight decay rate of $0.0001$, and learning rate decay rate of $0.9$. We use the batch size of $2$. The networks are trained for $150,000$ iterations in total ($30$ epochs and $5,000$ iterations per epoch). The learning rate decays every 500 iterations.

\section{t-distributed stochastic neighbor embedding (t-SNE)}
\label{sec:tsne}

\begin{figure*}[t!]
\vskip 0.2in
    \centering
    \begin{subfigure}[b]{0.49\linewidth}
        \centering
        \includegraphics[width=\linewidth]{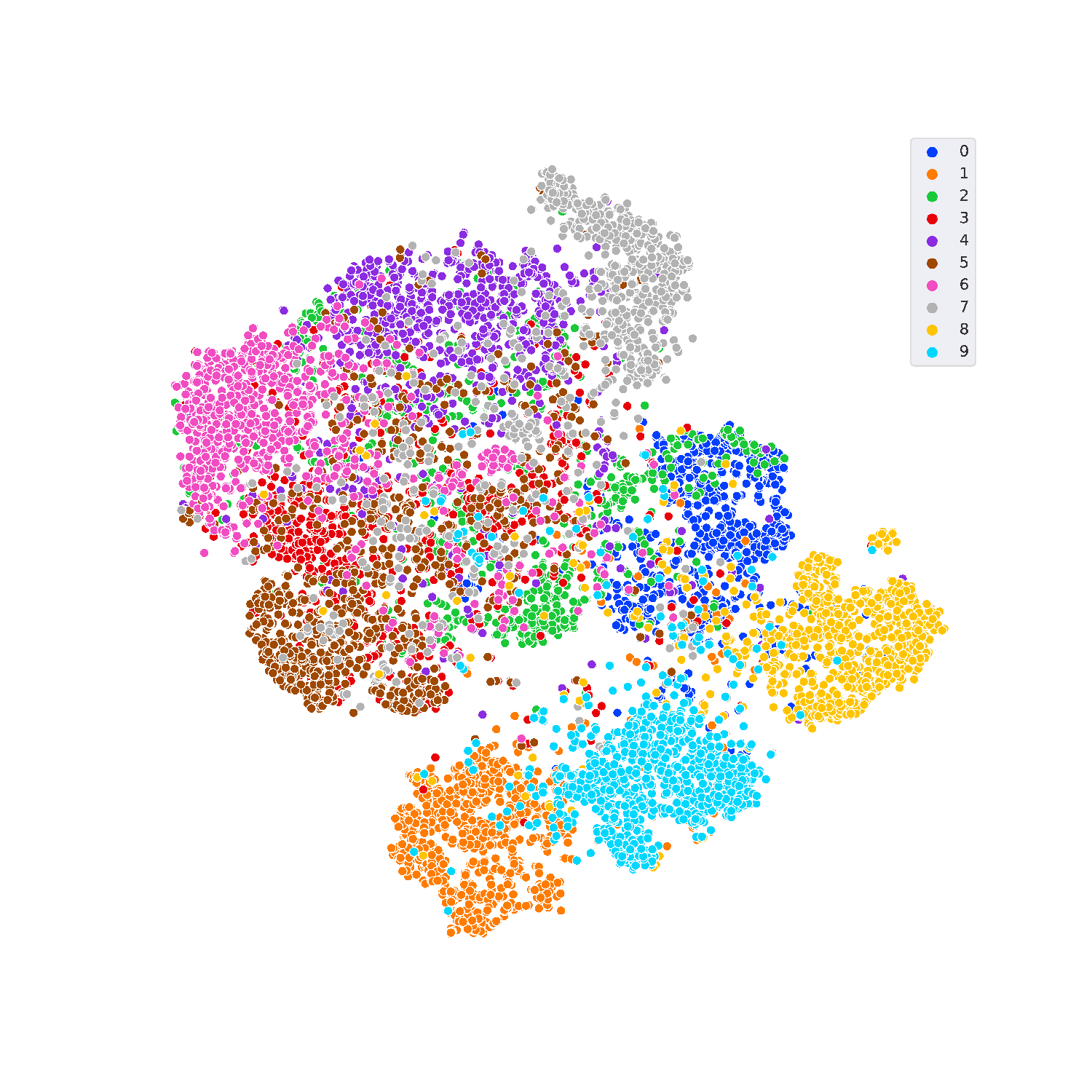}
        \caption{SimSiam}    
        \label{subfig:tsne_simsiam}
    \end{subfigure}
    \begin{subfigure}[b]{0.49\linewidth}  
        \centering 
        \includegraphics[width=\linewidth]{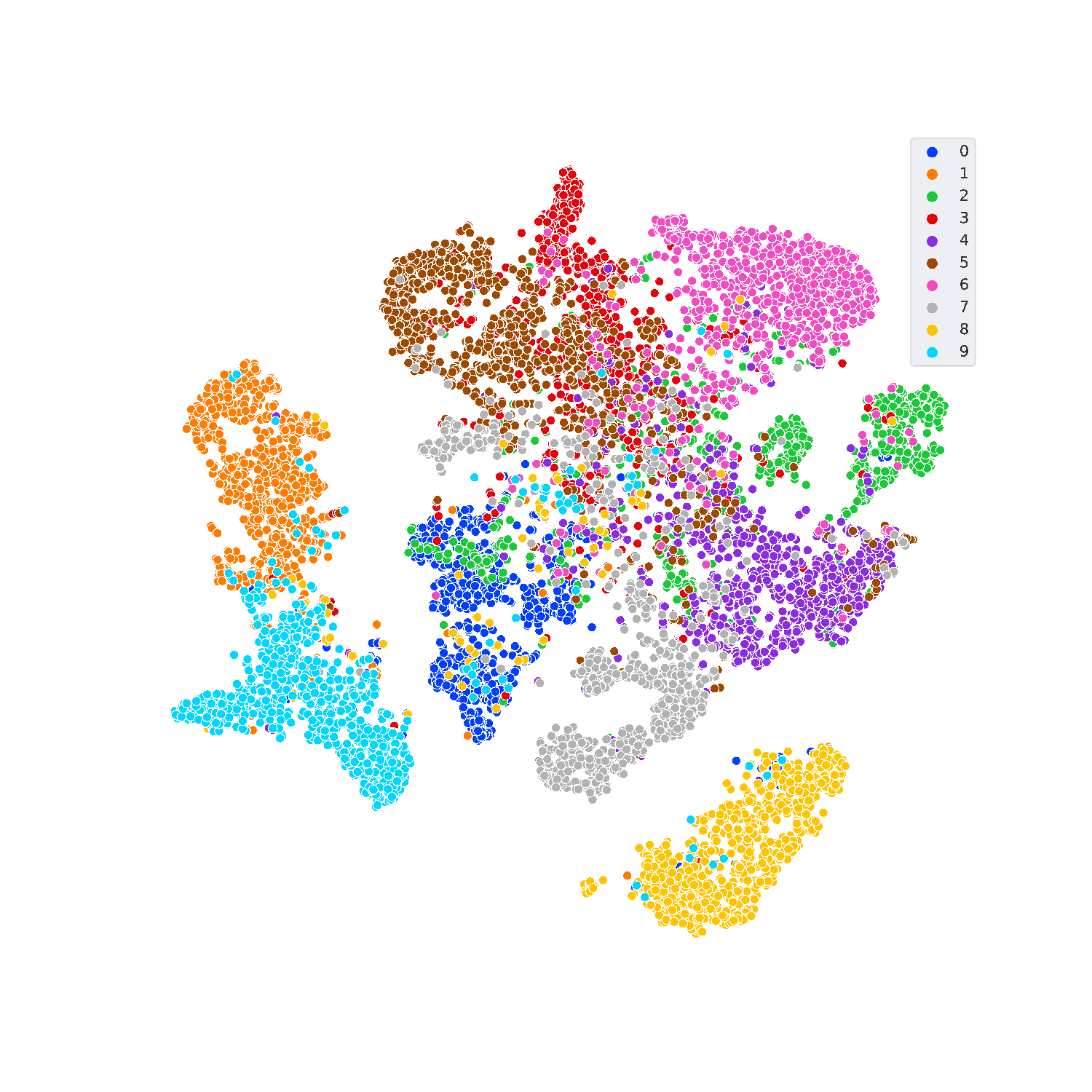}
        \caption{SimSiam w/ GSG}    
        \label{subfig:tsne_simsiam_gsg}
    \end{subfigure}
    \begin{subfigure}[b]{0.49\linewidth}   
        \centering 
        \includegraphics[width=\linewidth]{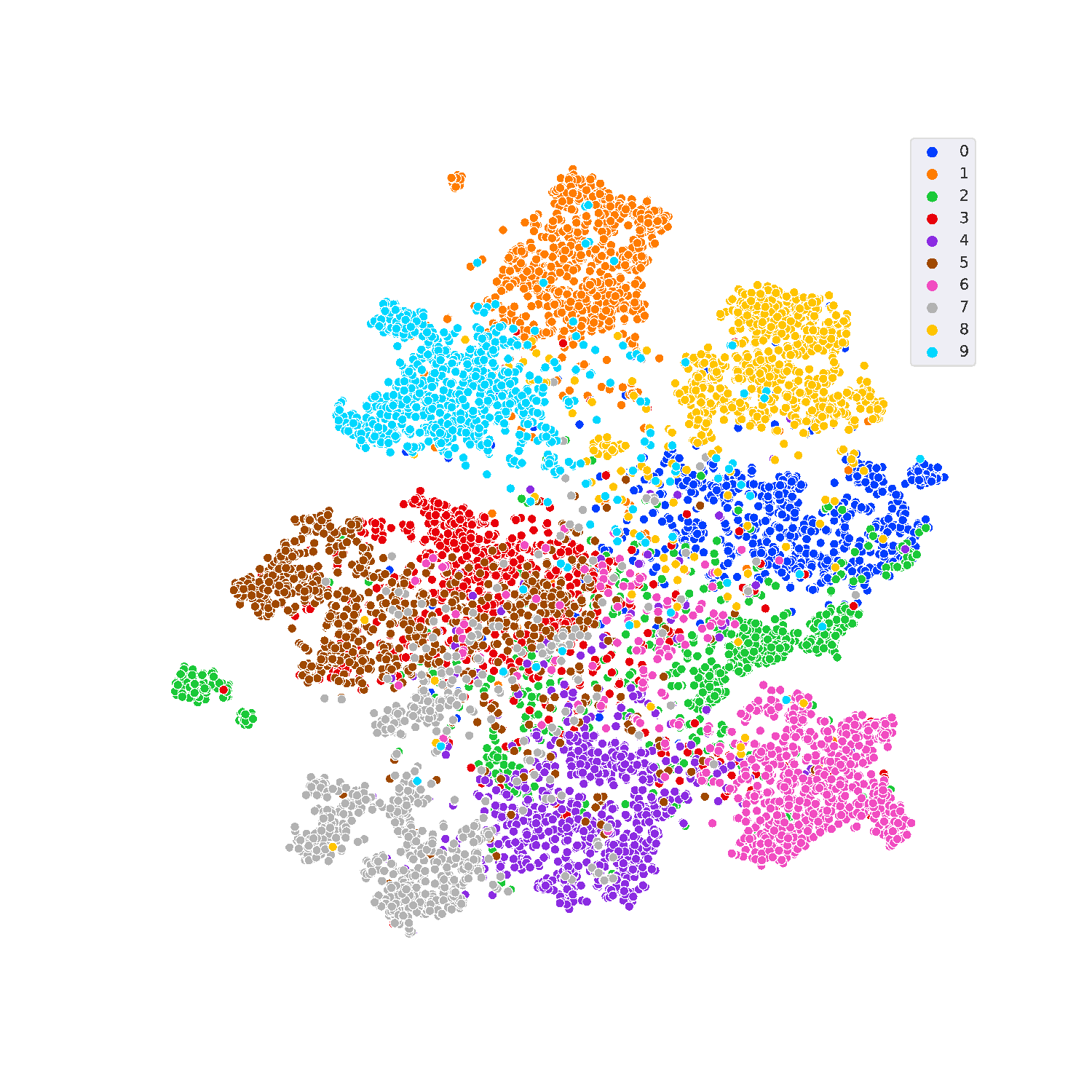}
        \caption{BYOL}   
        \label{subfig:tsne_byol}
    \end{subfigure}
    \begin{subfigure}[b]{0.49\linewidth}   
        \centering 
        \includegraphics[width=\linewidth]{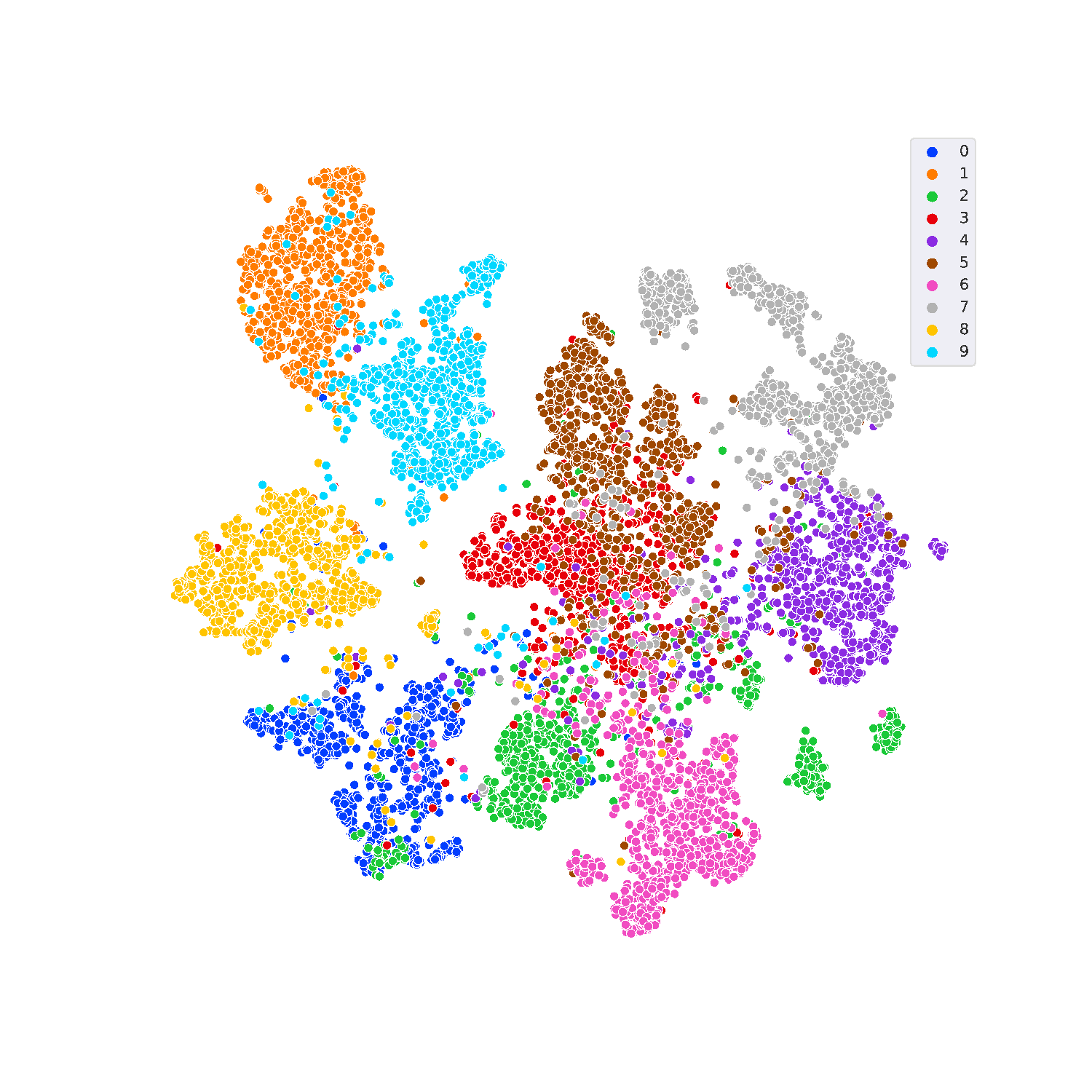}
        \caption{BYOL w/ GSG}    
        \label{subfig:tsne_byol_gsg}
    \end{subfigure}
    \caption{t-SNE visualizations of CIFAR-10. Clusters of our algorithms are better separated than those of the original algorithms as expected.} 
    \label{fig:tsne}
\vskip -0.2in
\end{figure*}

We present t-SNE visualizations \citep{van2008visualizing} of representations of images from the test set of CIFAR-10. Figure \ref{fig:tsne} shows that in both SimSiam and BYOL, applying our GSG made the clusters corresponding to the classes better separate.

\section{Semantic segmentation}
\label{sec:semantic_segmentation}

\begin{figure}
     \centering
     \begin{subfigure}[b]{0.157\linewidth}
         \centering
         \caption{Test image}
         \includegraphics[width=\linewidth]{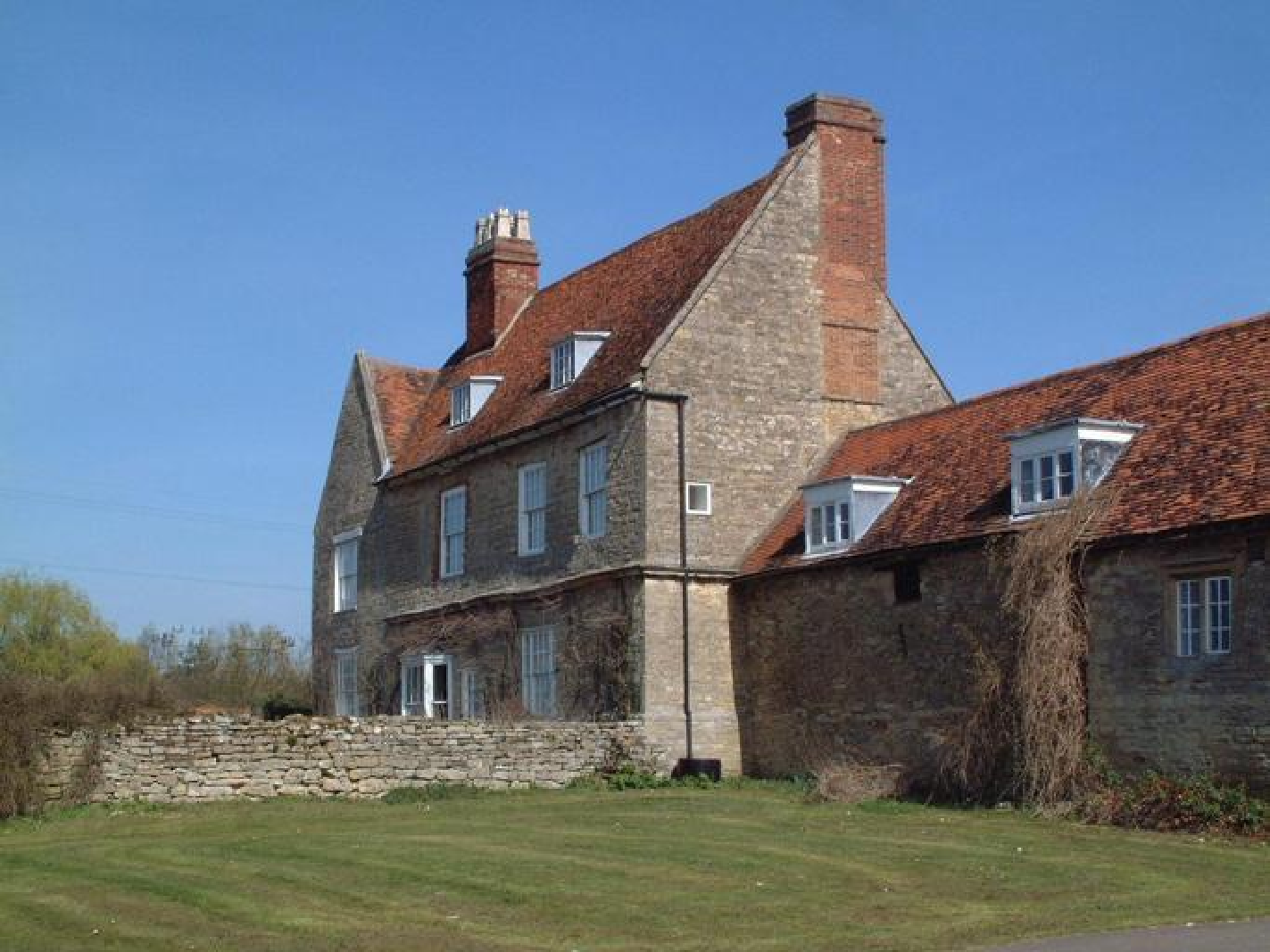}
     \end{subfigure}
     \hspace{-0.4em}
     \begin{subfigure}[b]{0.157\linewidth}
         \centering
         \caption{Ground truth}
         \includegraphics[width=\linewidth]{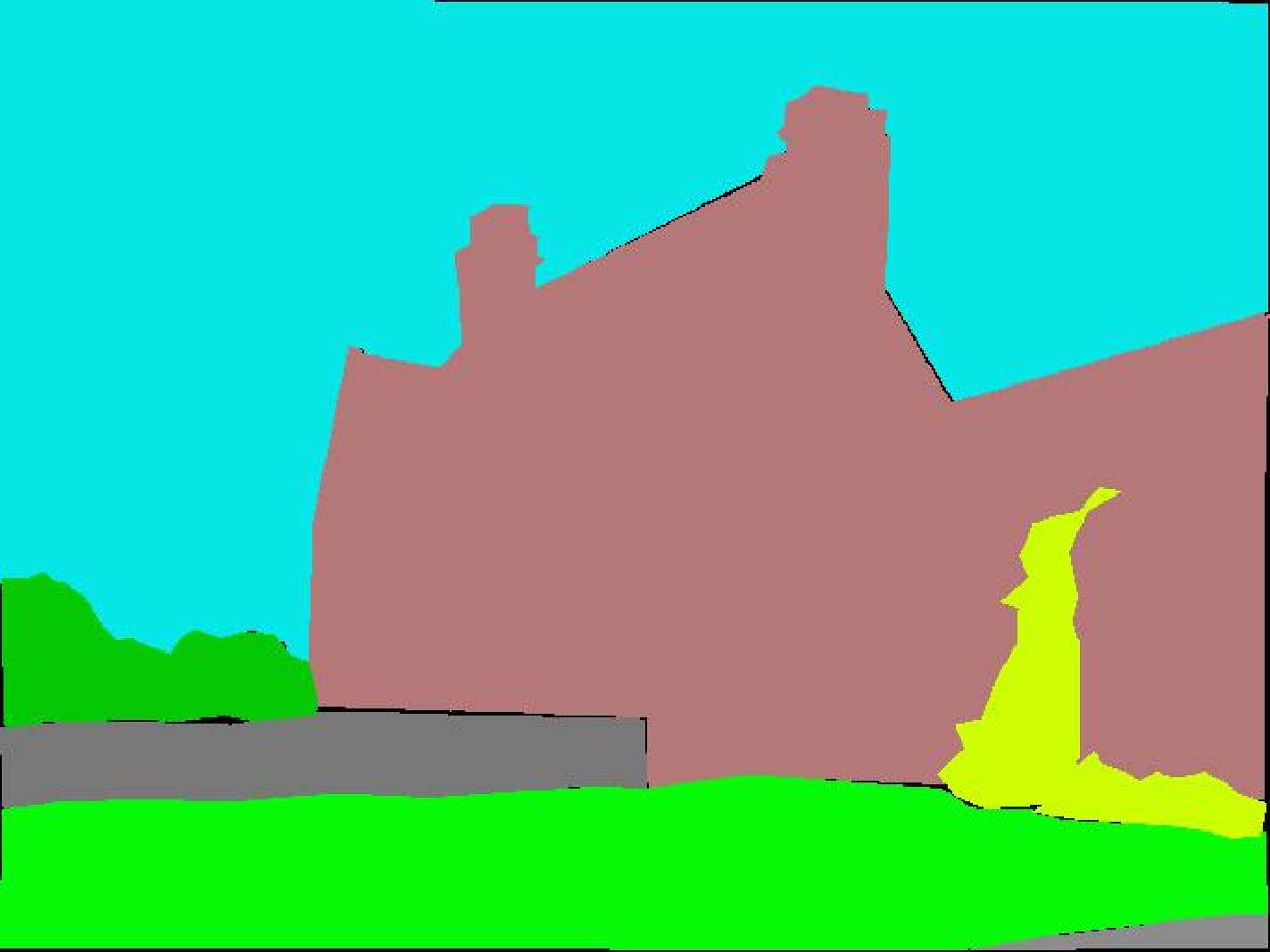}
     \end{subfigure}
     \hfill
     \begin{subfigure}[b]{0.157\linewidth}
         \centering
         \caption{SimSiam}
         \includegraphics[width=\linewidth]{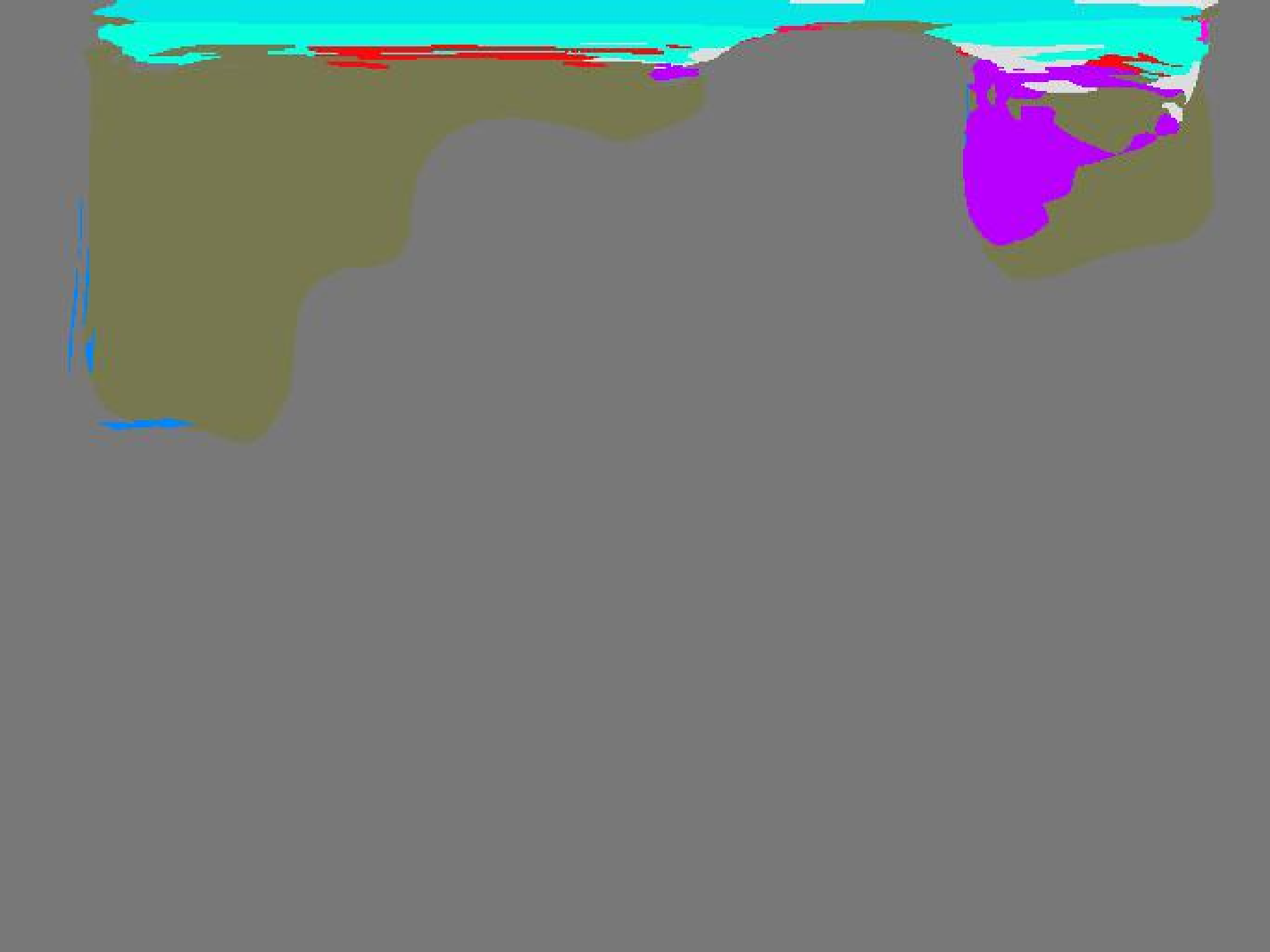}
     \end{subfigure}
     \hspace{-0.4em}
     \begin{subfigure}[b]{0.157\linewidth}
         \centering
         \caption{SimSiamGSG}
         \includegraphics[width=\linewidth]{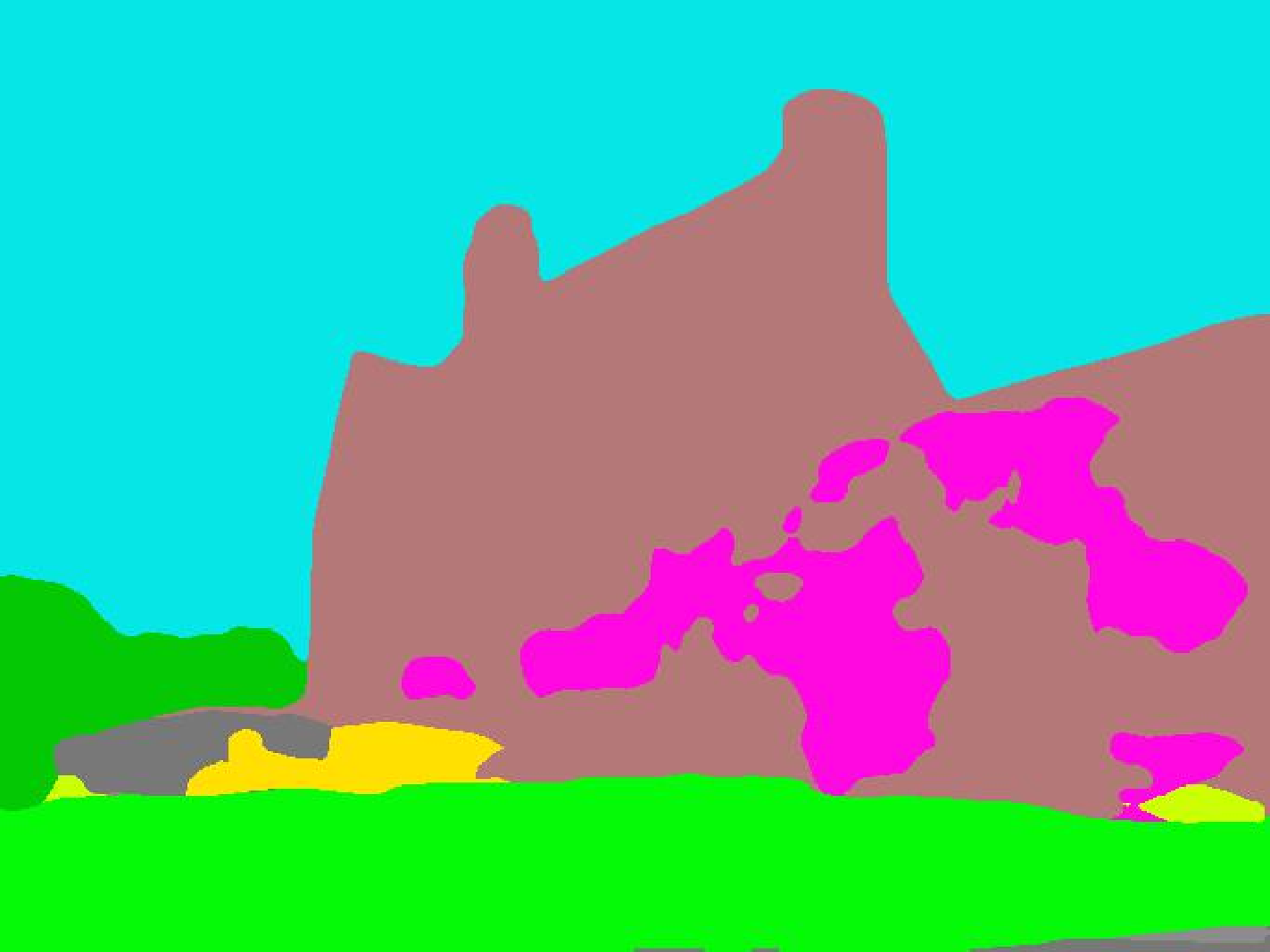}
     \end{subfigure}
     \hfill
     \begin{subfigure}[b]{0.157\linewidth}
         \centering
         \caption{BYOL}
         \includegraphics[width=\linewidth]{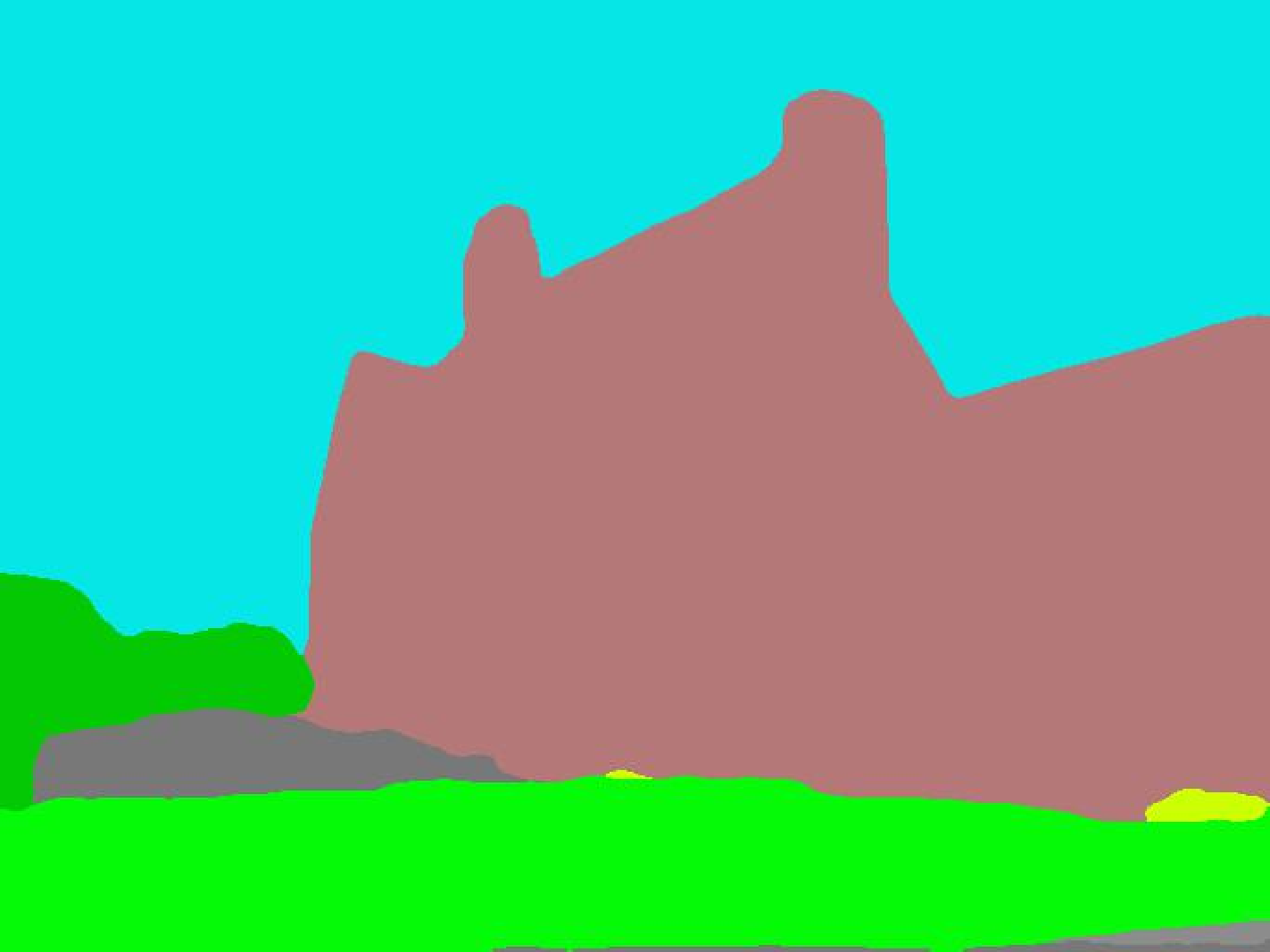}
     \end{subfigure}
     \hspace{-0.4em}
     \begin{subfigure}[b]{0.157\linewidth}
         \centering
         \caption{BYOLGSG}
         \includegraphics[width=\linewidth]{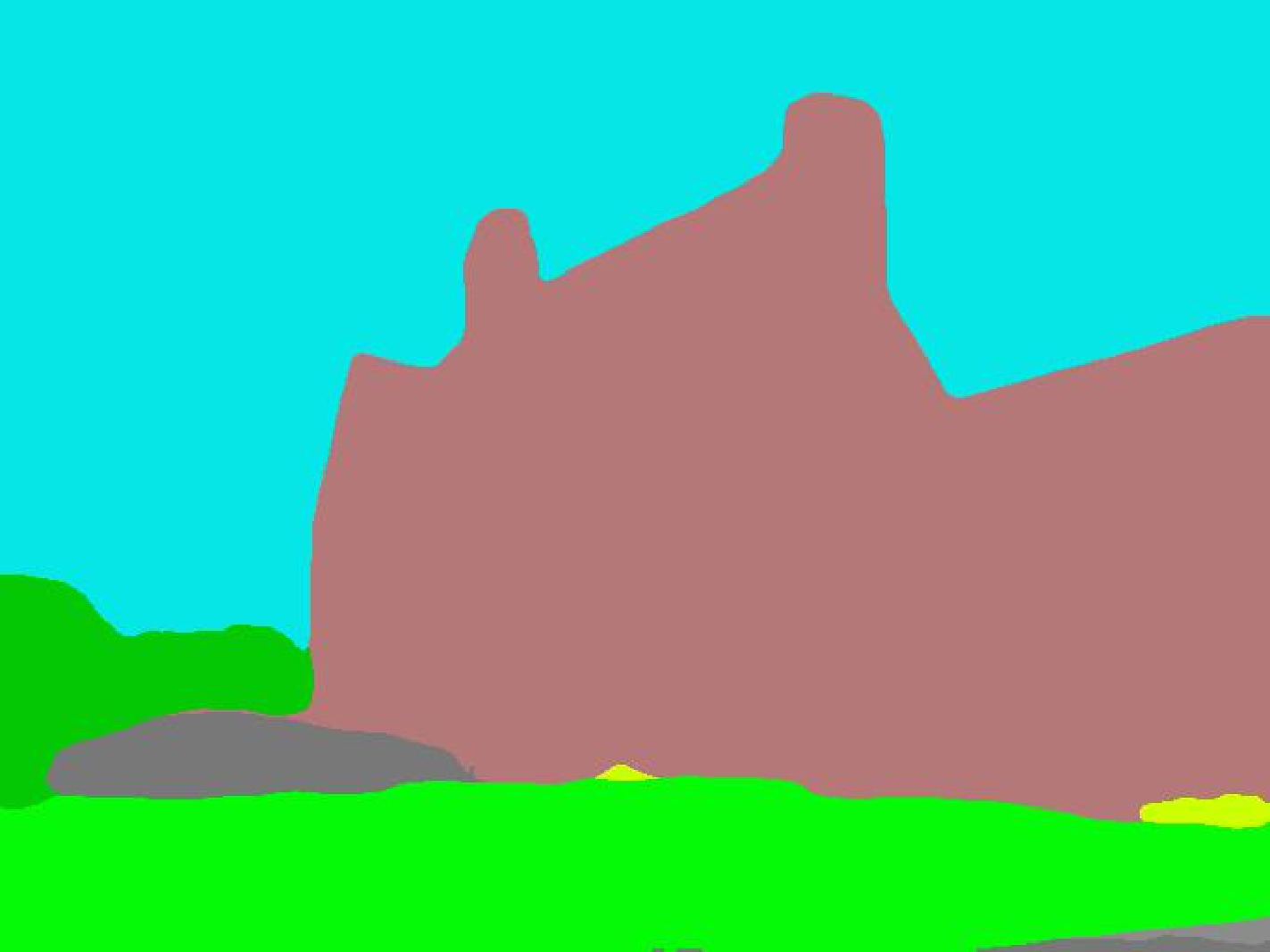}
     \end{subfigure}
     \vspace{1em}
     
     \begin{subfigure}[b]{0.157\linewidth}
         \centering
         \includegraphics[width=\linewidth]{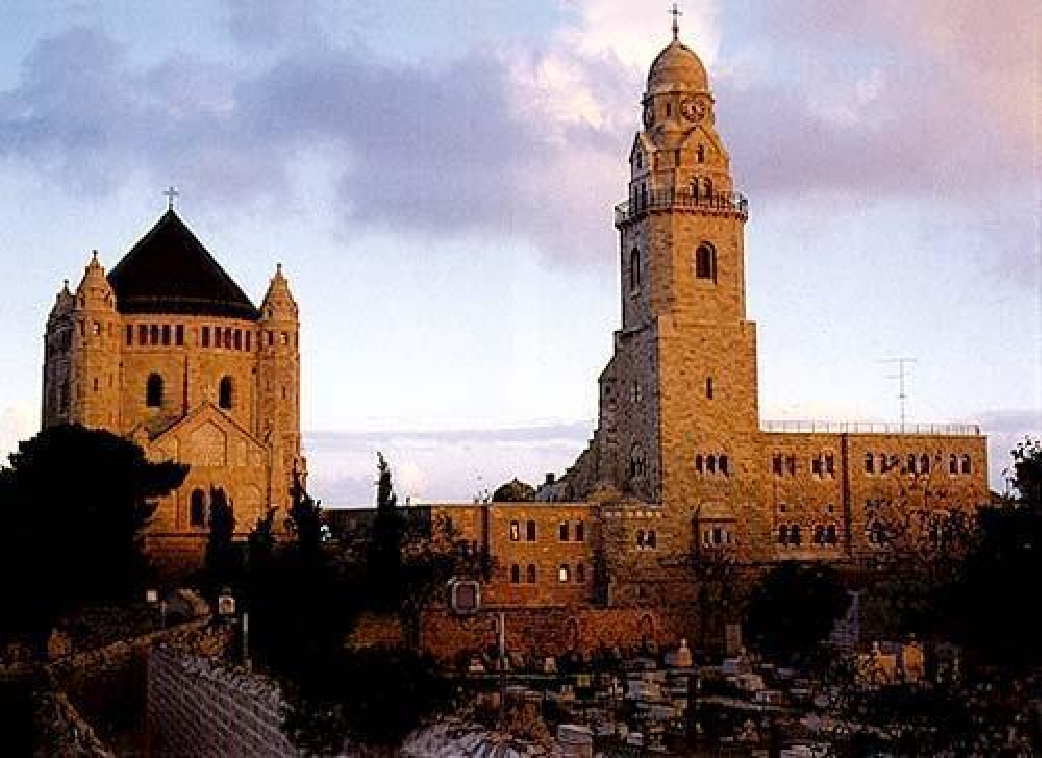}
     \end{subfigure}
     \hspace{-0.4em}
     \begin{subfigure}[b]{0.157\linewidth}
         \centering
         \includegraphics[width=\linewidth]{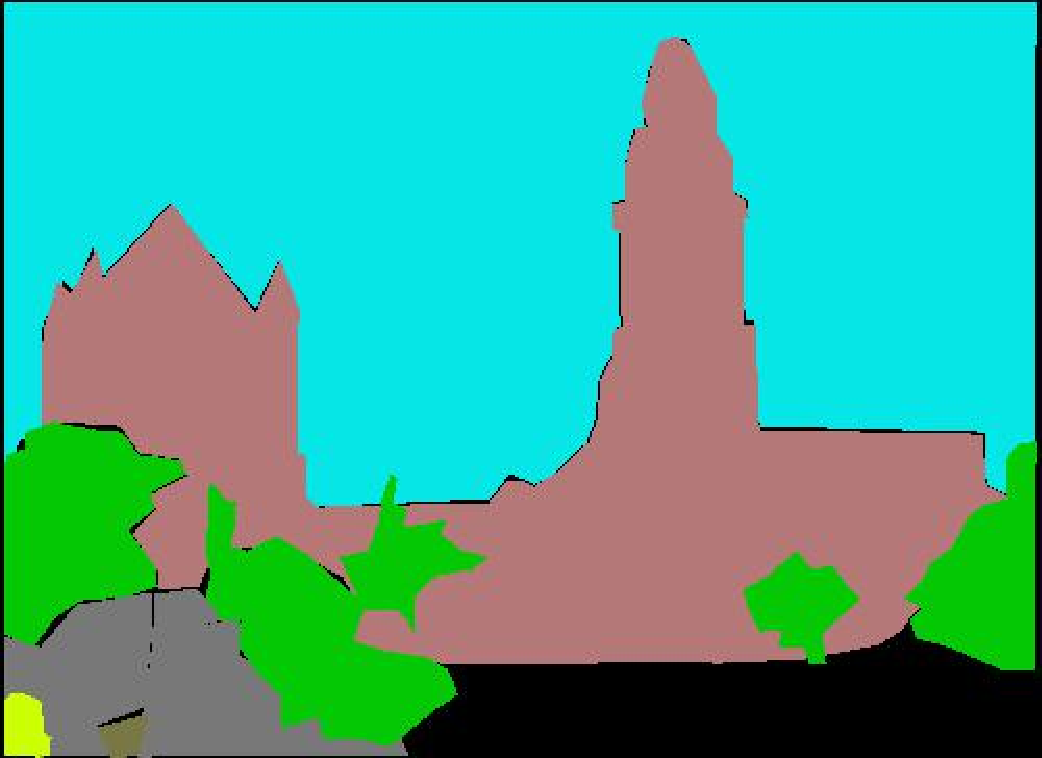}
     \end{subfigure}
     \hfill
     \begin{subfigure}[b]{0.157\linewidth}
         \centering
         \includegraphics[width=\linewidth]{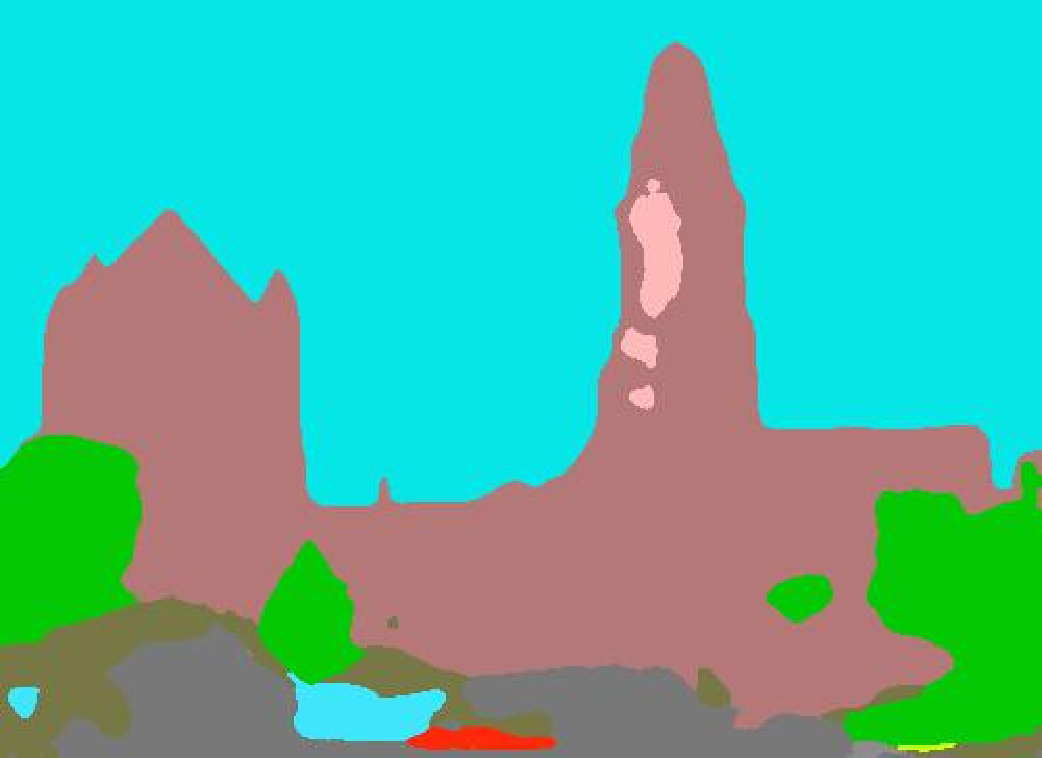}
     \end{subfigure}
     \hspace{-0.4em}
     \begin{subfigure}[b]{0.157\linewidth}
         \centering
         \includegraphics[width=\linewidth]{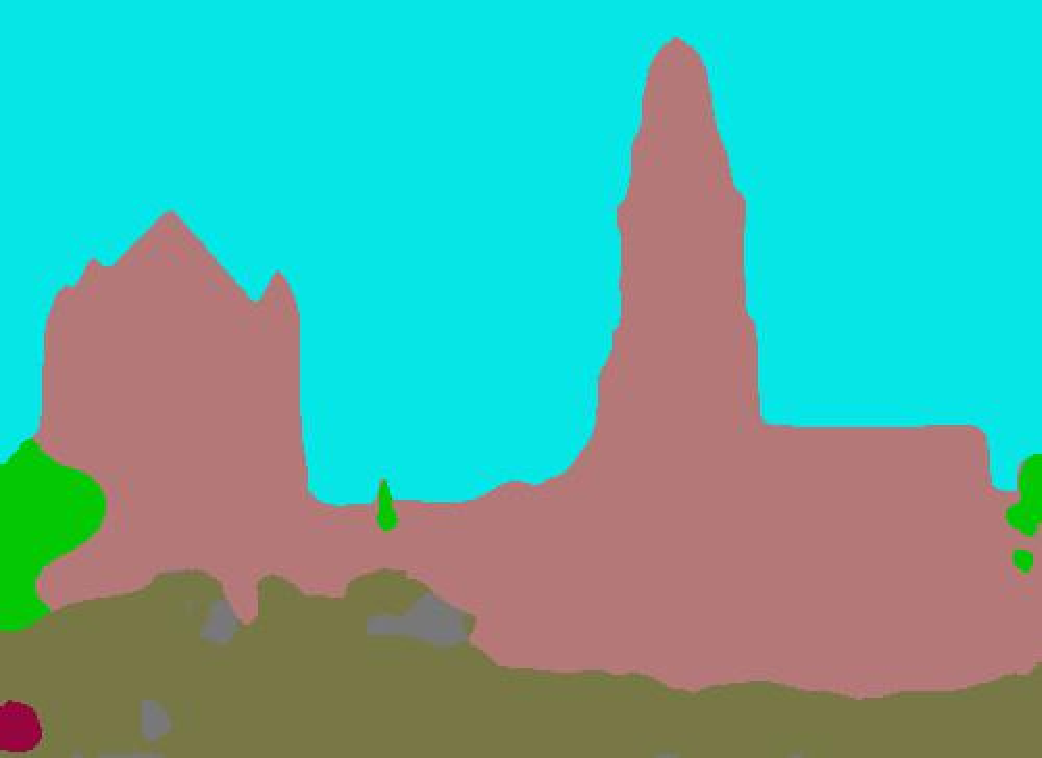}
     \end{subfigure}
     \hfill
     \begin{subfigure}[b]{0.157\linewidth}
         \centering
         \includegraphics[width=\linewidth]{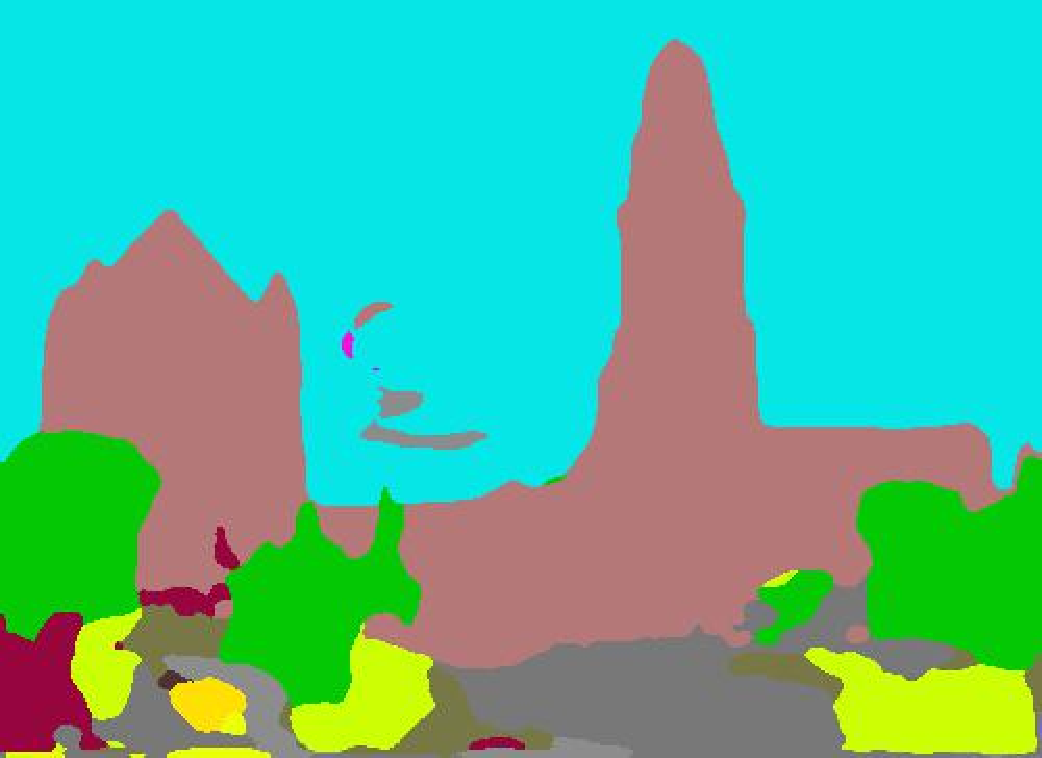}
     \end{subfigure}
     \hspace{-0.4em}
     \begin{subfigure}[b]{0.157\linewidth}
         \centering
         \includegraphics[width=\linewidth]{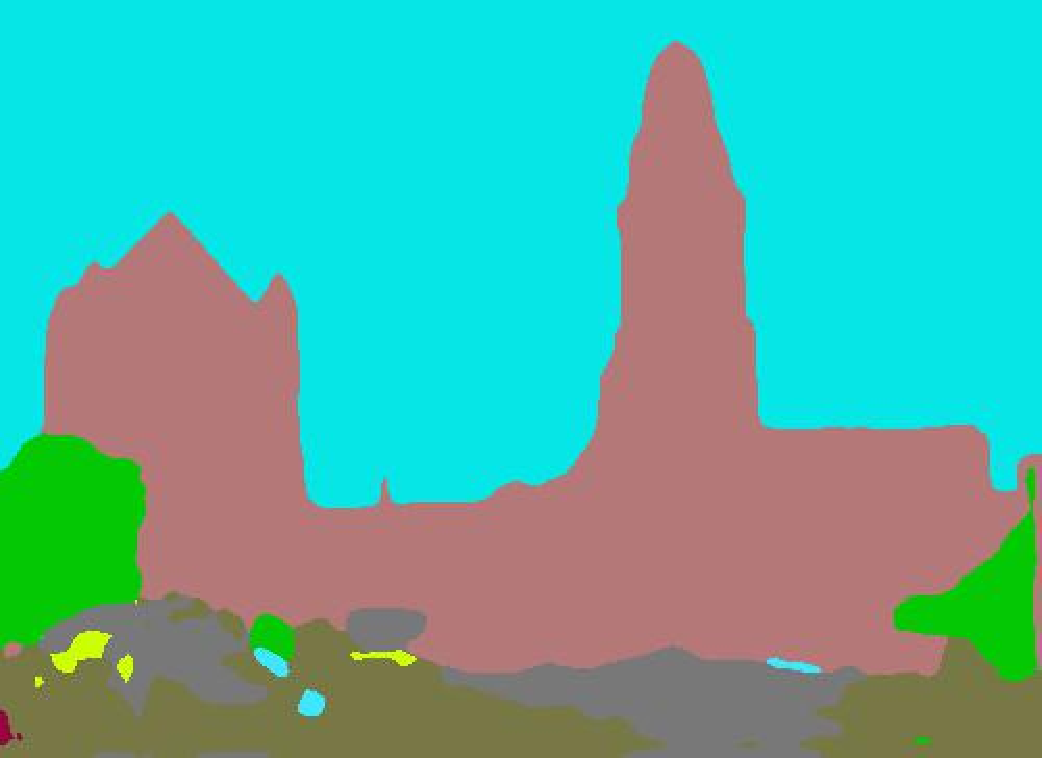}
     \end{subfigure}
     \vspace{1em}
     
     \begin{subfigure}[b]{0.157\linewidth}
         \centering
         \includegraphics[width=\linewidth]{figures/semantic-segmentation/img/ADE_val_00000003_img.eps}
     \end{subfigure}
     \hspace{-0.4em}
     \begin{subfigure}[b]{0.157\linewidth}
         \centering
         \includegraphics[width=\linewidth]{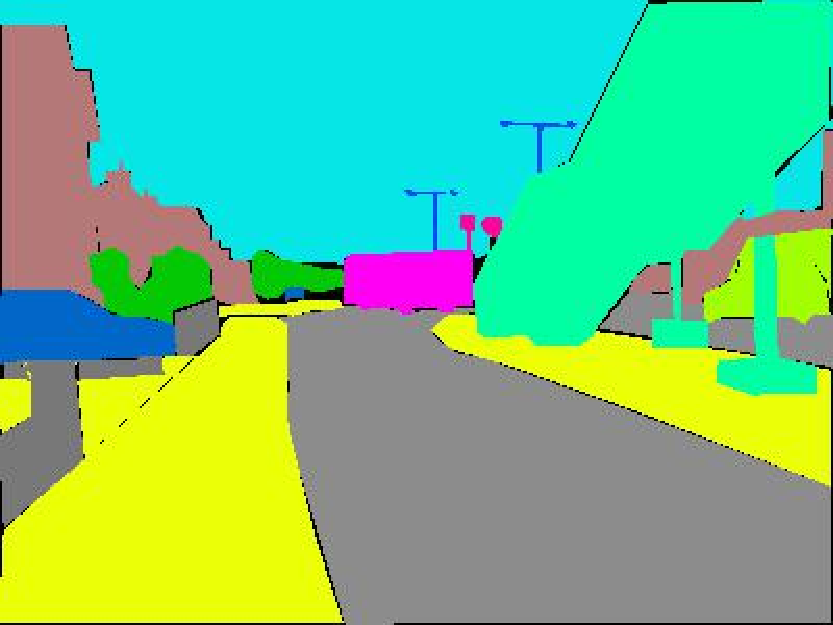}
     \end{subfigure}
     \hfill
     \begin{subfigure}[b]{0.157\linewidth}
         \centering
         \includegraphics[width=\linewidth]{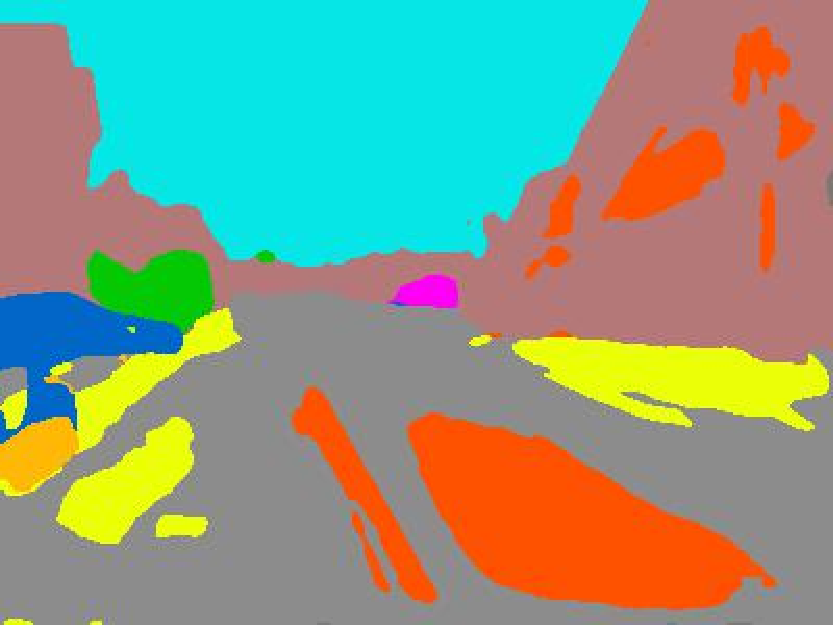}
     \end{subfigure}
     \hspace{-0.4em}
     \begin{subfigure}[b]{0.157\linewidth}
         \centering
         \includegraphics[width=\linewidth]{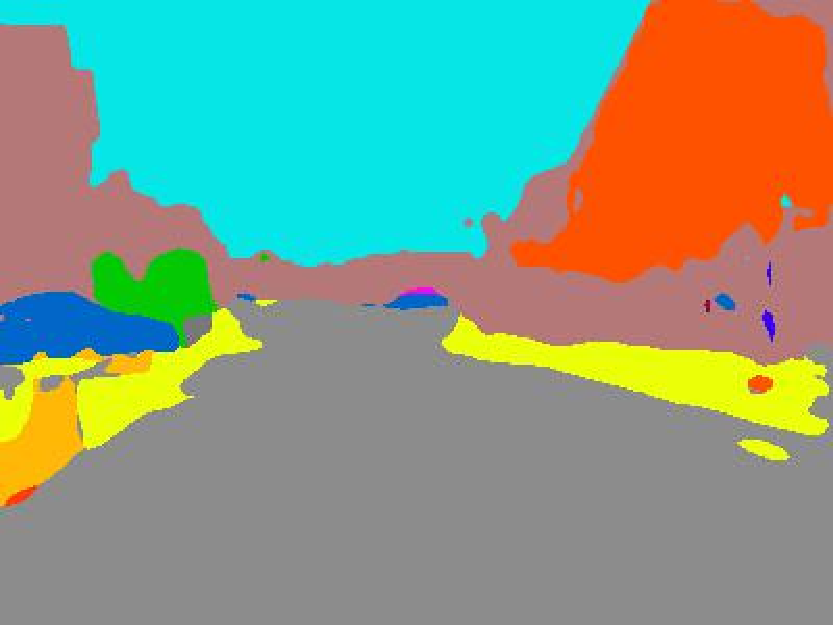}
     \end{subfigure}
     \hfill
     \begin{subfigure}[b]{0.157\linewidth}
         \centering
         \includegraphics[width=\linewidth]{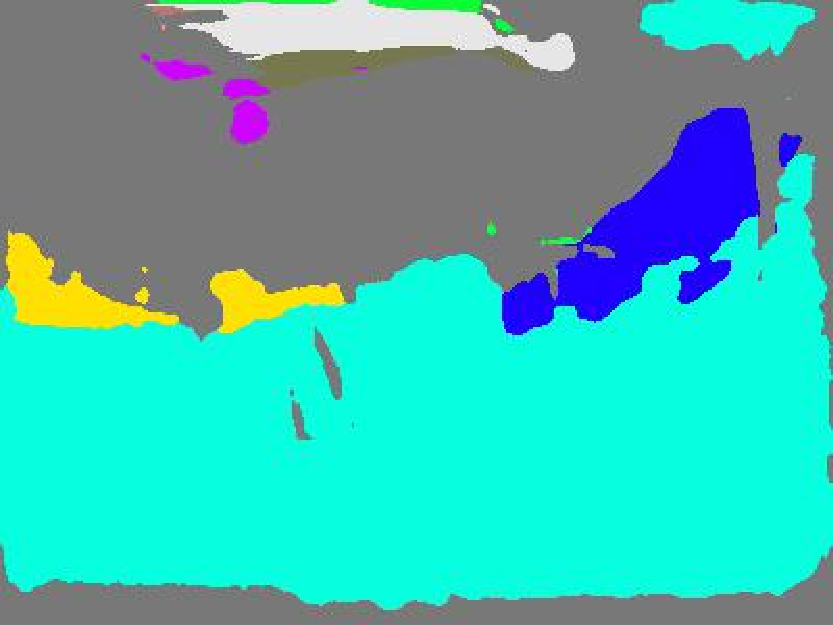}
     \end{subfigure}
     \hspace{-0.4em}
     \begin{subfigure}[b]{0.157\linewidth}
         \centering
         \includegraphics[width=\linewidth]{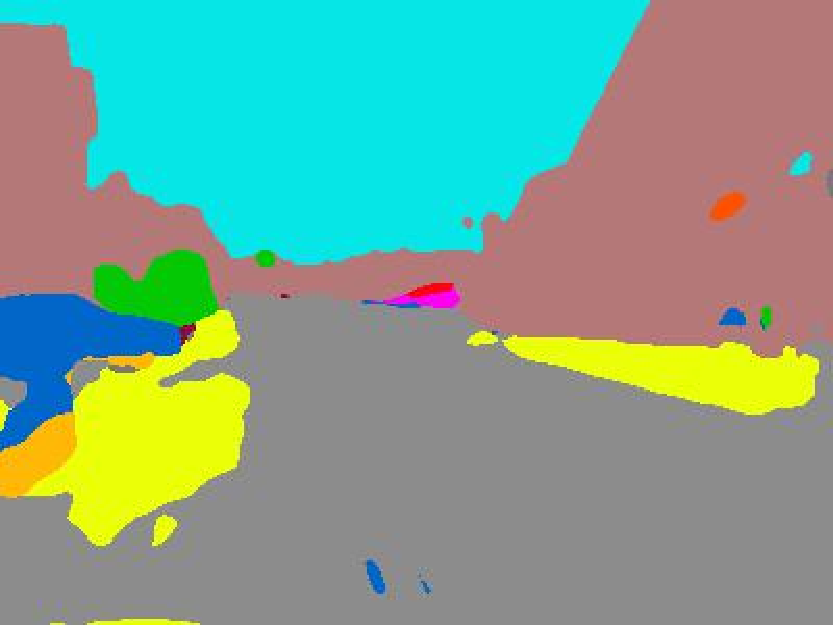}
     \end{subfigure}
     \vspace{1em}
     
     \begin{subfigure}[b]{0.157\linewidth}
         \centering
         \includegraphics[width=\linewidth]{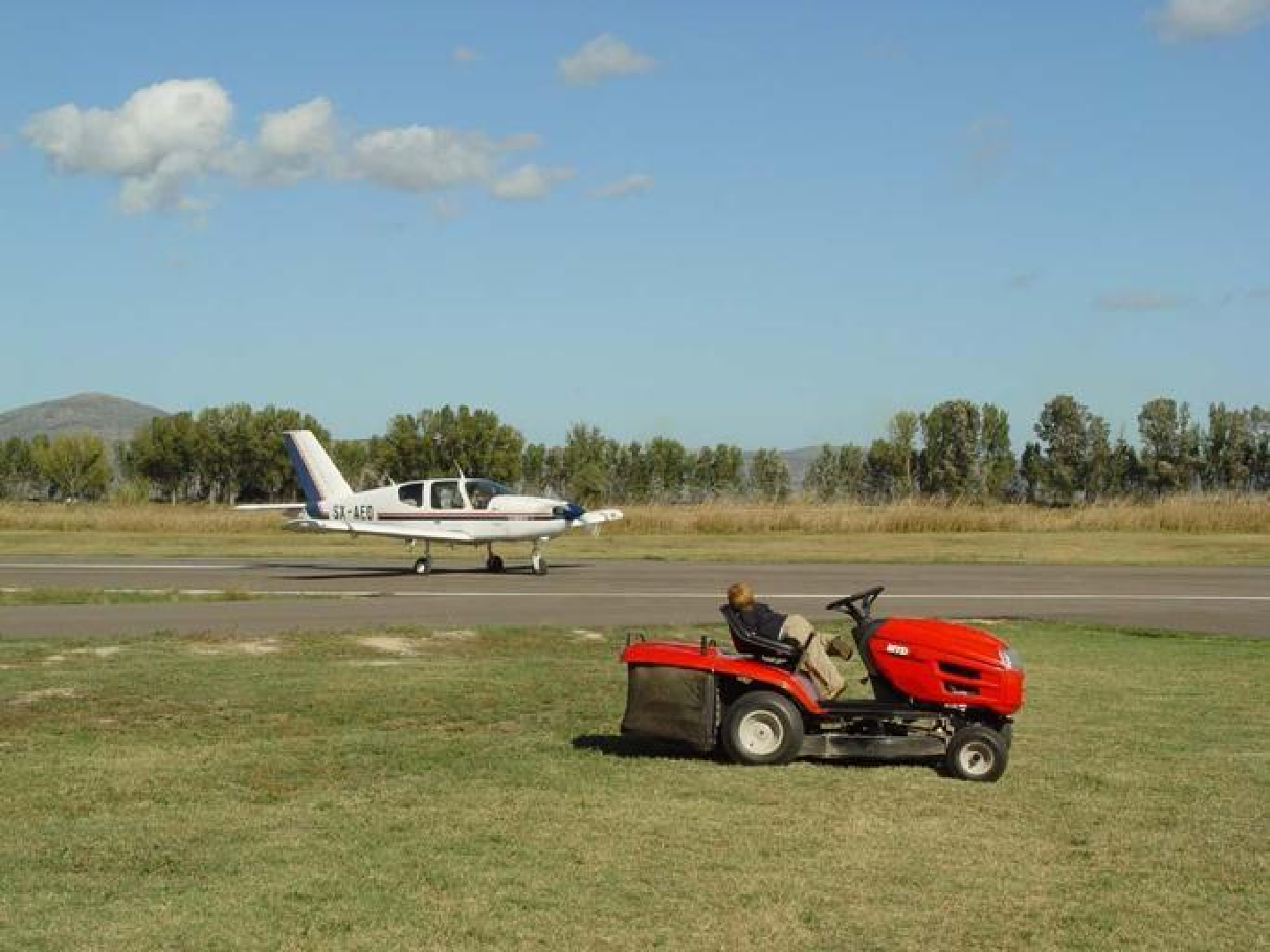}
     \end{subfigure}
     \hspace{-0.4em}
     \begin{subfigure}[b]{0.157\linewidth}
         \centering
         \includegraphics[width=\linewidth]{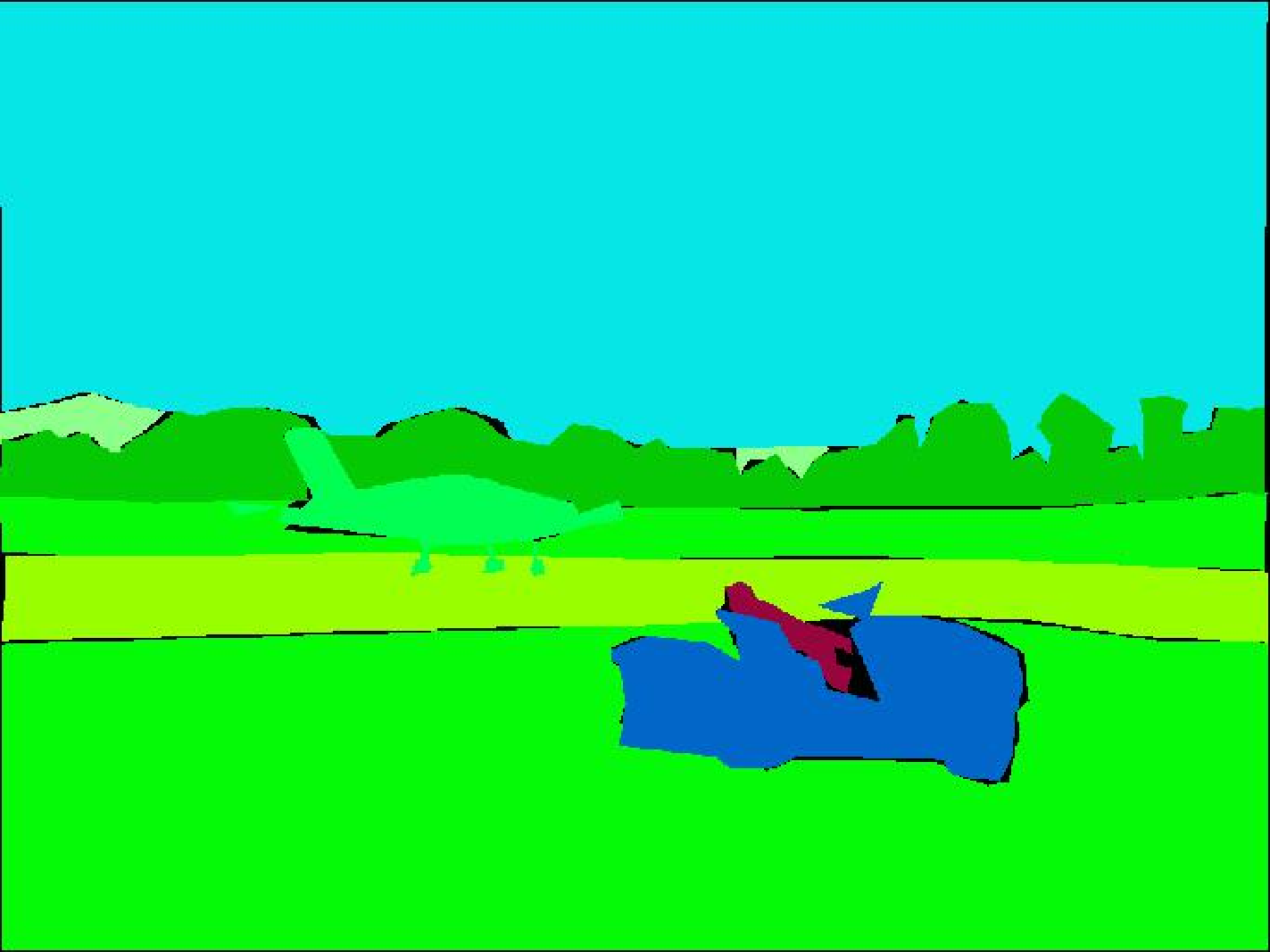}
     \end{subfigure}
     \hfill
     \begin{subfigure}[b]{0.157\linewidth}
         \centering
         \includegraphics[width=\linewidth]{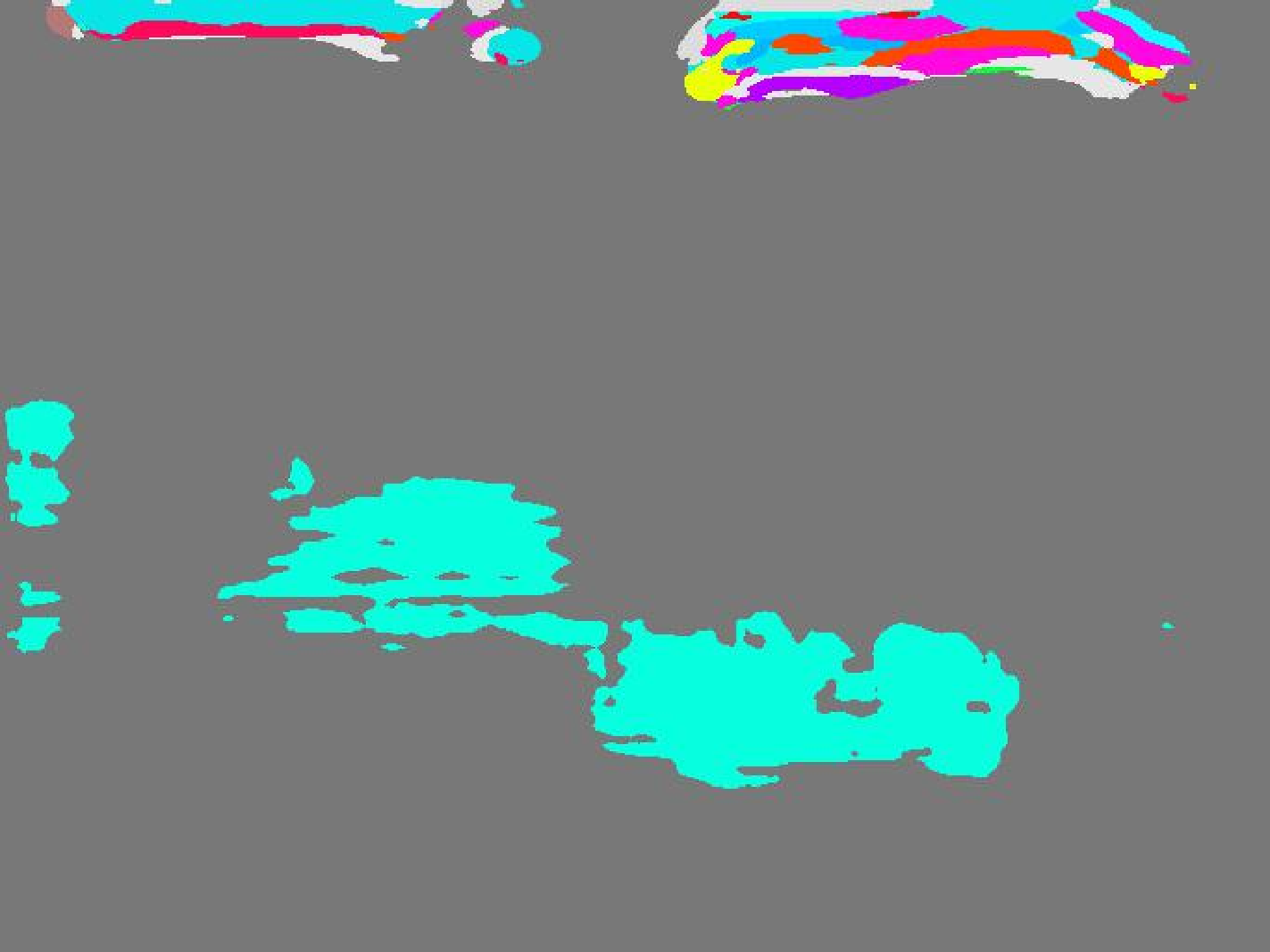}
     \end{subfigure}
     \hspace{-0.4em}
     \begin{subfigure}[b]{0.157\linewidth}
         \centering
         \includegraphics[width=\linewidth]{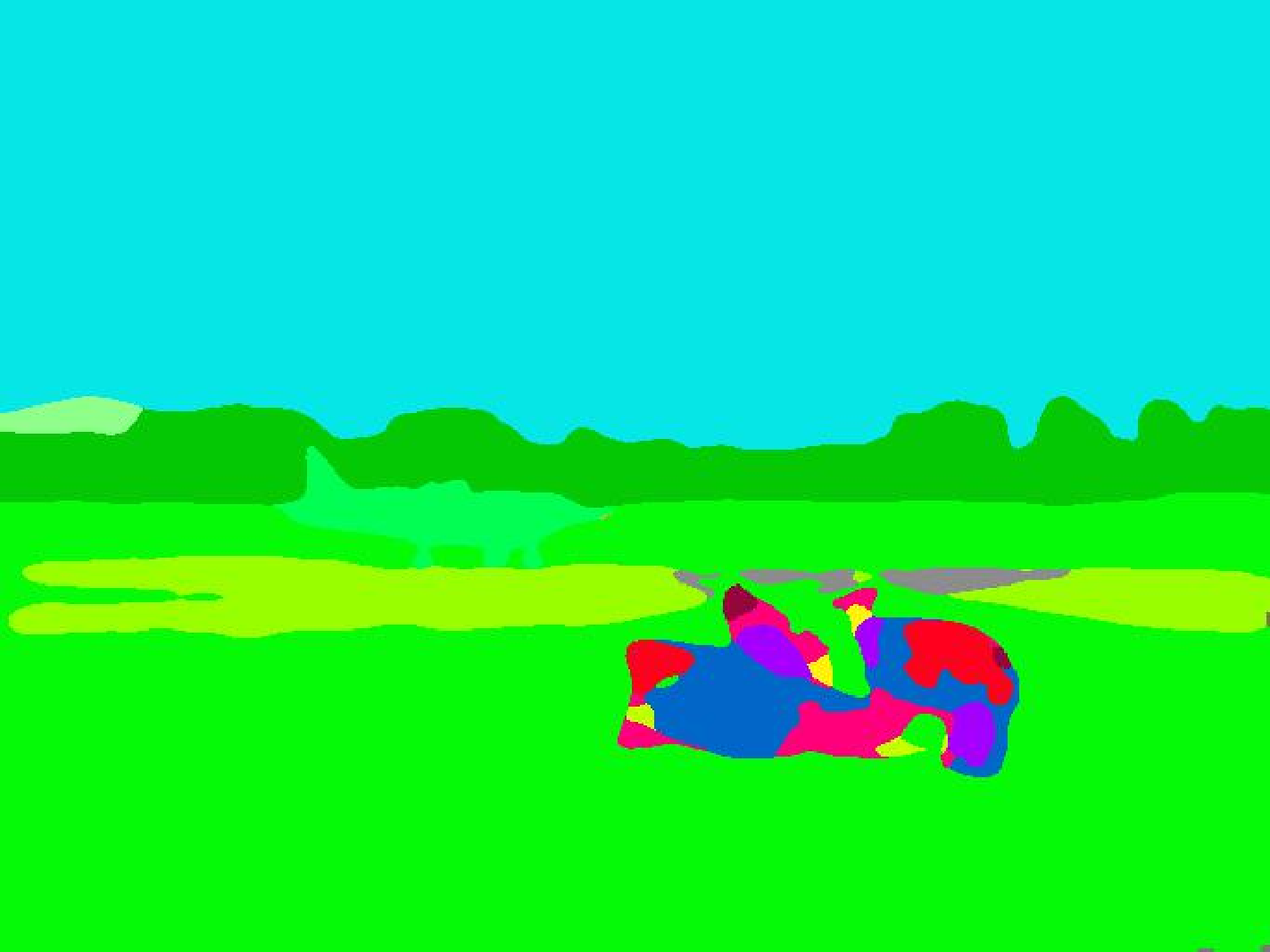}
     \end{subfigure}
     \hfill
     \begin{subfigure}[b]{0.157\linewidth}
         \centering
         \includegraphics[width=\linewidth]{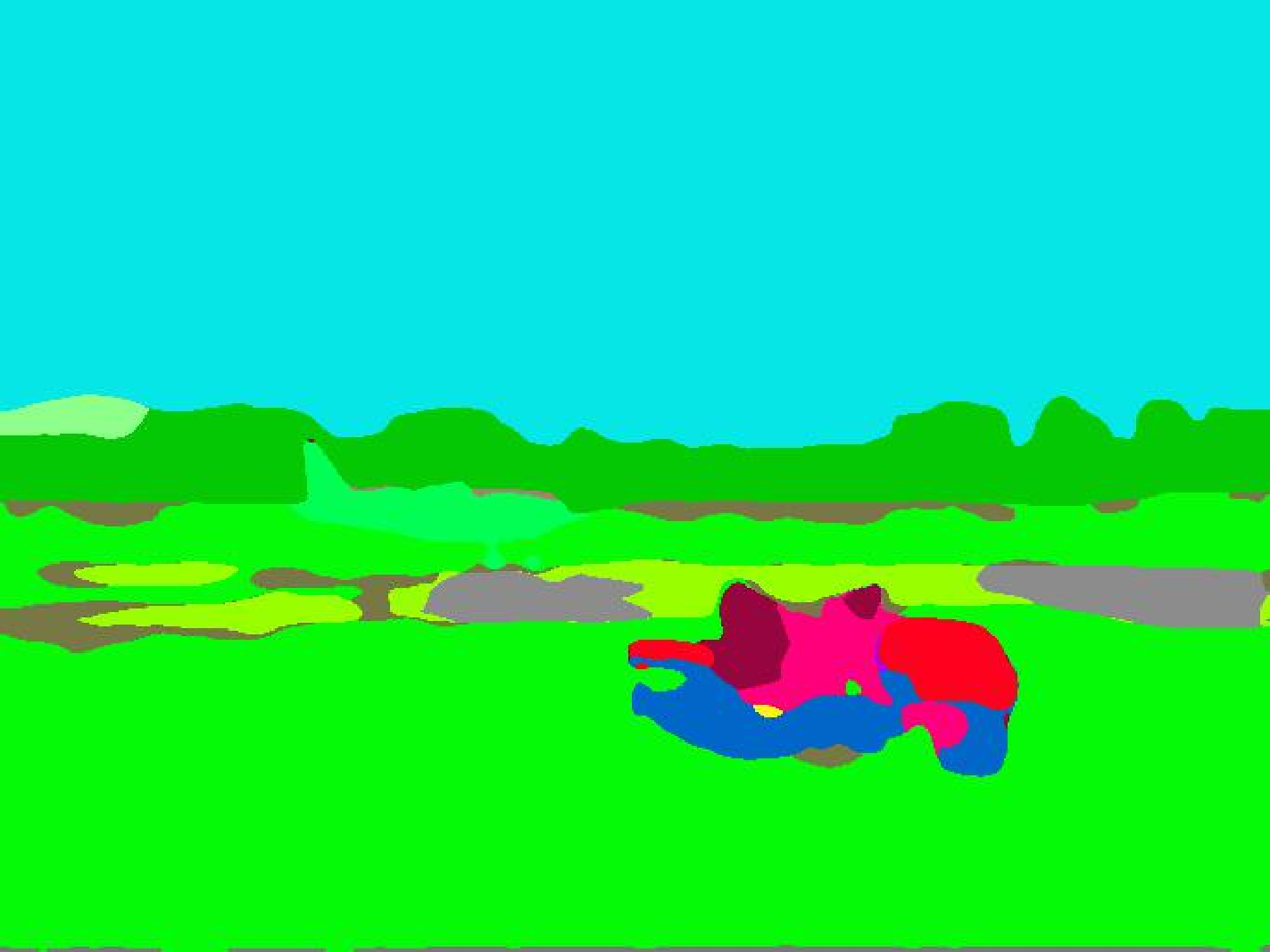}
     \end{subfigure}
     \hspace{-0.4em}
     \begin{subfigure}[b]{0.157\linewidth}
         \centering
         \includegraphics[width=\linewidth]{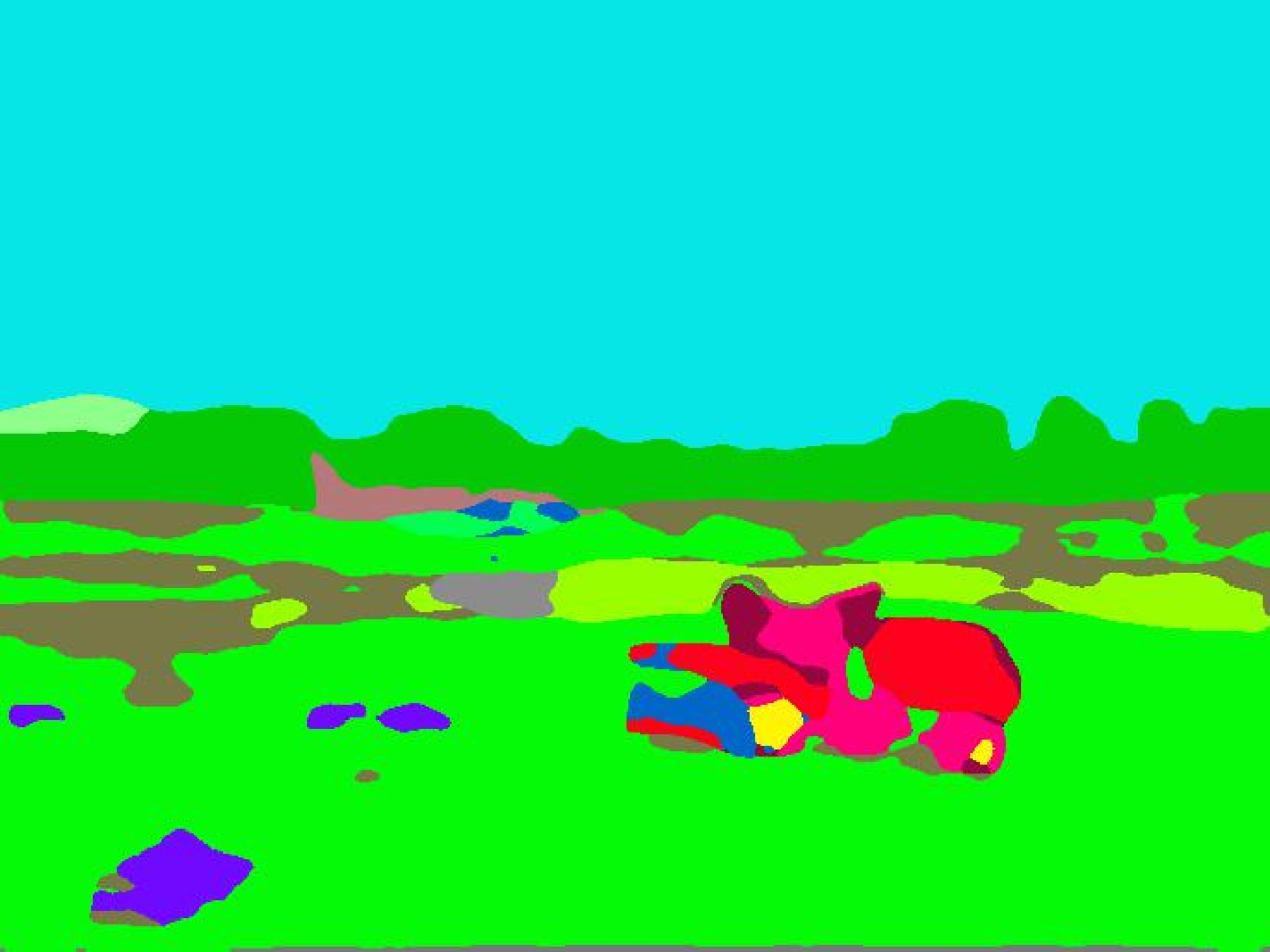}
     \end{subfigure}
     \vspace{1em}
     
     \begin{subfigure}[b]{0.157\linewidth}
         \centering
         \includegraphics[width=\linewidth]{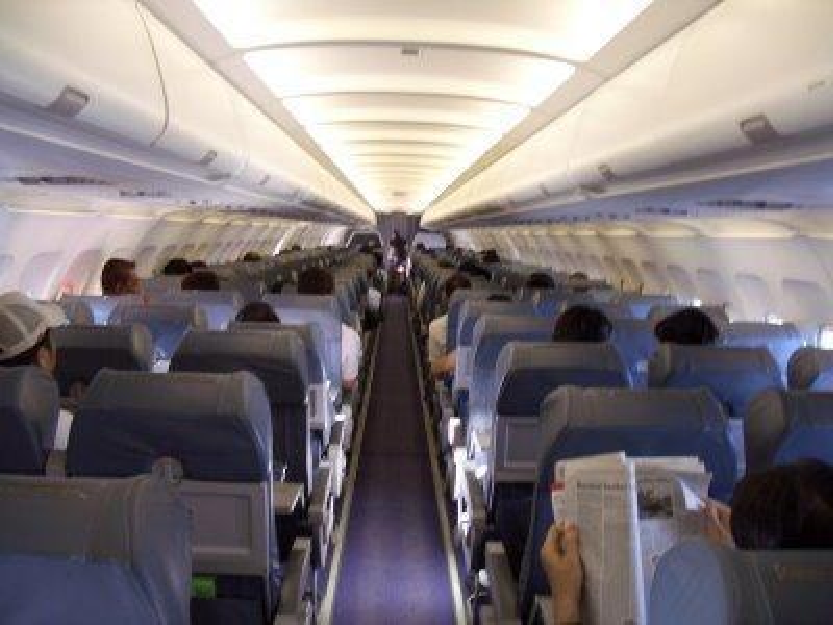}
     \end{subfigure}
     \hspace{-0.4em}
     \begin{subfigure}[b]{0.157\linewidth}
         \centering
         \includegraphics[width=\linewidth]{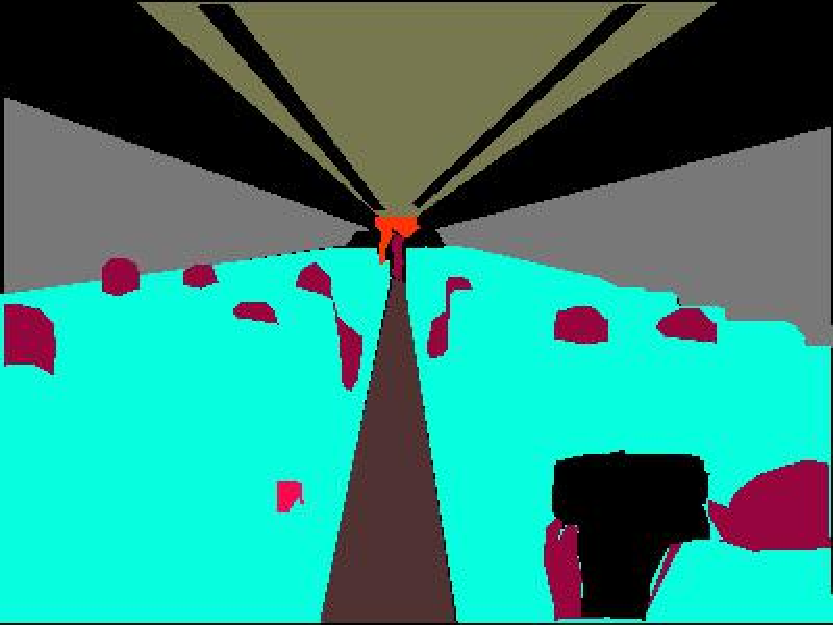}
     \end{subfigure}
     \hfill
     \begin{subfigure}[b]{0.157\linewidth}
         \centering
         \includegraphics[width=\linewidth]{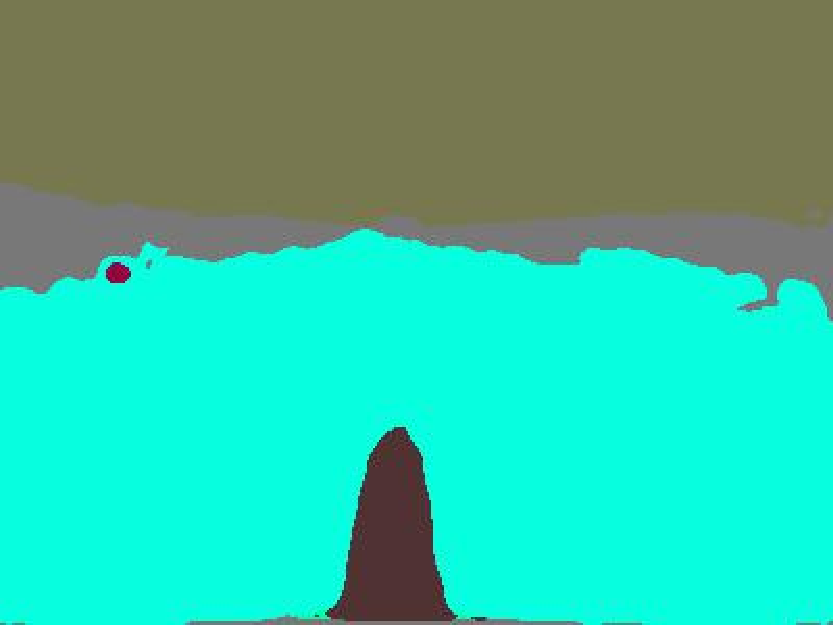}
     \end{subfigure}
     \hspace{-0.4em}
     \begin{subfigure}[b]{0.157\linewidth}
         \centering
         \includegraphics[width=\linewidth]{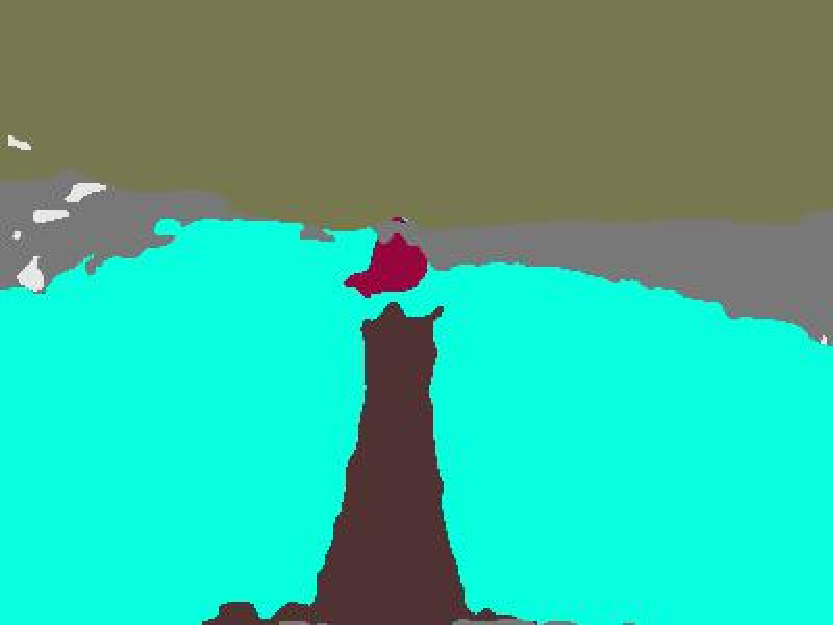}
     \end{subfigure}
     \hfill
     \begin{subfigure}[b]{0.157\linewidth}
         \centering
         \includegraphics[width=\linewidth]{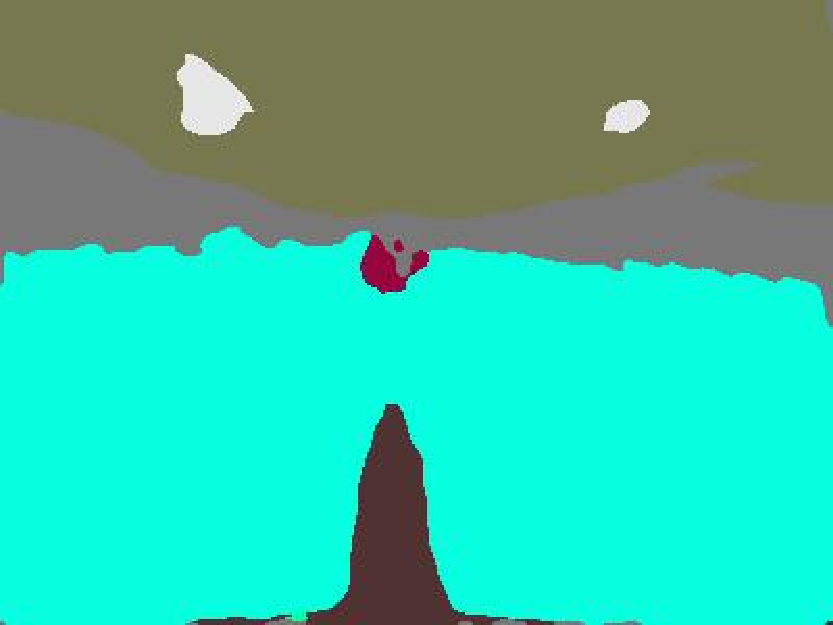}
     \end{subfigure}
     \hspace{-0.4em}
     \begin{subfigure}[b]{0.157\linewidth}
         \centering
         \includegraphics[width=\linewidth]{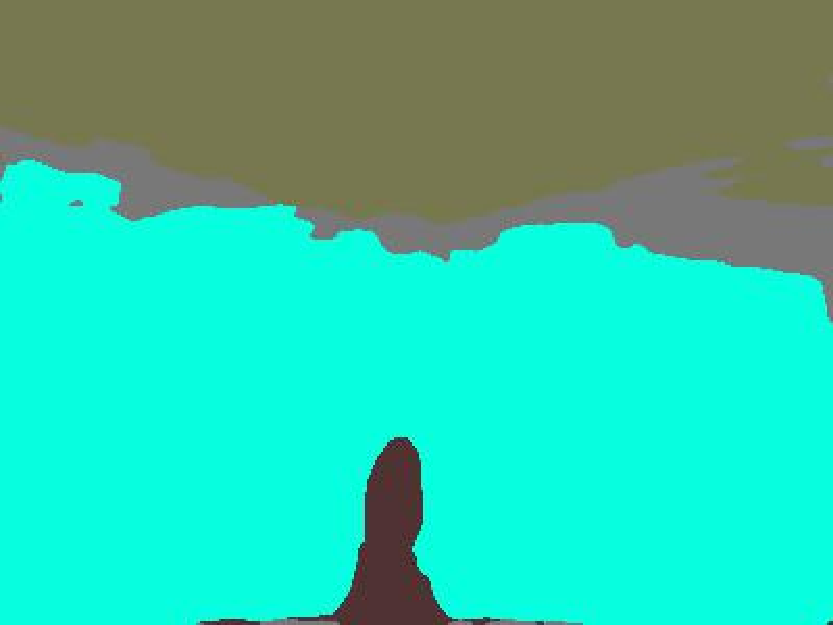}
     \end{subfigure}
     \vspace{1em}

     \begin{subfigure}[b]{0.157\linewidth}
         \centering
         \includegraphics[width=\linewidth]{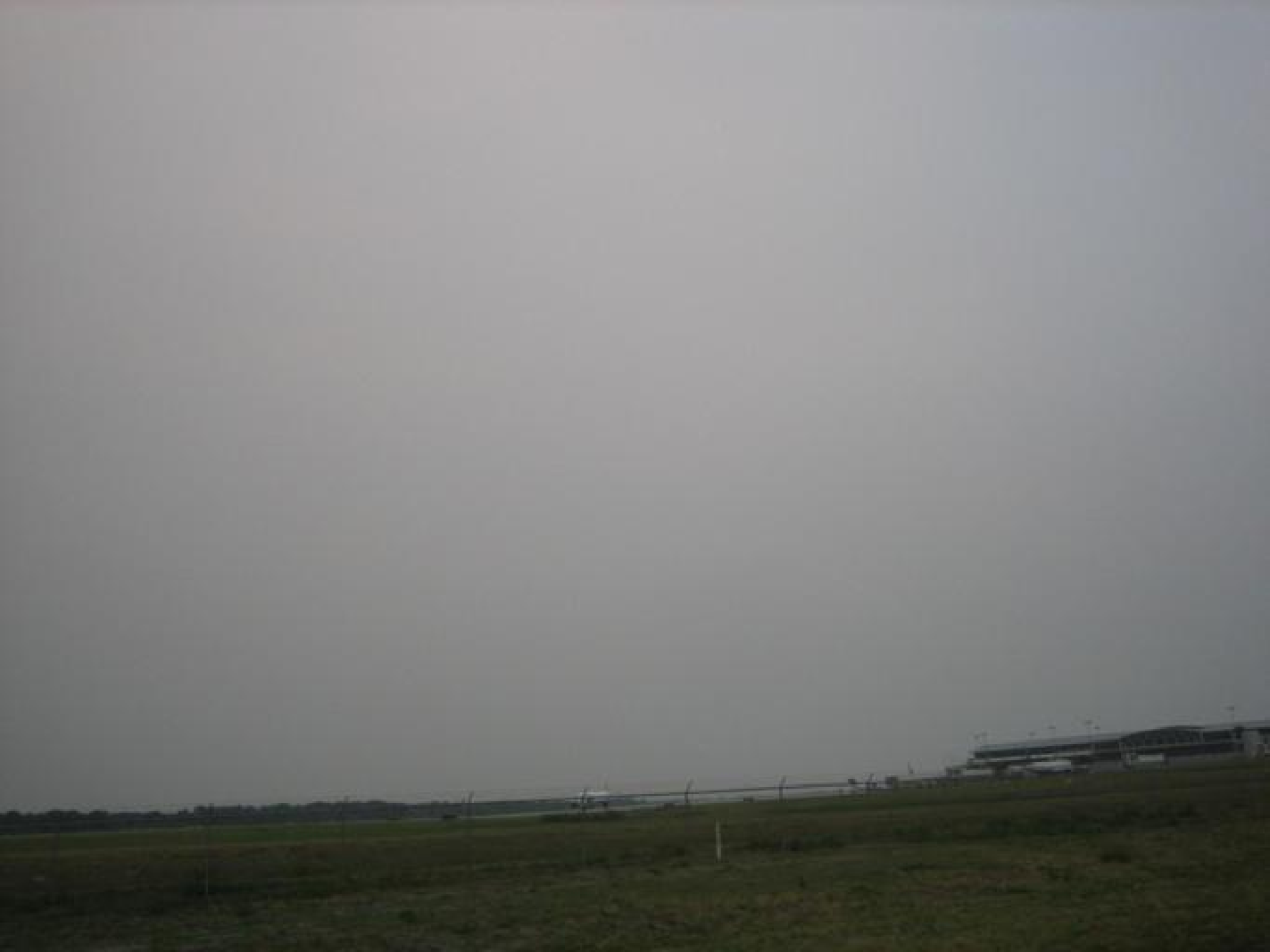}
     \end{subfigure}
     \hspace{-0.4em}
     \begin{subfigure}[b]{0.157\linewidth}
         \centering
         \includegraphics[width=\linewidth]{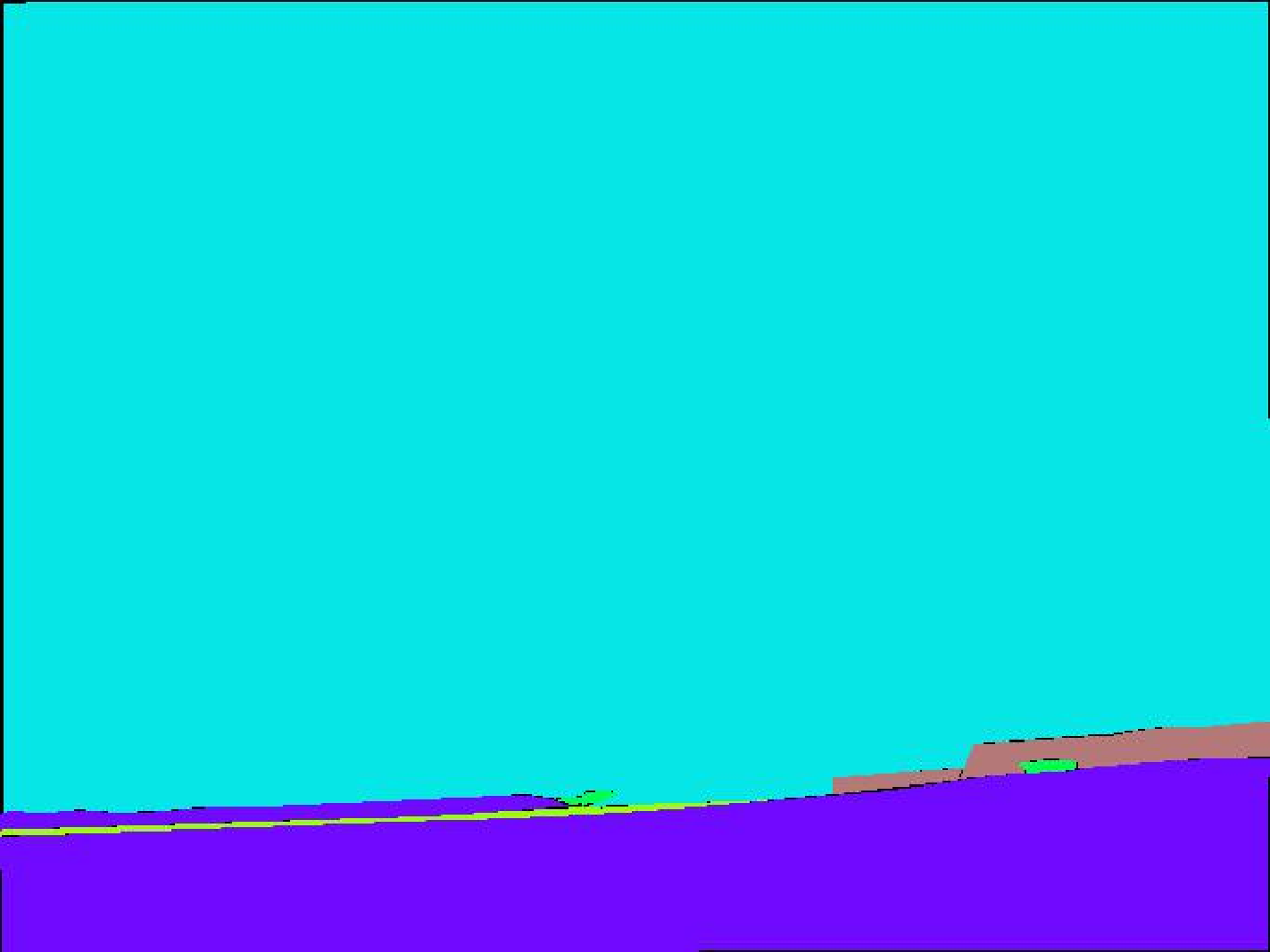}
     \end{subfigure}
     \hfill
     \begin{subfigure}[b]{0.157\linewidth}
         \centering
         \includegraphics[width=\linewidth]{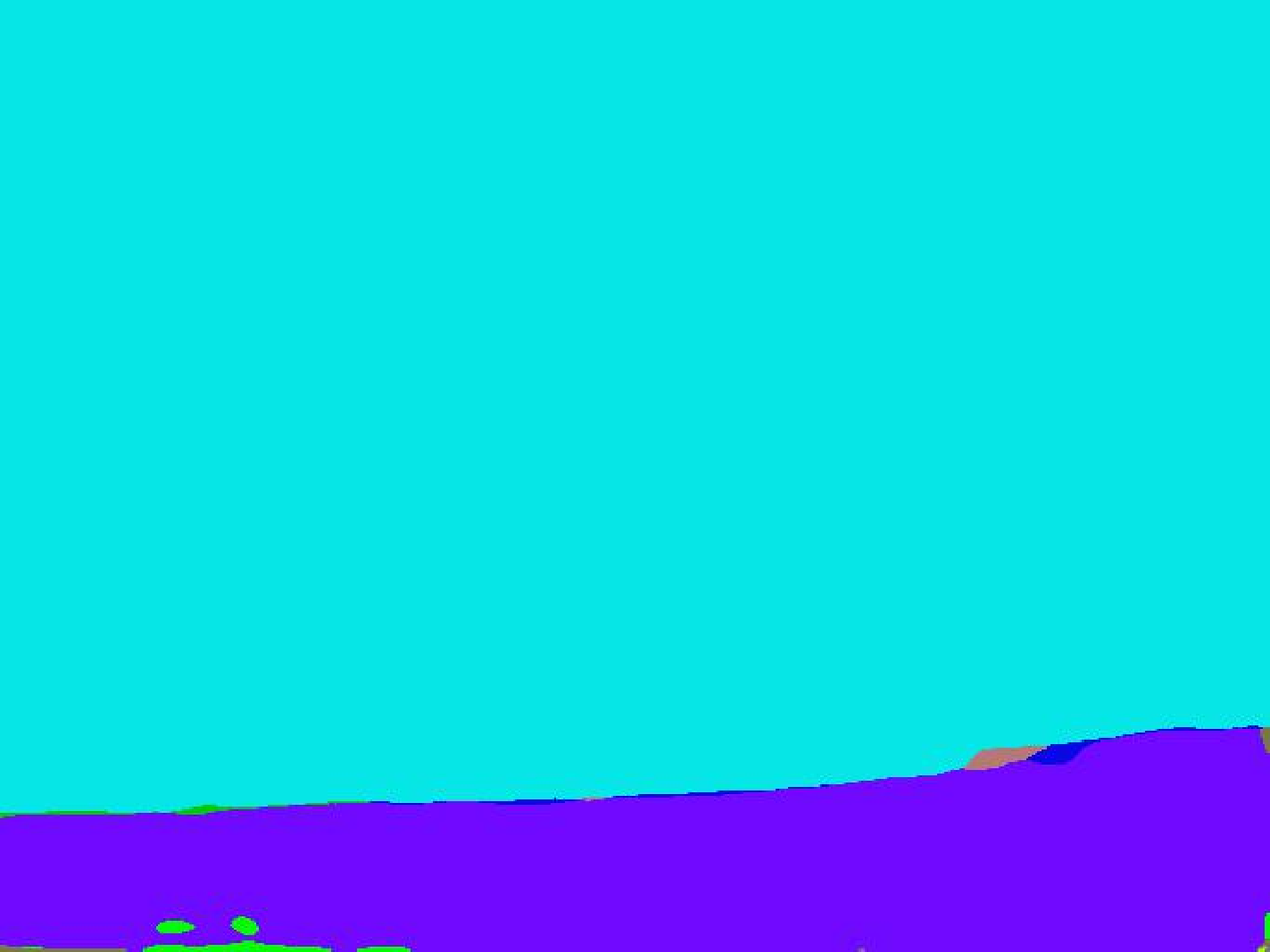}
     \end{subfigure}
     \hspace{-0.4em}
     \begin{subfigure}[b]{0.157\linewidth}
         \centering
         \includegraphics[width=\linewidth]{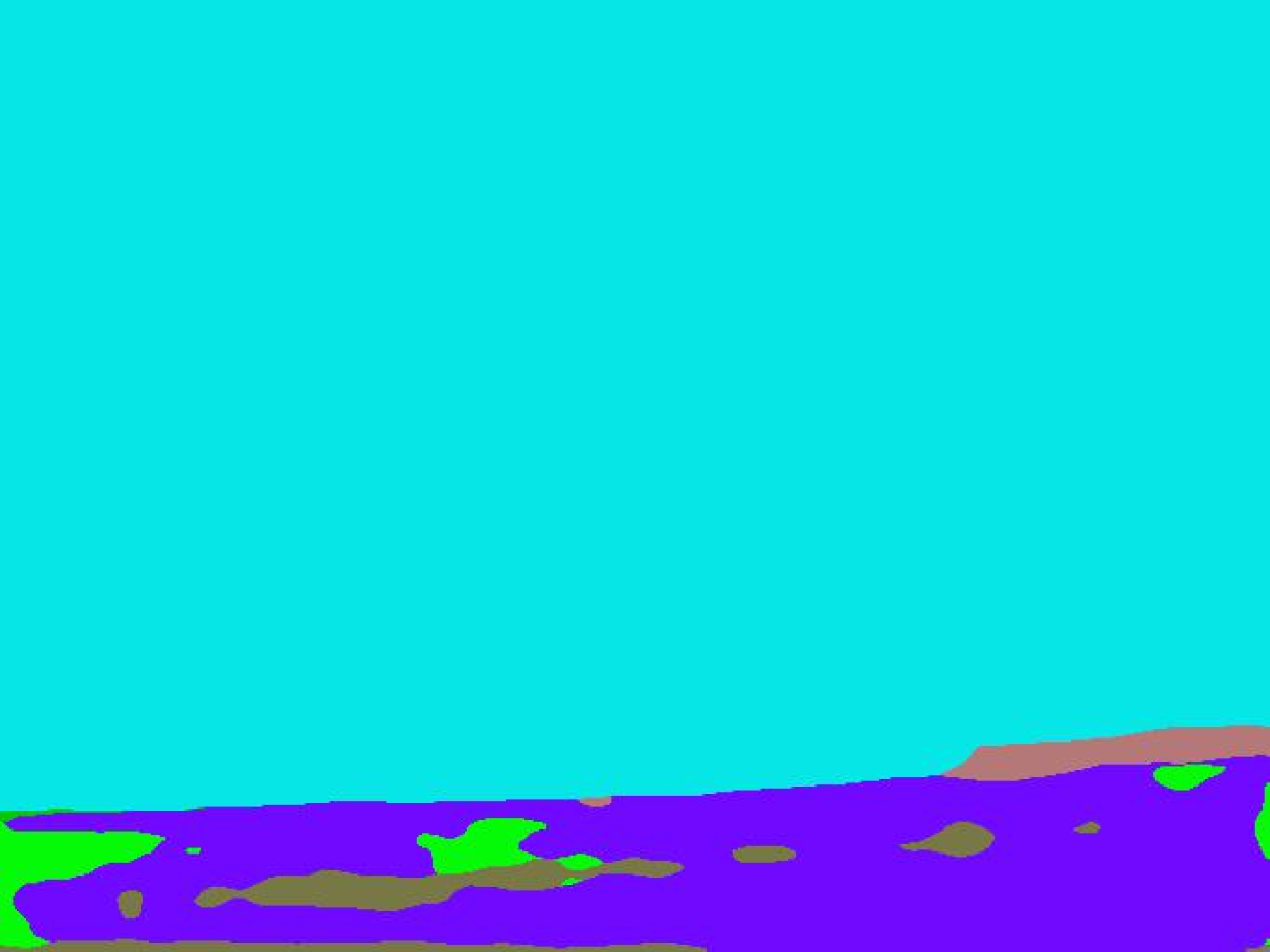}
     \end{subfigure}
     \hfill
     \begin{subfigure}[b]{0.157\linewidth}
         \centering
         \includegraphics[width=\linewidth]{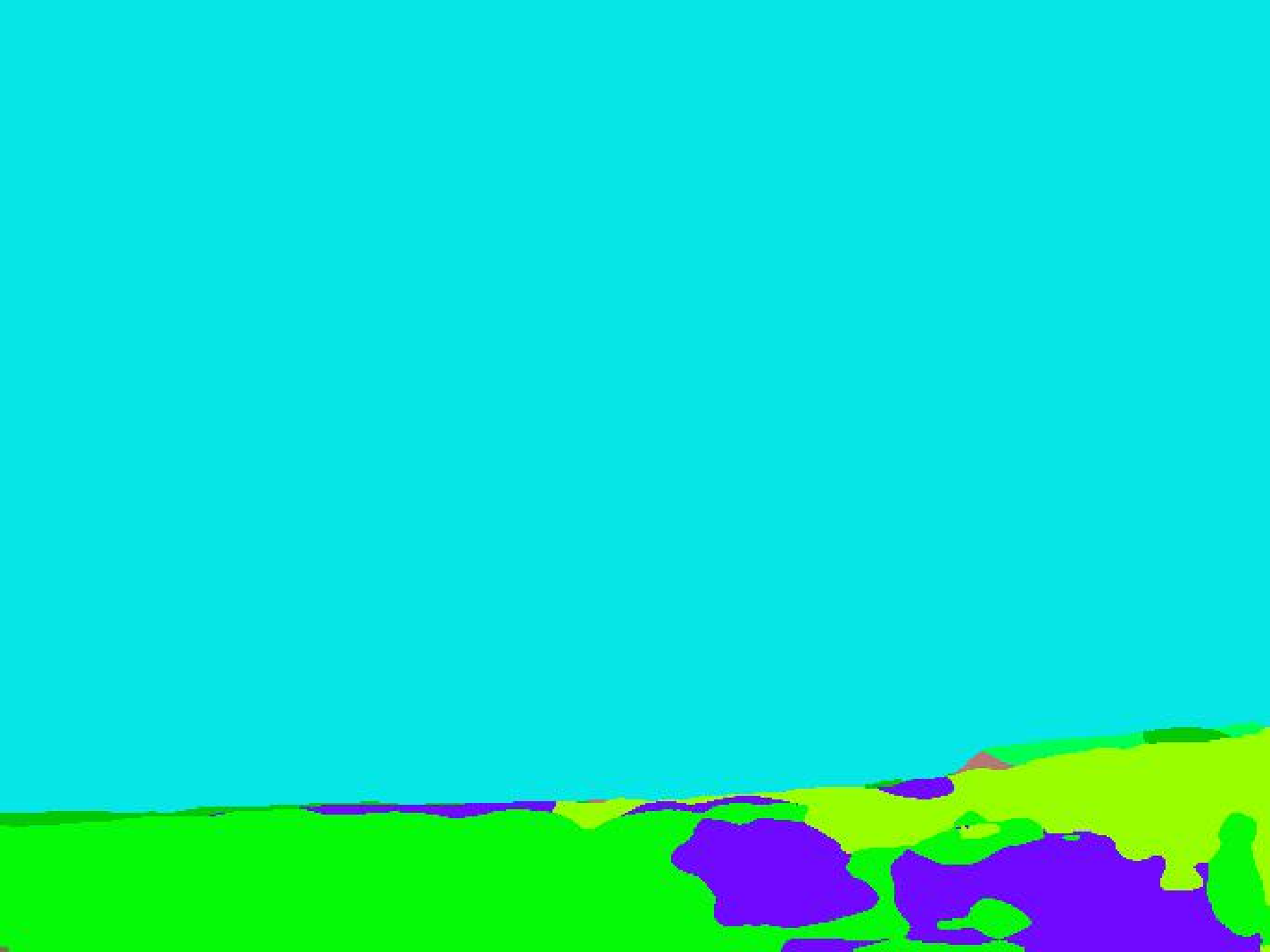}
     \end{subfigure}
     \hspace{-0.4em}
     \begin{subfigure}[b]{0.157\linewidth}
         \centering
         \includegraphics[width=\linewidth]{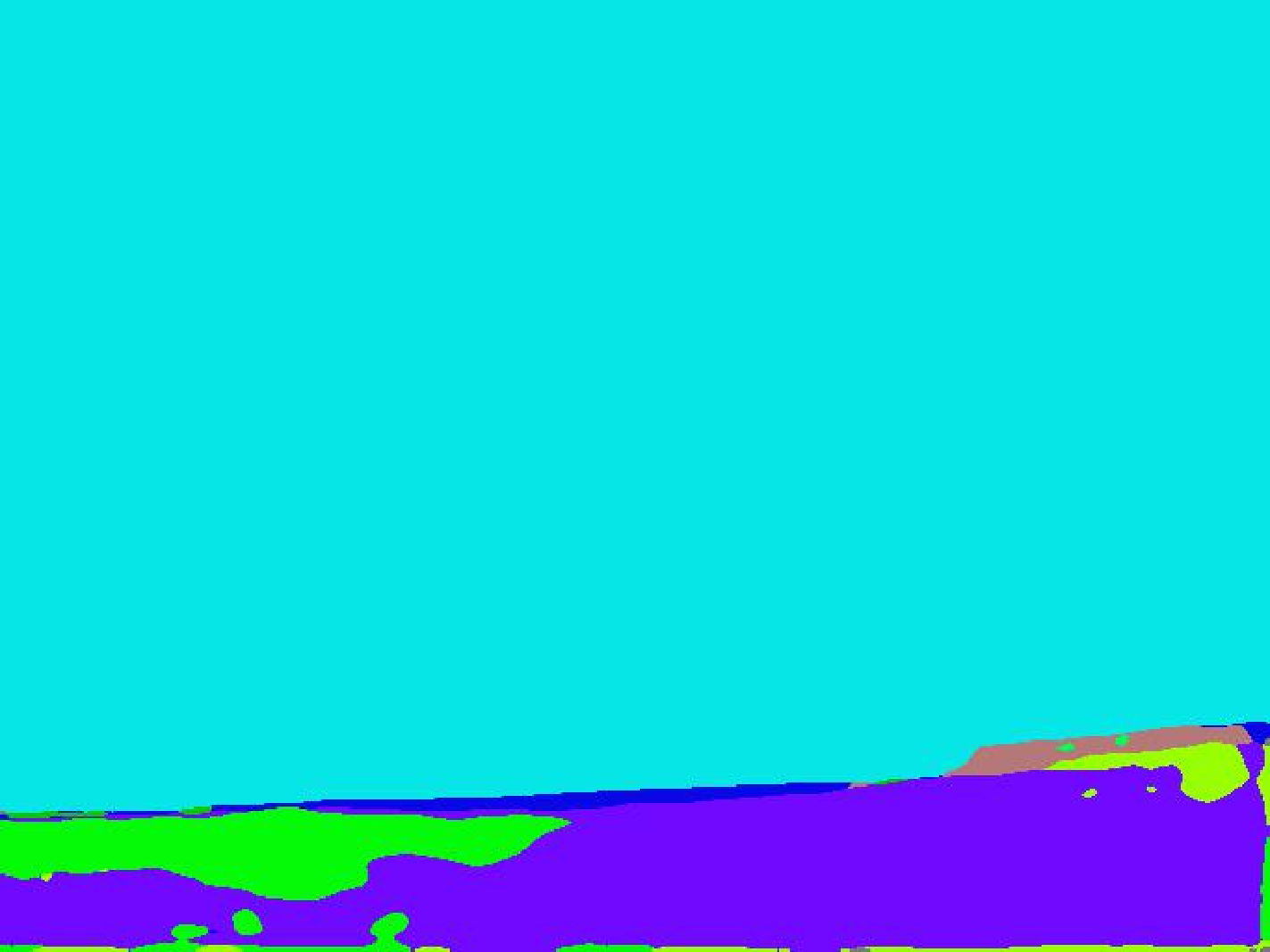}
     \end{subfigure}
     \vspace{1em}
     
     \begin{subfigure}[b]{0.157\linewidth}
         \centering
         \includegraphics[width=\linewidth]{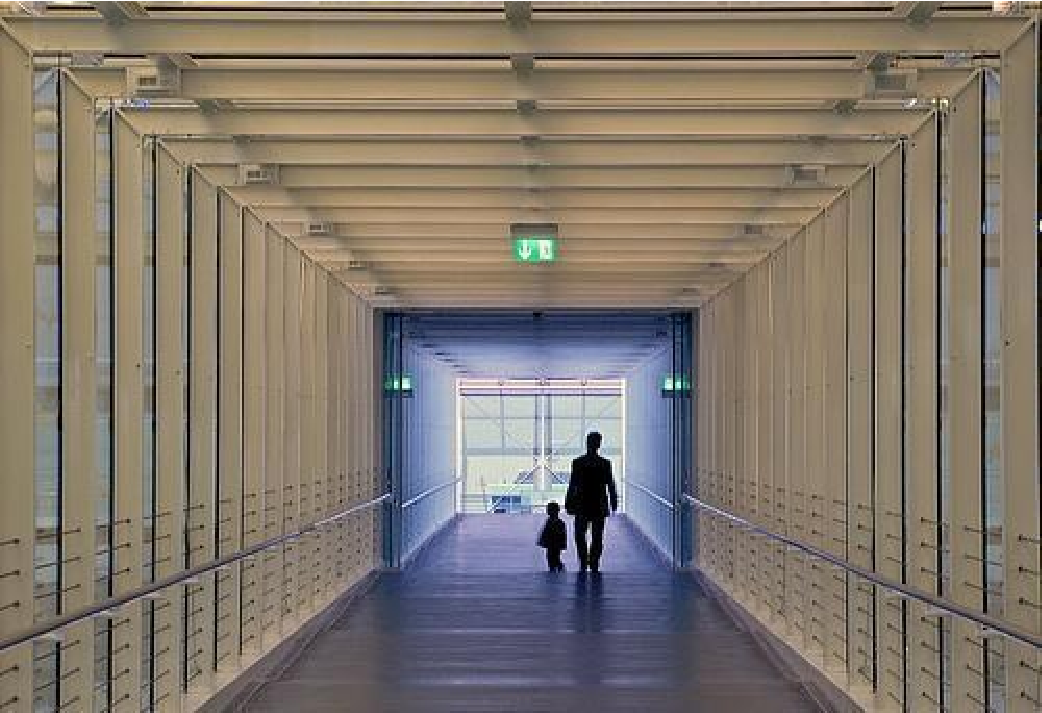}
     \end{subfigure}
     \hspace{-0.4em}
     \begin{subfigure}[b]{0.157\linewidth}
         \centering
         \includegraphics[width=\linewidth]{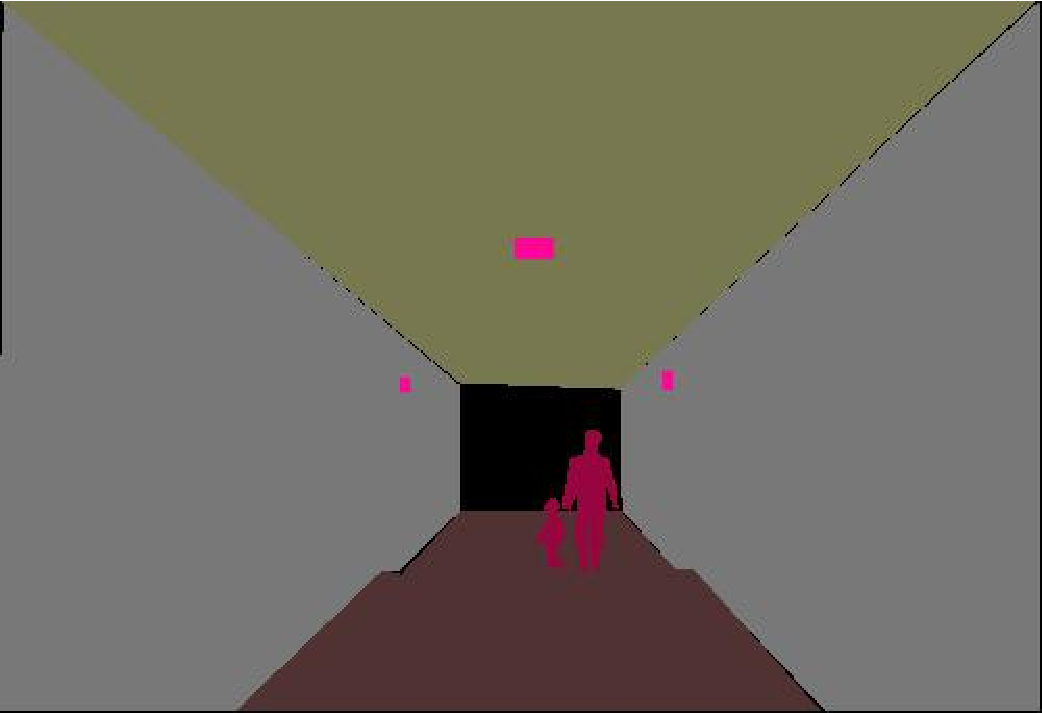}
     \end{subfigure}
     \hfill
     \begin{subfigure}[b]{0.157\linewidth}
         \centering
         \includegraphics[width=\linewidth]{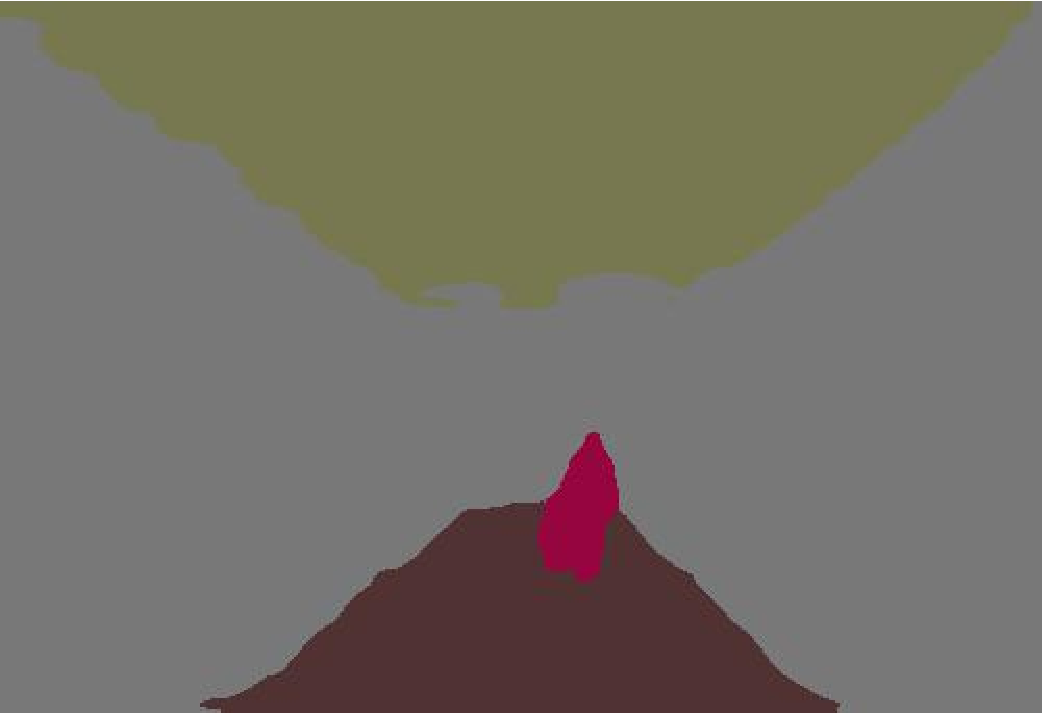}
     \end{subfigure}
     \hspace{-0.4em}
     \begin{subfigure}[b]{0.157\linewidth}
         \centering
         \includegraphics[width=\linewidth]{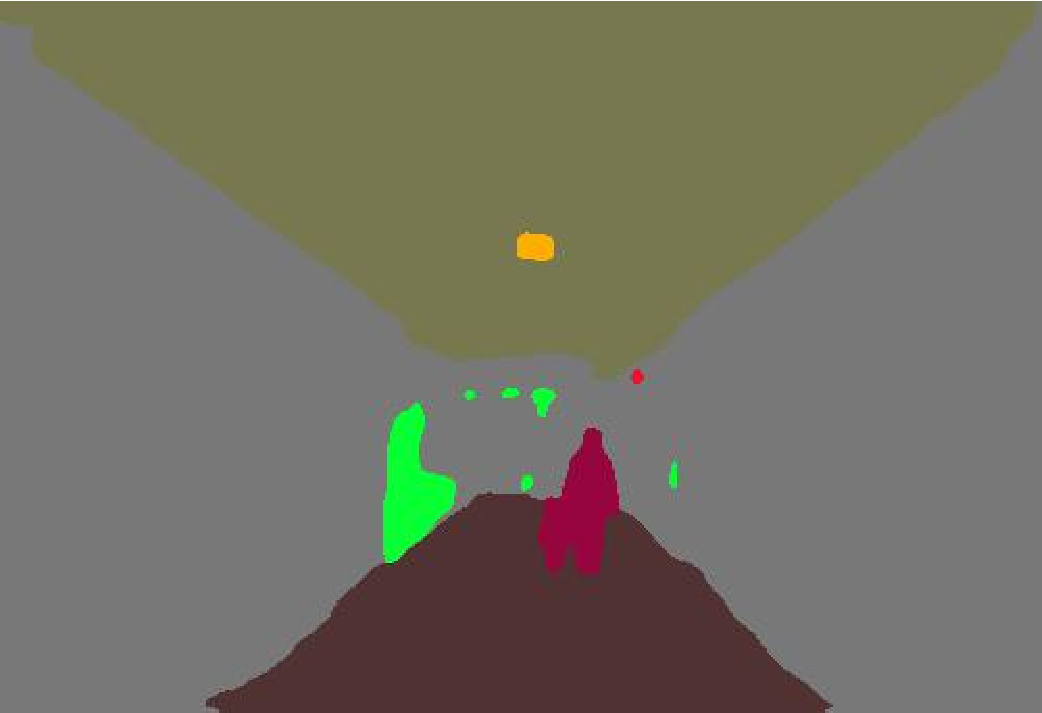}
     \end{subfigure}
     \hfill
     \begin{subfigure}[b]{0.157\linewidth}
         \centering
         \includegraphics[width=\linewidth]{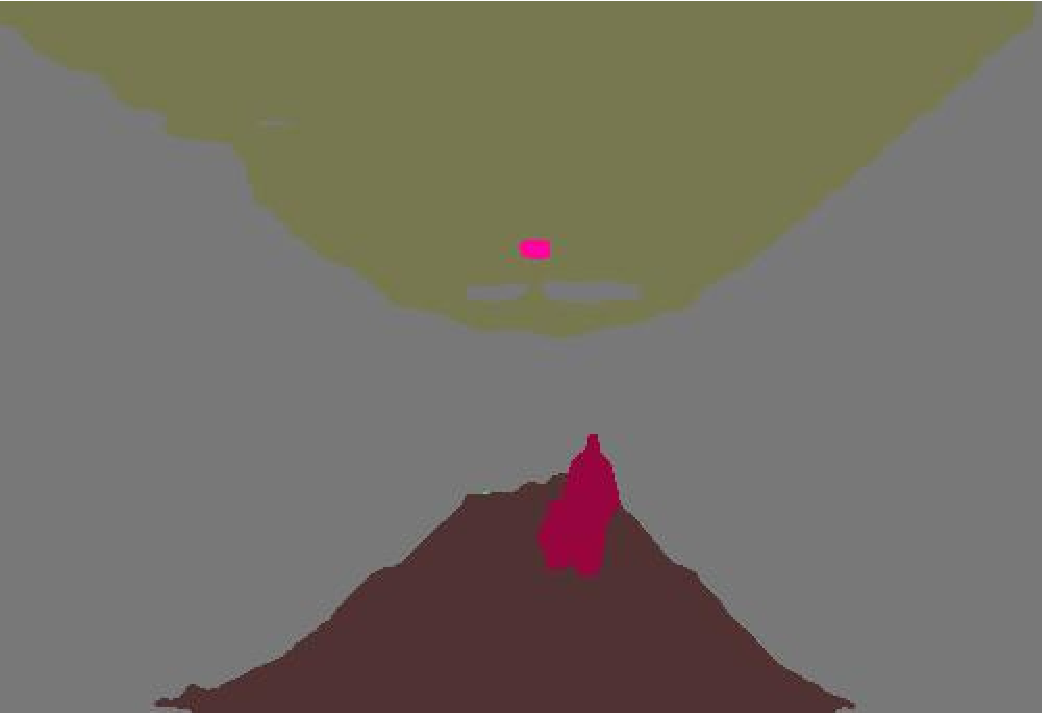}
     \end{subfigure}
     \hspace{-0.4em}
     \begin{subfigure}[b]{0.157\linewidth}
         \centering
         \includegraphics[width=\linewidth]{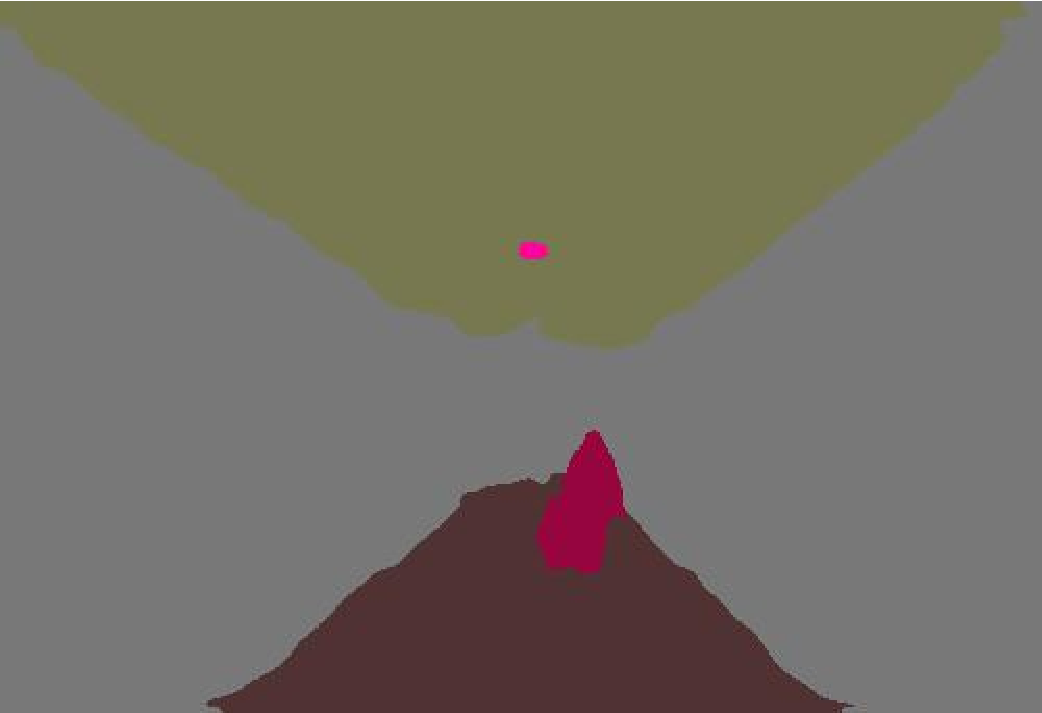}
     \end{subfigure}
     \vspace{1em}
     
     \begin{subfigure}[b]{0.157\linewidth}
         \centering
         \includegraphics[width=\linewidth]{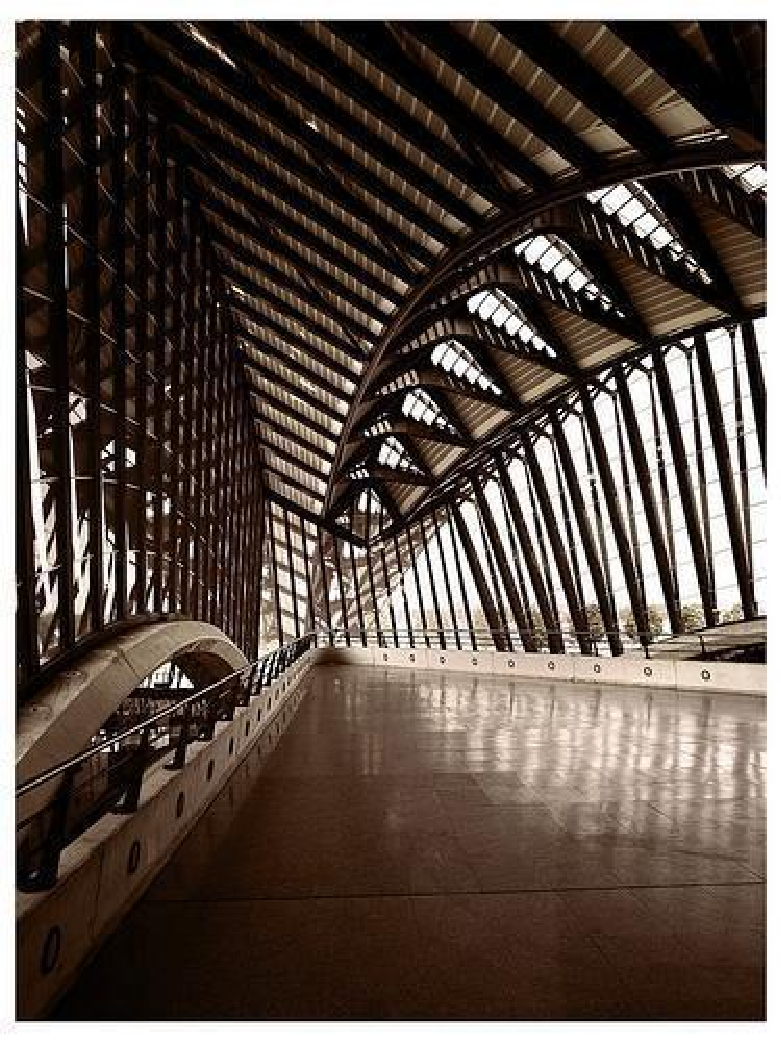}
     \end{subfigure}
     \hspace{-0.4em}
     \begin{subfigure}[b]{0.157\linewidth}
         \centering
         \includegraphics[width=\linewidth]{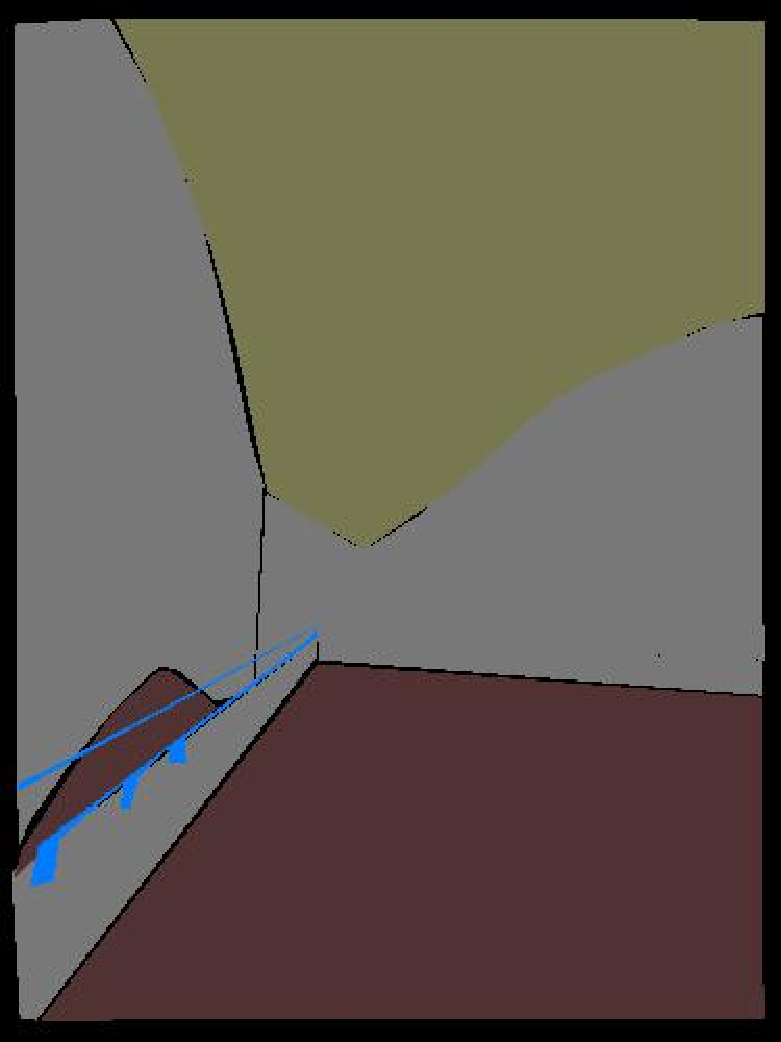}
     \end{subfigure}
     \hfill
     \begin{subfigure}[b]{0.157\linewidth}
         \centering
         \includegraphics[width=\linewidth]{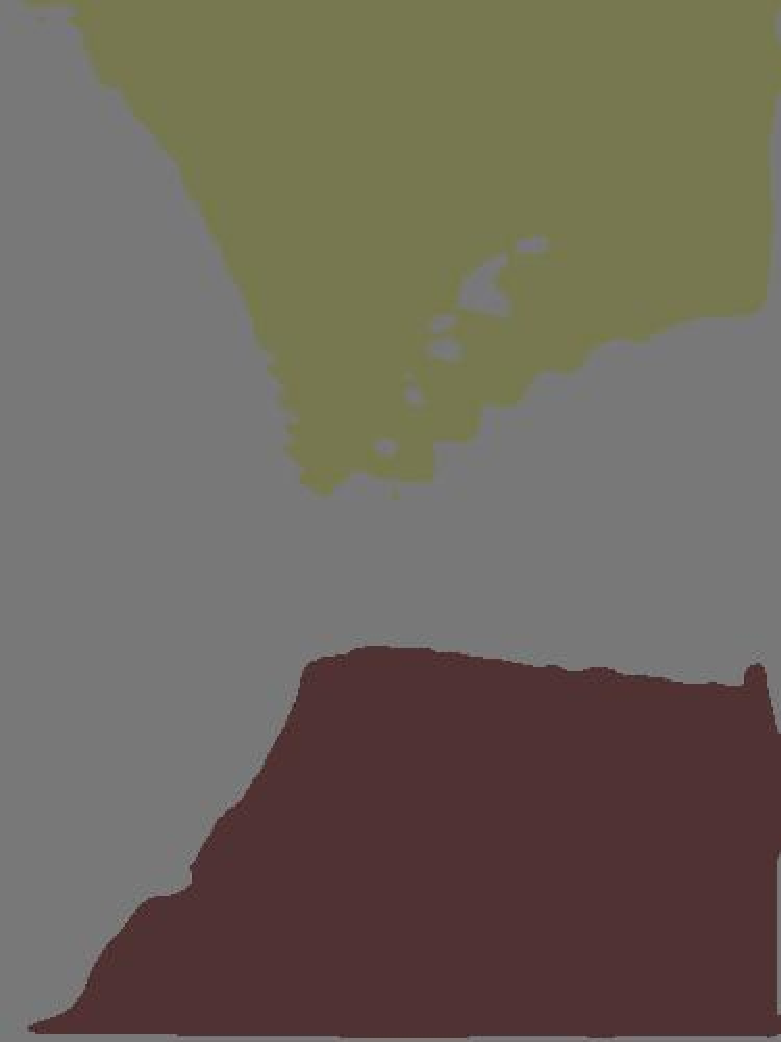}
     \end{subfigure}
     \hspace{-0.4em}
     \begin{subfigure}[b]{0.157\linewidth}
         \centering
         \includegraphics[width=\linewidth]{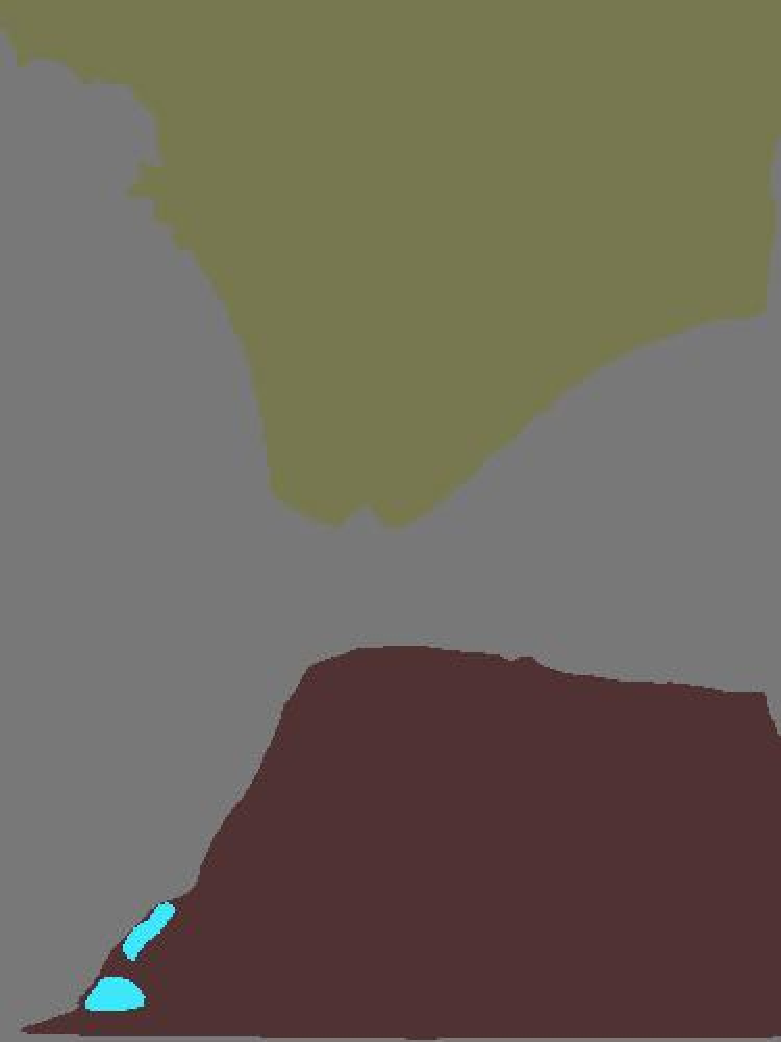}
     \end{subfigure}
     \hfill
     \begin{subfigure}[b]{0.157\linewidth}
         \centering
         \includegraphics[width=\linewidth]{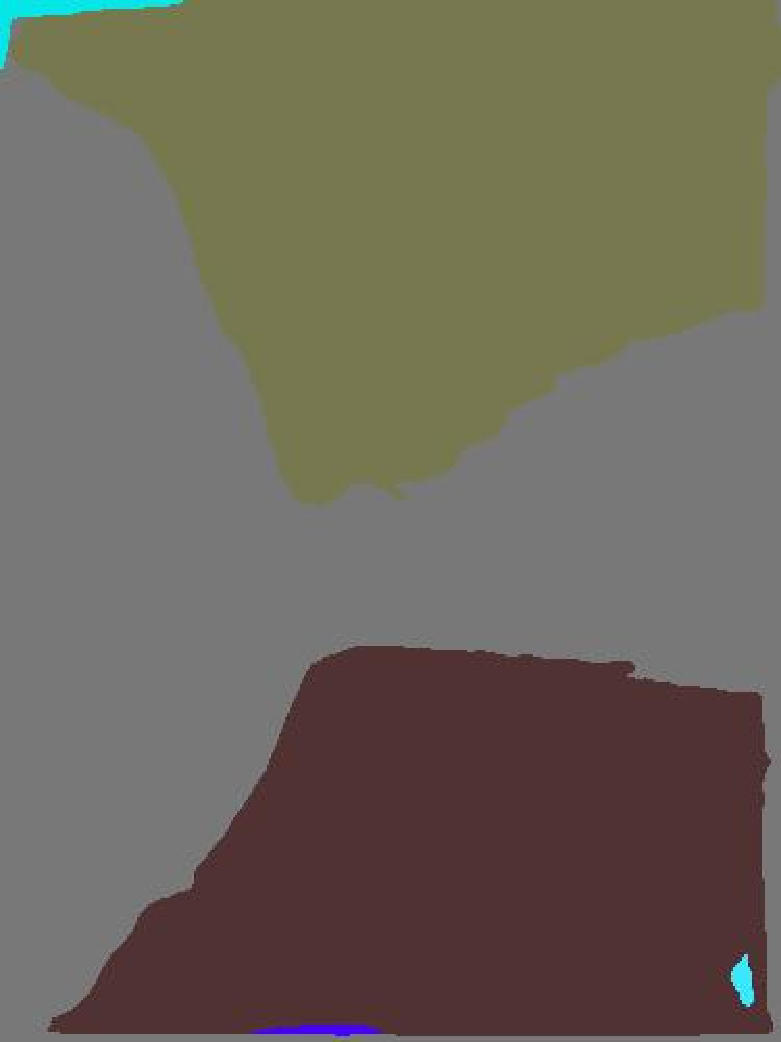}
     \end{subfigure}
     \hspace{-0.4em}
     \begin{subfigure}[b]{0.157\linewidth}
         \centering
         \includegraphics[width=\linewidth]{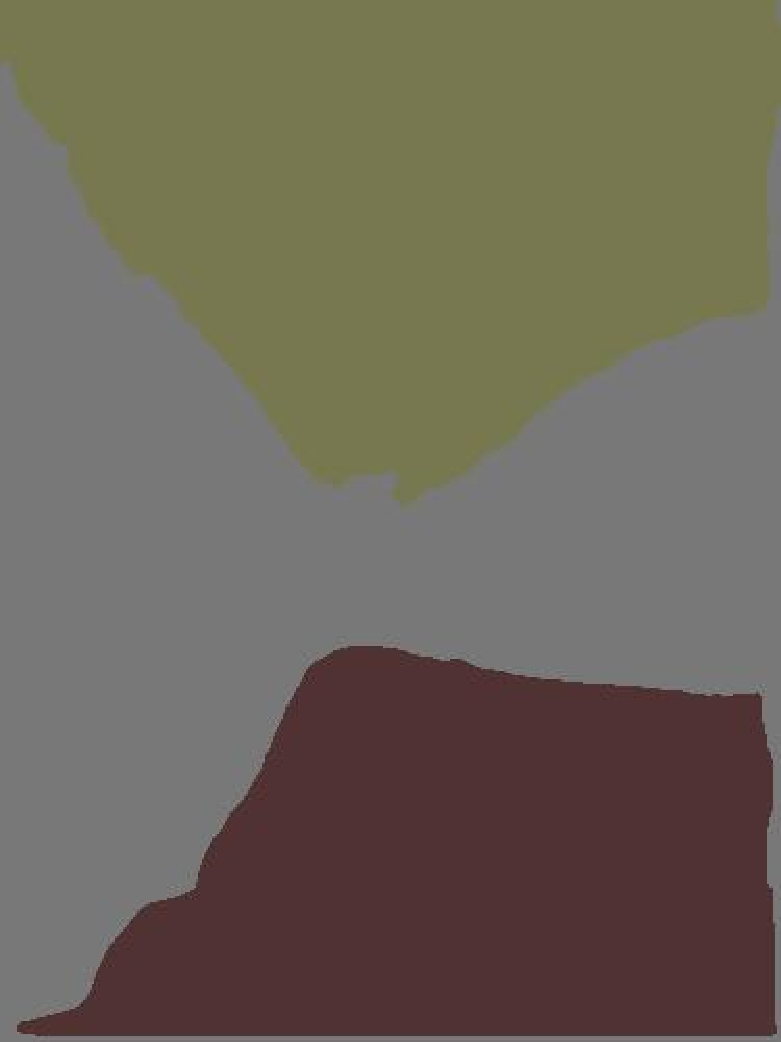}
     \end{subfigure}
     \vspace{1em}
     
     \begin{subfigure}[b]{0.157\linewidth}
         \centering
         \includegraphics[width=\linewidth]{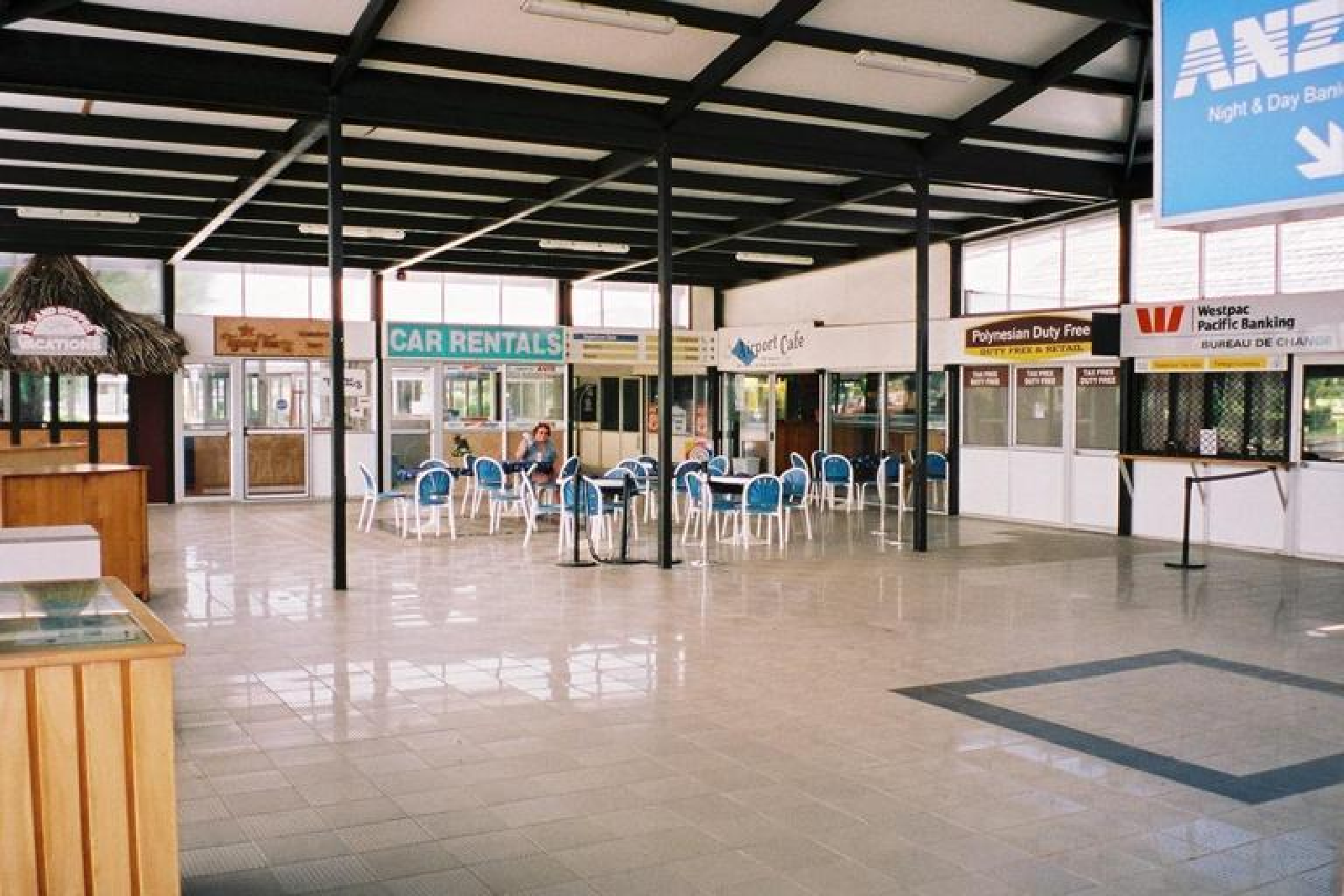}
     \end{subfigure}
     \hspace{-0.4em}
     \begin{subfigure}[b]{0.157\linewidth}
         \centering
         \includegraphics[width=\linewidth]{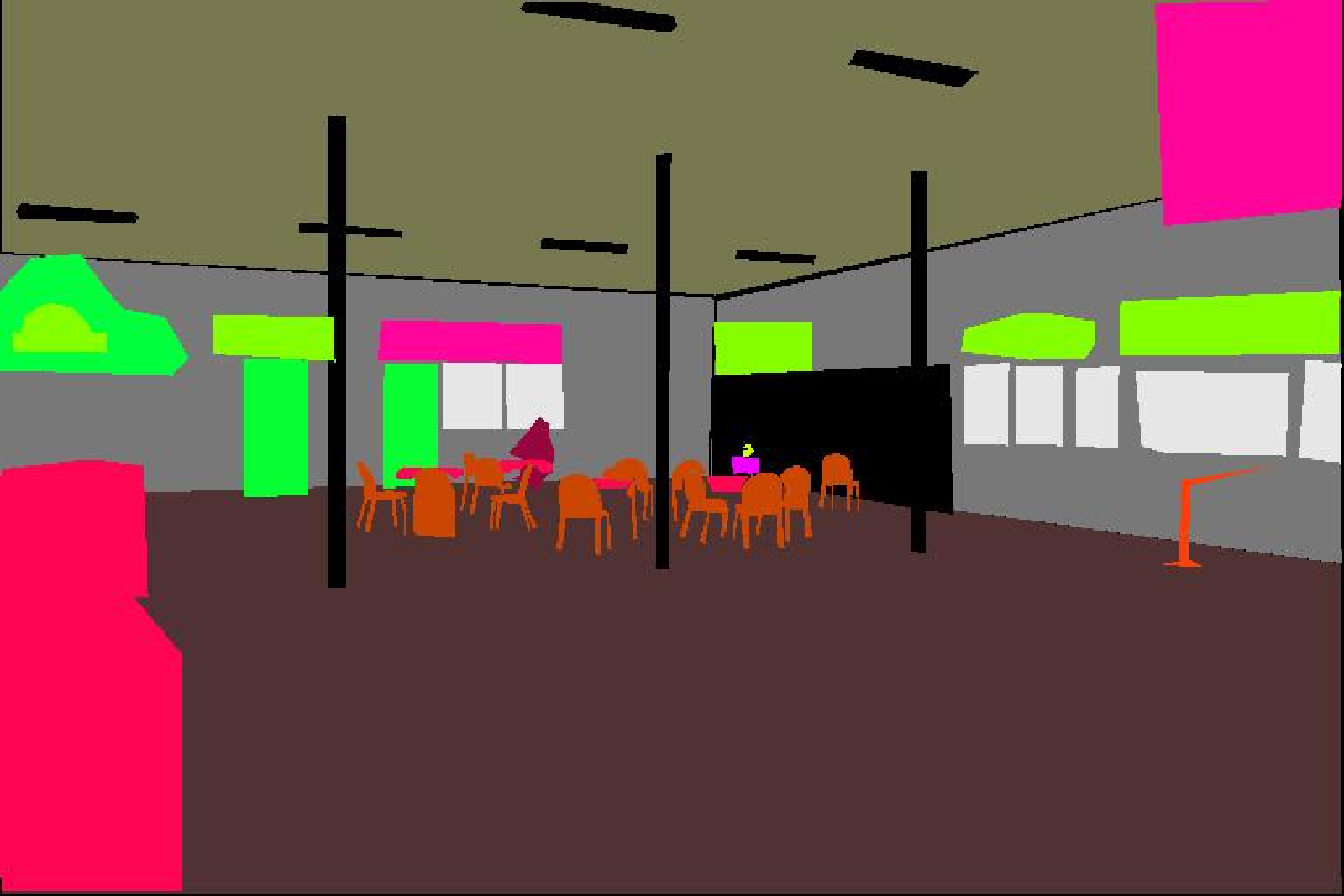}
     \end{subfigure}
     \hfill
     \begin{subfigure}[b]{0.157\linewidth}
         \centering
         \includegraphics[width=\linewidth]{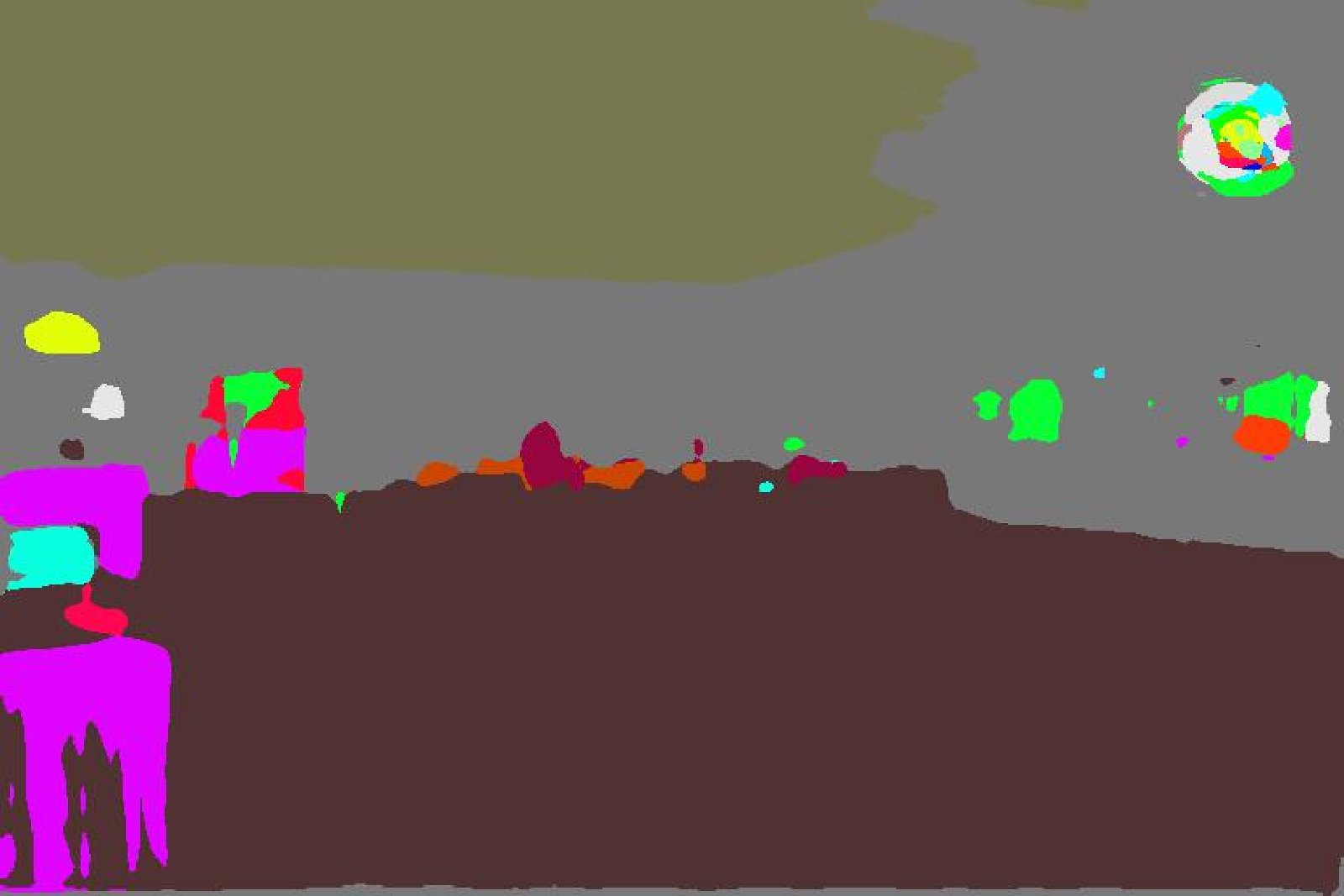}
     \end{subfigure}
     \hspace{-0.4em}
     \begin{subfigure}[b]{0.157\linewidth}
         \centering
         \includegraphics[width=\linewidth]{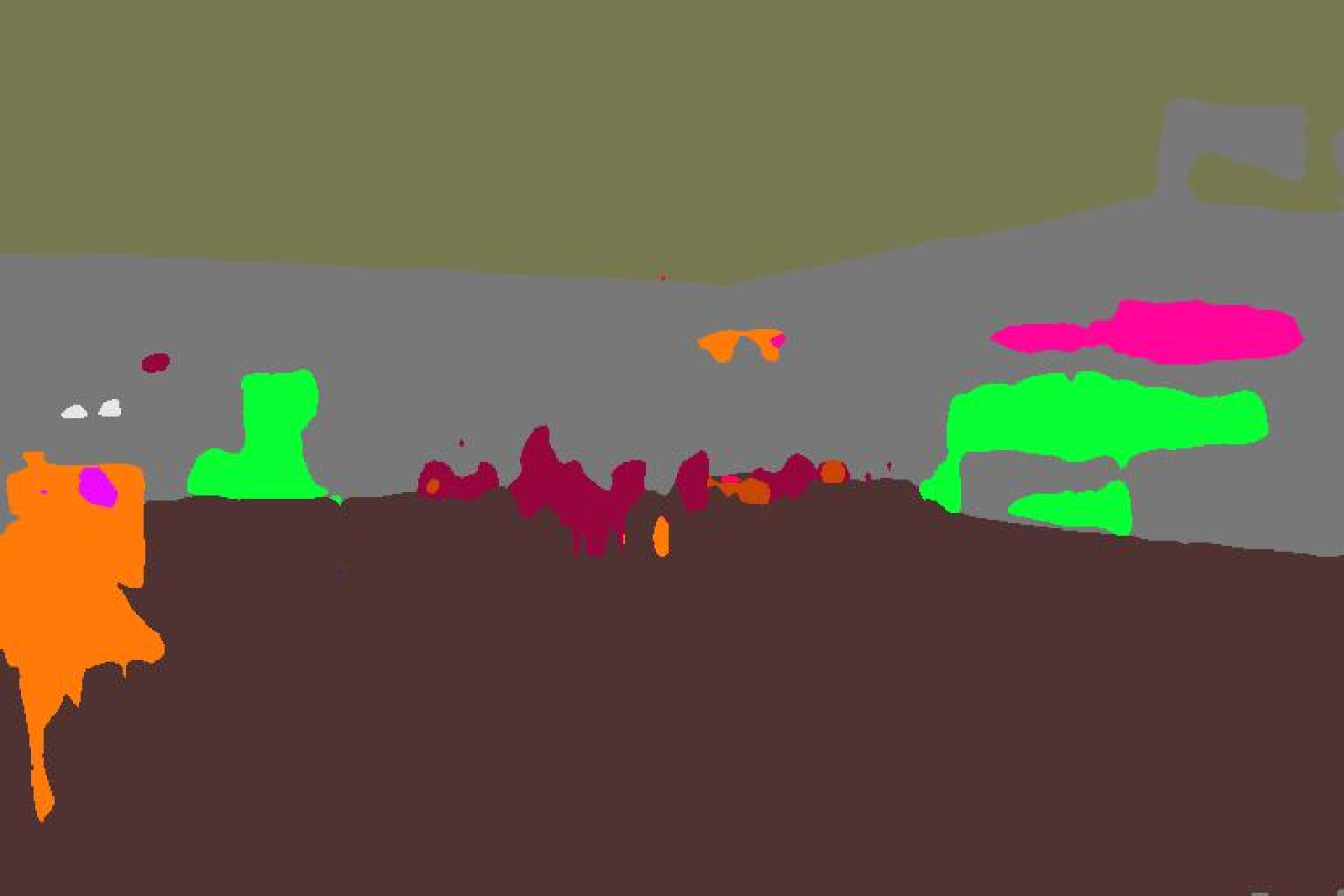}
     \end{subfigure}
     \hfill
     \begin{subfigure}[b]{0.157\linewidth}
         \centering
         \includegraphics[width=\linewidth]{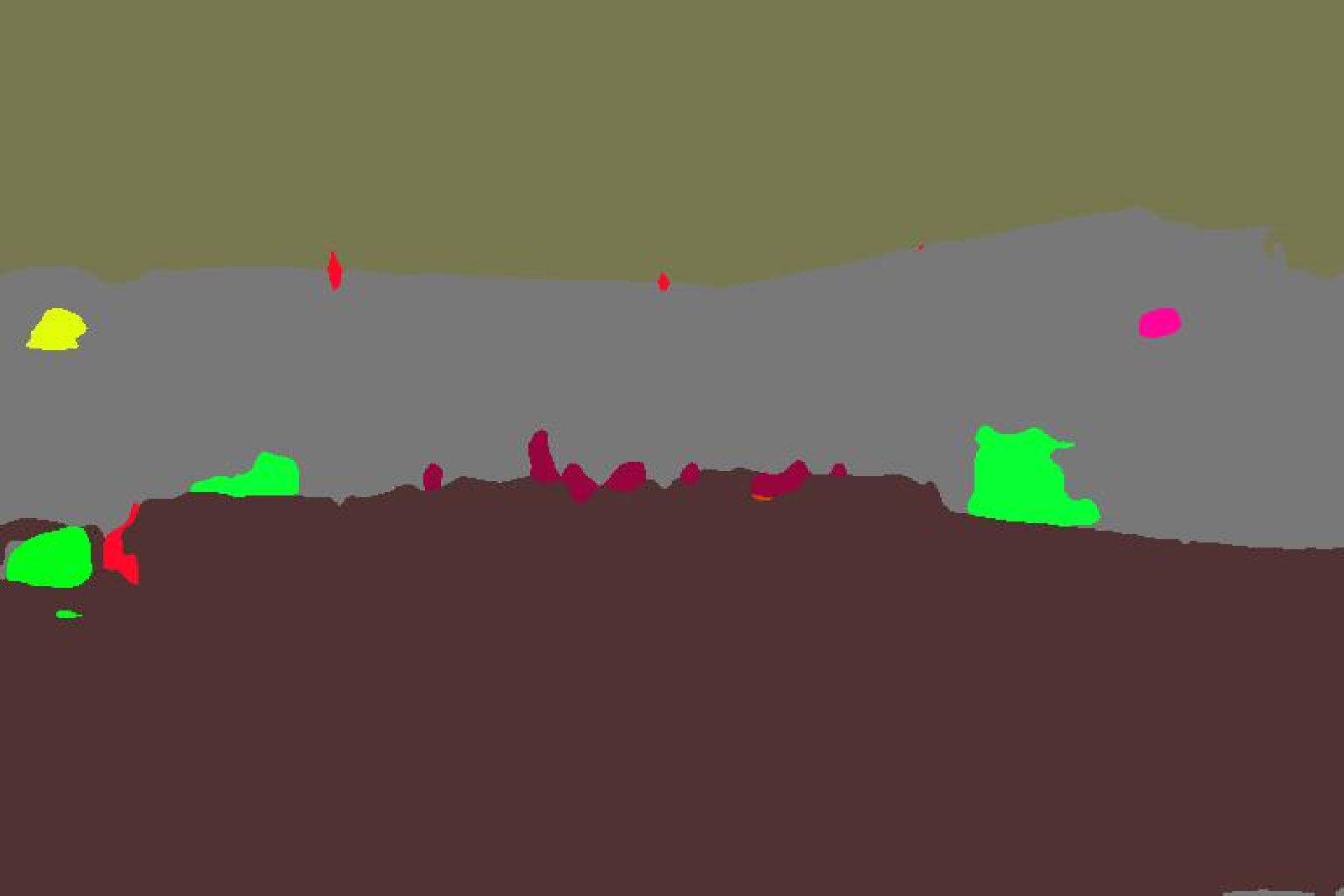}
     \end{subfigure}
     \hspace{-0.4em}
     \begin{subfigure}[b]{0.157\linewidth}
         \centering
         \includegraphics[width=\linewidth]{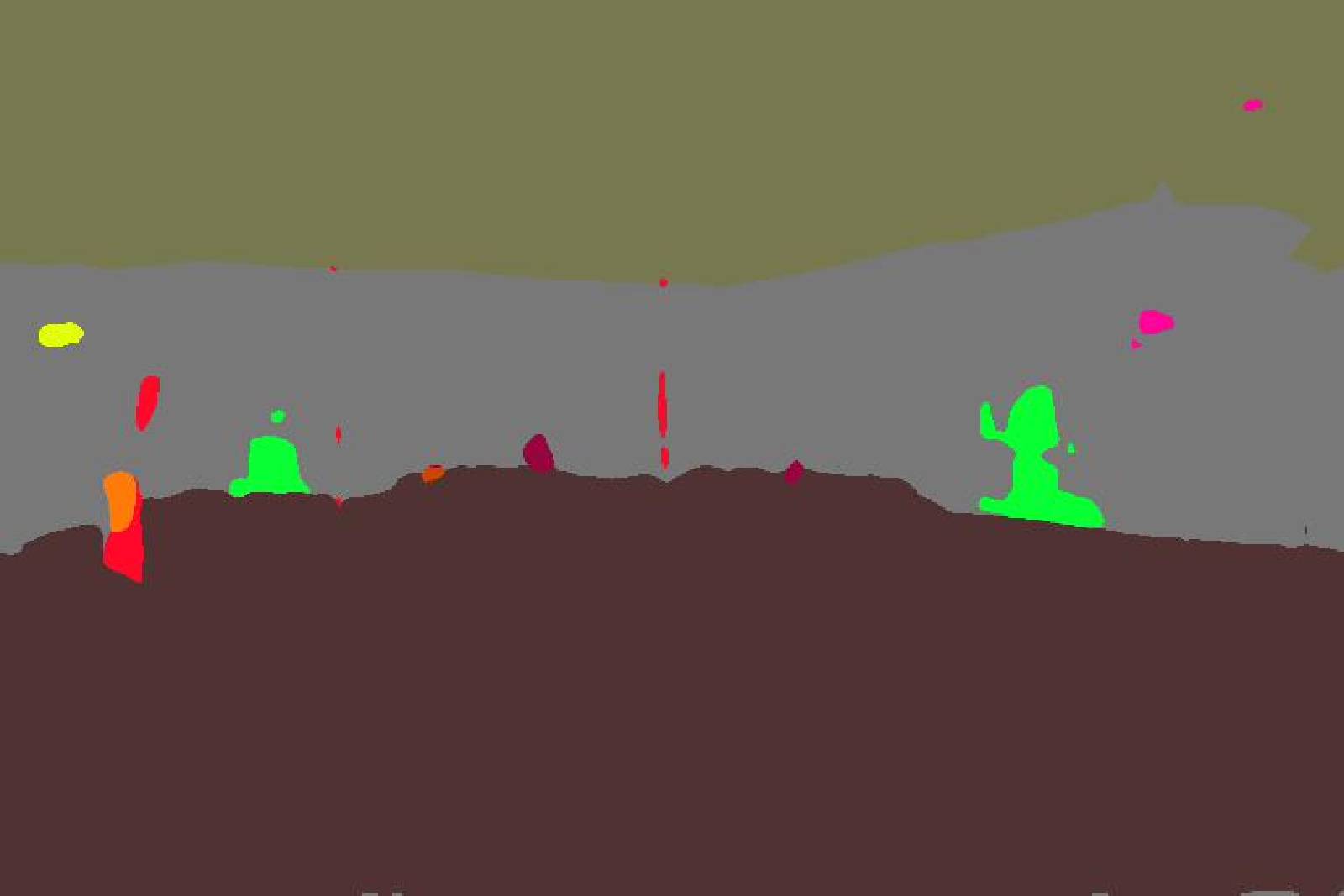}
     \end{subfigure}
     \vspace{1em}
     
     \begin{subfigure}[b]{0.157\linewidth}
         \centering
         \includegraphics[width=\linewidth]{figures/semantic-segmentation/img/ADE_val_00000010_img.eps}
     \end{subfigure}
     \hspace{-0.4em}
     \begin{subfigure}[b]{0.157\linewidth}
         \centering
         \includegraphics[width=\linewidth]{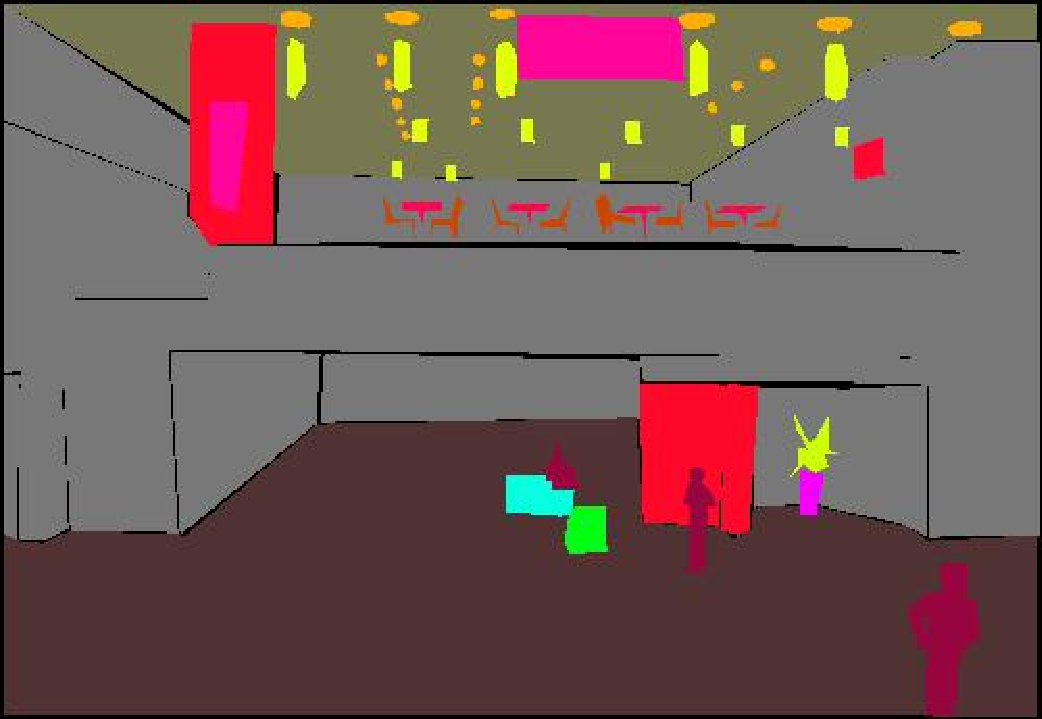}
     \end{subfigure}
     \hfill
     \begin{subfigure}[b]{0.157\linewidth}
         \centering
         \includegraphics[width=\linewidth]{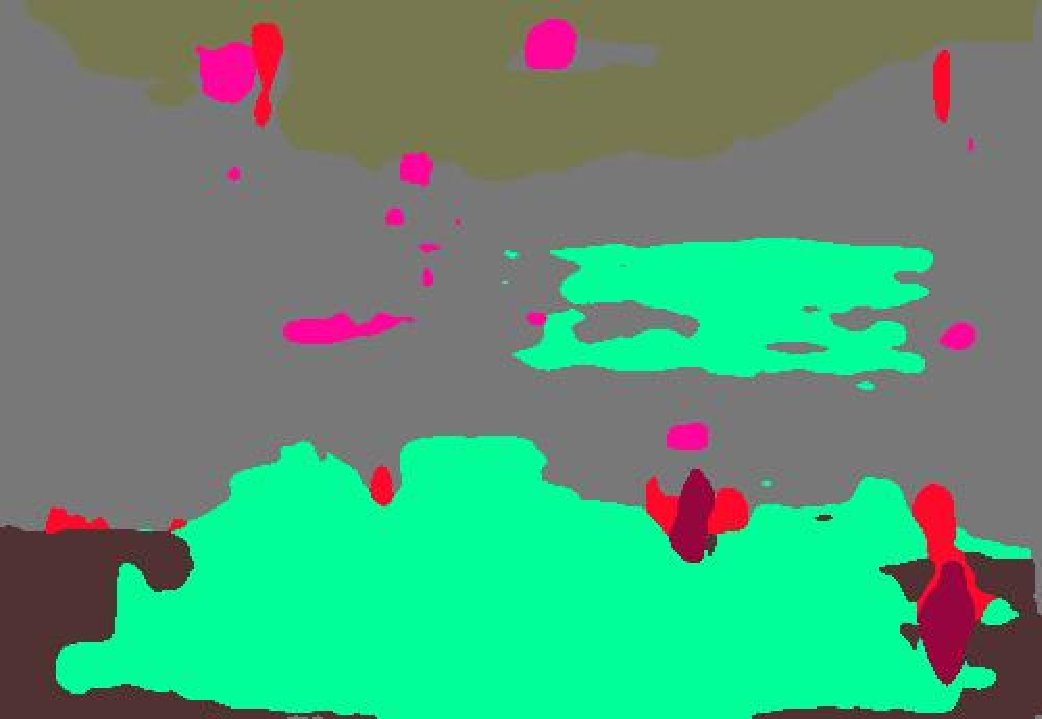}
     \end{subfigure}
     \hspace{-0.4em}
     \begin{subfigure}[b]{0.157\linewidth}
         \centering
         \includegraphics[width=\linewidth]{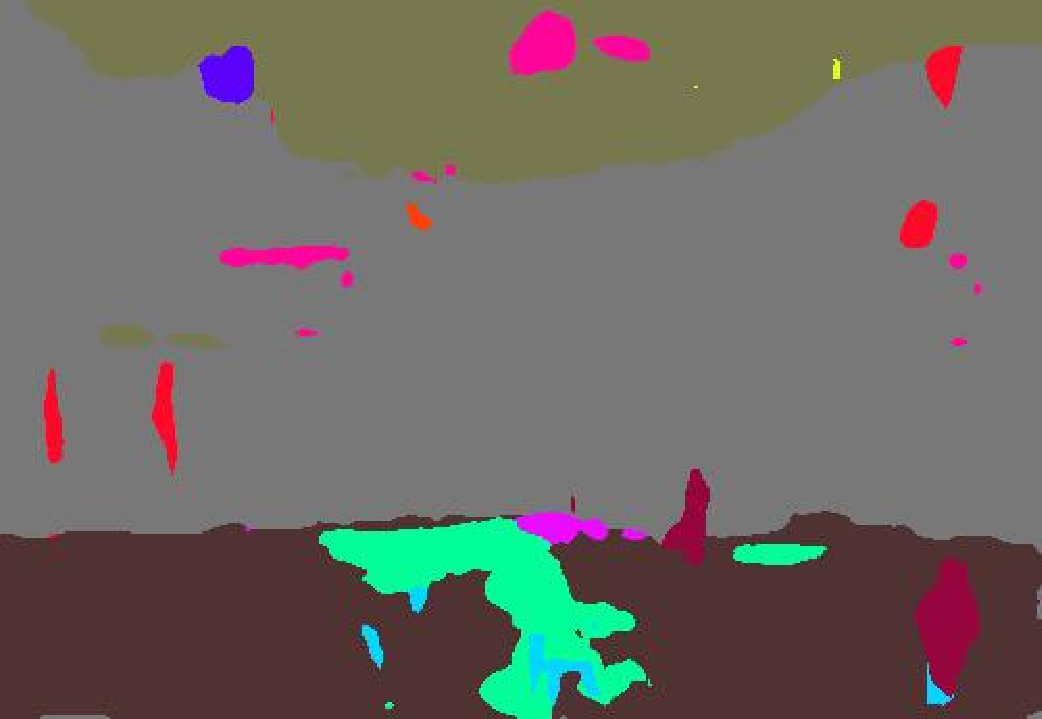}
     \end{subfigure}
     \hfill
     \begin{subfigure}[b]{0.157\linewidth}
         \centering
         \includegraphics[width=\linewidth]{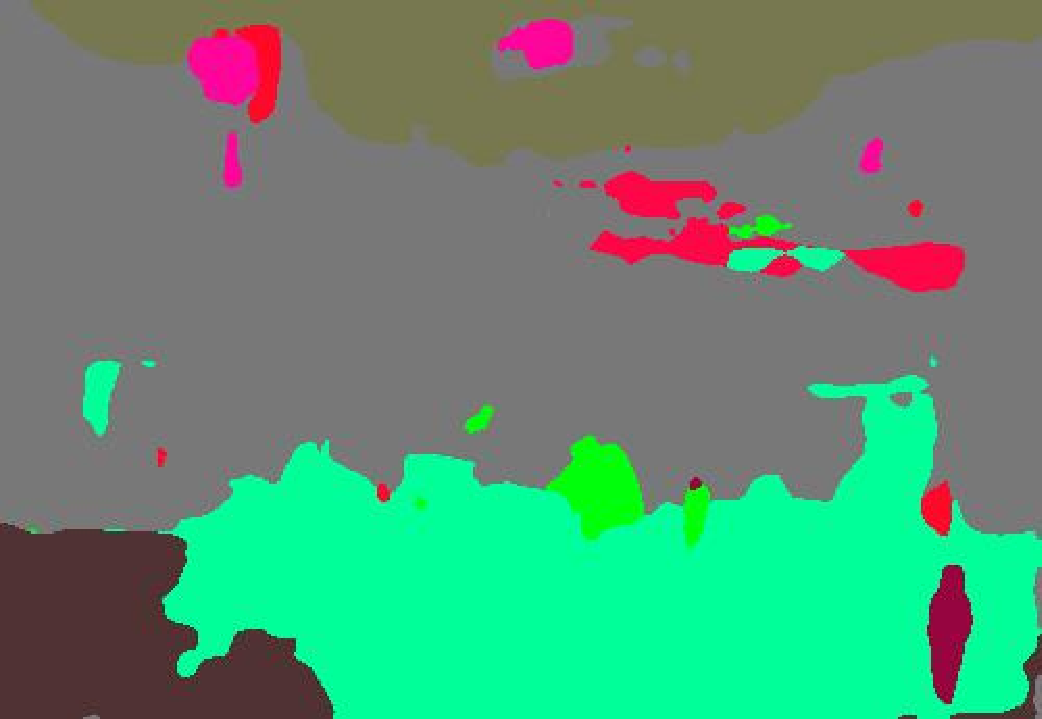}
     \end{subfigure}
     \hspace{-0.4em}
     \begin{subfigure}[b]{0.157\linewidth}
         \centering
         \includegraphics[width=\linewidth]{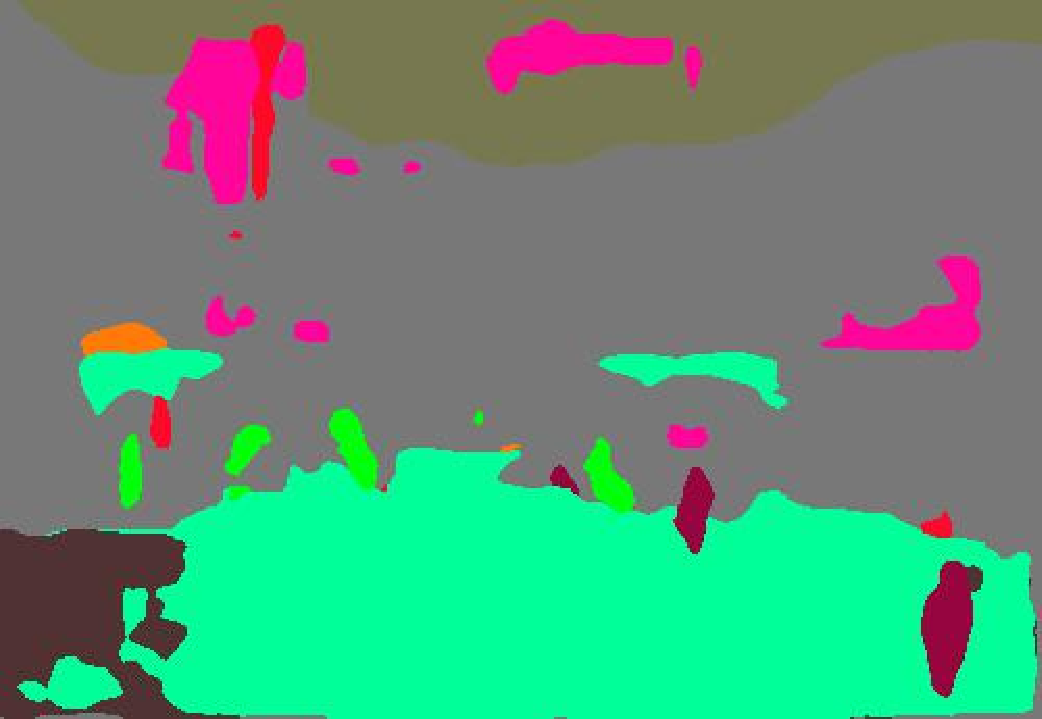}
     \end{subfigure}
     
        \caption{Semantic segmentation results of the first ten images from the validation set of ADE20K. The performance of our algorithms is similar or better than that of the original algorithms.}
        \label{fig:semantic_segmentation}
\end{figure}

We present semantic segmentation results (test image, ground truth, and predicted results of each algorithm) of the first ten images from the validation set of ADE20K. Figure \ref{fig:semantic_segmentation} shows that the representations pre-trained by our algorithms transfer well to the semantic segmentation task. Also, the performance of our algorithms is similar to or better than that of the existing algorithms.

\section{Limitations and future directions}
\label{sec:limitations_and_future_directions}

Since this is the first study in the direction of implicitly applying the idea of contrastive learning, we adopt algorithms with simple losses (such as SimSiam and BYOL) as the benchmark algorithms. To do so, we want to see vividly and not to obscure the effect of our method. The performance may be further improved when applied to algorithms that utilize more complicated strategies. For example, there are algorithms that use a multi-crop strategy \citep{caron2020unsupervised} where both global and local views are considered \citep{caron2021emerging, zhang2022leverage, wanyan2023dino}. Combining our ideas with these algorithms may further improve transfer learning (e.g., dense prediction tasks) performance on multi-object datasets. Since our methodology is an intermediate between contrastive and non-contrastive learning, it may help the community to better understand the differences and similarities between them \citep{garrido2022duality}.

\bibliographystyle{plainnat}
\bibliography{wrapper}

\begin{thebibliography}{11}
\providecommand{\natexlab}[1]{#1}
\providecommand{\url}[1]{\texttt{#1}}
\expandafter\ifx\csname urlstyle\endcsname\relax
  \providecommand{\doi}[1]{doi: #1}\else
  \providecommand{\doi}{doi: \begingroup \urlstyle{rm}\Url}\fi

\bibitem[Caron et~al.(2020)Caron, Misra, Mairal, Goyal, Bojanowski, and Joulin]{caron2020unsupervised}
Mathilde Caron, Ishan Misra, Julien Mairal, Priya Goyal, Piotr Bojanowski, and Armand Joulin.
\newblock Unsupervised learning of visual features by contrasting cluster assignments.
\newblock \emph{Advances in Neural Information Processing Systems}, 33:\penalty0 9912--9924, 2020.

\bibitem[Caron et~al.(2021)Caron, Touvron, Misra, J{\'e}gou, Mairal, Bojanowski, and Joulin]{caron2021emerging}
Mathilde Caron, Hugo Touvron, Ishan Misra, Herv{\'e} J{\'e}gou, Julien Mairal, Piotr Bojanowski, and Armand Joulin.
\newblock Emerging properties in self-supervised vision transformers.
\newblock In \emph{Proceedings of the IEEE/CVF International Conference on Computer Vision}, pages 9650--9660, 2021.

\bibitem[Chen and He(2021)]{chen2021exploring}
Xinlei Chen and Kaiming He.
\newblock Exploring simple siamese representation learning.
\newblock In \emph{Proceedings of the IEEE/CVF Conference on Computer Vision and Pattern Recognition}, pages 15750--15758, 2021.

\bibitem[Ericsson et~al.(2021)Ericsson, Gouk, and Hospedales]{ericsson2021well}
Linus Ericsson, Henry Gouk, and Timothy~M Hospedales.
\newblock How well do self-supervised models transfer?
\newblock In \emph{Proceedings of the IEEE/CVF Conference on Computer Vision and Pattern Recognition}, pages 5414--5423, 2021.

\bibitem[Garrido et~al.(2022)Garrido, Chen, Bardes, Najman, and Lecun]{garrido2022duality}
Quentin Garrido, Yubei Chen, Adrien Bardes, Laurent Najman, and Yann Lecun.
\newblock On the duality between contrastive and non-contrastive self-supervised learning.
\newblock \emph{arXiv preprint arXiv:2206.02574}, 2022.

\bibitem[He et~al.(2016)He, Zhang, Ren, and Sun]{he2016deep}
Kaiming He, Xiangyu Zhang, Shaoqing Ren, and Jian Sun.
\newblock Deep residual learning for image recognition.
\newblock In \emph{Proceedings of the IEEE conference on computer vision and pattern recognition}, pages 770--778, 2016.

\bibitem[Liu and Nocedal(1989)]{liu1989limited}
Dong~C Liu and Jorge Nocedal.
\newblock On the limited memory bfgs method for large scale optimization.
\newblock \emph{Mathematical programming}, 45\penalty0 (1-3):\penalty0 503--528, 1989.

\bibitem[Susmelj et~al.(2020)Susmelj, Heller, Wirth, Prescott, and et~al.]{susmelj2020lightly}
Igor Susmelj, Matthias Heller, Philipp Wirth, Jeremy Prescott, and Malte~Ebner et~al.
\newblock Lightly.
\newblock \emph{GitHub. Note: https://github.com/lightly-ai/lightly}, 2020.

\bibitem[Van~der Maaten and Hinton(2008)]{van2008visualizing}
Laurens Van~der Maaten and Geoffrey Hinton.
\newblock Visualizing data using t-sne.
\newblock \emph{Journal of machine learning research}, 9\penalty0 (11), 2008.

\bibitem[Wanyan et~al.(2023)Wanyan, Seneviratne, Shen, and Kirley]{wanyan2023dino}
Xinye Wanyan, Sachith Seneviratne, Shuchang Shen, and Michael Kirley.
\newblock Dino-mc: Self-supervised contrastive learning for remote sensing imagery with multi-sized local crops.
\newblock \emph{arXiv preprint arXiv:2303.06670}, 2023.

\bibitem[Zhang et~al.(2022)Zhang, Qiu, Ke, S{\"u}sstrunk, and Salzmann]{zhang2022leverage}
Tong Zhang, Congpei Qiu, Wei Ke, Sabine S{\"u}sstrunk, and Mathieu Salzmann.
\newblock Leverage your local and global representations: A new self-supervised learning strategy.
\newblock In \emph{Proceedings of the IEEE/CVF Conference on Computer Vision and Pattern Recognition}, pages 16580--16589, 2022.

\end{thebibliography}


\begin{thebibliography}{65}
\providecommand{\natexlab}[1]{#1}
\providecommand{\url}[1]{\texttt{#1}}
\expandafter\ifx\csname urlstyle\endcsname\relax
  \providecommand{\doi}[1]{doi: #1}\else
  \providecommand{\doi}{doi: \begingroup \urlstyle{rm}\Url}\fi

\bibitem[Asano et~al.(2019)Asano, Rupprecht, and Vedaldi]{asano2019self}
Yuki~Markus Asano, Christian Rupprecht, and Andrea Vedaldi.
\newblock Self-labelling via simultaneous clustering and representation learning.
\newblock \emph{arXiv preprint arXiv:1911.05371}, 2019.

\bibitem[Bachman et~al.(2019)Bachman, Hjelm, and Buchwalter]{bachman2019learning}
Philip Bachman, R~Devon Hjelm, and William Buchwalter.
\newblock Learning representations by maximizing mutual information across views.
\newblock \emph{Advances in neural information processing systems}, 32, 2019.

\bibitem[Bardes et~al.(2021)Bardes, Ponce, and LeCun]{bardes2021vicreg}
Adrien Bardes, Jean Ponce, and Yann LeCun.
\newblock Vicreg: Variance-invariance-covariance regularization for self-supervised learning.
\newblock \emph{arXiv preprint arXiv:2105.04906}, 2021.

\bibitem[Bossard et~al.(2014)Bossard, Guillaumin, and Van~Gool]{bossard2014food}
Lukas Bossard, Matthieu Guillaumin, and Luc Van~Gool.
\newblock Food-101--mining discriminative components with random forests.
\newblock In \emph{Computer Vision--ECCV 2014: 13th European Conference, Zurich, Switzerland, September 6-12, 2014, Proceedings, Part VI 13}, pages 446--461. Springer, 2014.

\bibitem[Caron et~al.(2018)Caron, Bojanowski, Joulin, and Douze]{caron2018deep}
Mathilde Caron, Piotr Bojanowski, Armand Joulin, and Matthijs Douze.
\newblock Deep clustering for unsupervised learning of visual features.
\newblock In \emph{Proceedings of the European conference on computer vision (ECCV)}, pages 132--149, 2018.

\bibitem[Caron et~al.(2019)Caron, Bojanowski, Mairal, and Joulin]{caron2019unsupervised}
Mathilde Caron, Piotr Bojanowski, Julien Mairal, and Armand Joulin.
\newblock Unsupervised pre-training of image features on non-curated data.
\newblock In \emph{Proceedings of the IEEE/CVF International Conference on Computer Vision}, pages 2959--2968, 2019.

\bibitem[Caron et~al.(2020)Caron, Misra, Mairal, Goyal, Bojanowski, and Joulin]{caron2020unsupervised}
Mathilde Caron, Ishan Misra, Julien Mairal, Priya Goyal, Piotr Bojanowski, and Armand Joulin.
\newblock Unsupervised learning of visual features by contrasting cluster assignments.
\newblock \emph{Advances in Neural Information Processing Systems}, 33:\penalty0 9912--9924, 2020.

\bibitem[Caron et~al.(2021)Caron, Touvron, Misra, J{\'e}gou, Mairal, Bojanowski, and Joulin]{caron2021emerging}
Mathilde Caron, Hugo Touvron, Ishan Misra, Herv{\'e} J{\'e}gou, Julien Mairal, Piotr Bojanowski, and Armand Joulin.
\newblock Emerging properties in self-supervised vision transformers.
\newblock In \emph{Proceedings of the IEEE/CVF International Conference on Computer Vision}, pages 9650--9660, 2021.

\bibitem[Chen et~al.(2020{\natexlab{a}})Chen, Kornblith, Norouzi, and Hinton]{chen2020simple}
Ting Chen, Simon Kornblith, Mohammad Norouzi, and Geoffrey Hinton.
\newblock A simple framework for contrastive learning of visual representations.
\newblock In \emph{International conference on machine learning}, pages 1597--1607. PMLR, 2020{\natexlab{a}}.

\bibitem[Chen and He(2021)]{chen2021exploring}
Xinlei Chen and Kaiming He.
\newblock Exploring simple siamese representation learning.
\newblock In \emph{Proceedings of the IEEE/CVF Conference on Computer Vision and Pattern Recognition}, pages 15750--15758, 2021.

\bibitem[Chen et~al.(2020{\natexlab{b}})Chen, Fan, Girshick, and He]{chen2020improved}
Xinlei Chen, Haoqi Fan, Ross Girshick, and Kaiming He.
\newblock Improved baselines with momentum contrastive learning.
\newblock \emph{arXiv preprint arXiv:2003.04297}, 2020{\natexlab{b}}.

\bibitem[Chen et~al.(2021)Chen, Xie, and He]{chen2021empirical}
Xinlei Chen, Saining Xie, and Kaiming He.
\newblock An empirical study of training self-supervised vision transformers.
\newblock In \emph{Proceedings of the IEEE/CVF International Conference on Computer Vision}, pages 9640--9649, 2021.

\bibitem[Chopra et~al.(2005)Chopra, Hadsell, and LeCun]{chopra2005learning}
Sumit Chopra, Raia Hadsell, and Yann LeCun.
\newblock Learning a similarity metric discriminatively, with application to face verification.
\newblock In \emph{2005 IEEE Computer Society Conference on Computer Vision and Pattern Recognition (CVPR'05)}, volume~1, pages 539--546. IEEE, 2005.

\bibitem[Chuang et~al.(2020)Chuang, Robinson, Lin, Torralba, and Jegelka]{chuang2020debiased}
Ching-Yao Chuang, Joshua Robinson, Yen-Chen Lin, Antonio Torralba, and Stefanie Jegelka.
\newblock Debiased contrastive learning.
\newblock \emph{Advances in neural information processing systems}, 33:\penalty0 8765--8775, 2020.

\bibitem[Cimpoi et~al.(2014)Cimpoi, Maji, Kokkinos, Mohamed, and Vedaldi]{cimpoi2014describing}
Mircea Cimpoi, Subhransu Maji, Iasonas Kokkinos, Sammy Mohamed, and Andrea Vedaldi.
\newblock Describing textures in the wild.
\newblock In \emph{Proceedings of the IEEE conference on computer vision and pattern recognition}, pages 3606--3613, 2014.

\bibitem[Deng et~al.(2009)Deng, Dong, Socher, Li, Li, and Fei-Fei]{deng2009imagenet}
Jia Deng, Wei Dong, Richard Socher, Li-Jia Li, Kai Li, and Li~Fei-Fei.
\newblock Imagenet: A large-scale hierarchical image database.
\newblock In \emph{2009 IEEE conference on computer vision and pattern recognition}, pages 248--255. Ieee, 2009.

\bibitem[Doersch et~al.(2015)Doersch, Gupta, and Efros]{doersch2015unsupervised}
Carl Doersch, Abhinav Gupta, and Alexei~A Efros.
\newblock Unsupervised visual representation learning by context prediction.
\newblock In \emph{Proceedings of the IEEE international conference on computer vision}, pages 1422--1430, 2015.

\bibitem[Ericsson et~al.(2021)Ericsson, Gouk, and Hospedales]{ericsson2021well}
Linus Ericsson, Henry Gouk, and Timothy~M Hospedales.
\newblock How well do self-supervised models transfer?
\newblock In \emph{Proceedings of the IEEE/CVF Conference on Computer Vision and Pattern Recognition}, pages 5414--5423, 2021.

\bibitem[Everingham et~al.(2010)Everingham, Van~Gool, Williams, Winn, and Zisserman]{everingham2010pascal}
Mark Everingham, Luc Van~Gool, Christopher~KI Williams, John Winn, and Andrew Zisserman.
\newblock The pascal visual object classes (voc) challenge.
\newblock \emph{International journal of computer vision}, 88\penalty0 (2):\penalty0 303--338, 2010.

\bibitem[Fei-Fei et~al.(2004)Fei-Fei, Fergus, and Perona]{fei2004learning}
Li~Fei-Fei, Rob Fergus, and Pietro Perona.
\newblock Learning generative visual models from few training examples: An incremental bayesian approach tested on 101 object categories.
\newblock In \emph{2004 conference on computer vision and pattern recognition workshop}, pages 178--178. IEEE, 2004.

\bibitem[Frosst et~al.(2019)Frosst, Papernot, and Hinton]{frosst2019analyzing}
Nicholas Frosst, Nicolas Papernot, and Geoffrey Hinton.
\newblock Analyzing and improving representations with the soft nearest neighbor loss.
\newblock In \emph{International conference on machine learning}, pages 2012--2020. PMLR, 2019.

\bibitem[Gidaris et~al.(2018)Gidaris, Singh, and Komodakis]{gidaris2018unsupervised}
Spyros Gidaris, Praveer Singh, and Nikos Komodakis.
\newblock Unsupervised representation learning by predicting image rotations.
\newblock \emph{arXiv preprint arXiv:1803.07728}, 2018.

\bibitem[Grill et~al.(2020)Grill, Strub, Altch{\'e}, Tallec, Richemond, Buchatskaya, Doersch, Avila~Pires, Guo, Gheshlaghi~Azar, et~al.]{grill2020bootstrap}
Jean-Bastien Grill, Florian Strub, Florent Altch{\'e}, Corentin Tallec, Pierre Richemond, Elena Buchatskaya, Carl Doersch, Bernardo Avila~Pires, Zhaohan Guo, Mohammad Gheshlaghi~Azar, et~al.
\newblock Bootstrap your own latent-a new approach to self-supervised learning.
\newblock \emph{Advances in neural information processing systems}, 33:\penalty0 21271--21284, 2020.

\bibitem[Gutmann and Hyv{\"a}rinen(2010)]{gutmann2010noise}
Michael Gutmann and Aapo Hyv{\"a}rinen.
\newblock Noise-contrastive estimation: A new estimation principle for unnormalized statistical models.
\newblock In \emph{Proceedings of the thirteenth international conference on artificial intelligence and statistics}, pages 297--304. JMLR Workshop and Conference Proceedings, 2010.

\bibitem[Hadsell et~al.(2006)Hadsell, Chopra, and LeCun]{hadsell2006dimensionality}
Raia Hadsell, Sumit Chopra, and Yann LeCun.
\newblock Dimensionality reduction by learning an invariant mapping.
\newblock In \emph{2006 IEEE Computer Society Conference on Computer Vision and Pattern Recognition (CVPR'06)}, volume~2, pages 1735--1742. IEEE, 2006.

\bibitem[He et~al.(2016)He, Zhang, Ren, and Sun]{he2016deep}
Kaiming He, Xiangyu Zhang, Shaoqing Ren, and Jian Sun.
\newblock Deep residual learning for image recognition.
\newblock In \emph{Proceedings of the IEEE conference on computer vision and pattern recognition}, pages 770--778, 2016.

\bibitem[He et~al.(2020)He, Fan, Wu, Xie, and Girshick]{he2020momentum}
Kaiming He, Haoqi Fan, Yuxin Wu, Saining Xie, and Ross Girshick.
\newblock Momentum contrast for unsupervised visual representation learning.
\newblock In \emph{Proceedings of the IEEE/CVF conference on computer vision and pattern recognition}, pages 9729--9738, 2020.

\bibitem[Hjelm et~al.(2018)Hjelm, Fedorov, Lavoie-Marchildon, Grewal, Bachman, Trischler, and Bengio]{hjelm2018learning}
R~Devon Hjelm, Alex Fedorov, Samuel Lavoie-Marchildon, Karan Grewal, Phil Bachman, Adam Trischler, and Yoshua Bengio.
\newblock Learning deep representations by mutual information estimation and maximization.
\newblock \emph{arXiv preprint arXiv:1808.06670}, 2018.

\bibitem[Jaiswal et~al.(2020)Jaiswal, Babu, Zadeh, Banerjee, and Makedon]{jaiswal2020survey}
Ashish Jaiswal, Ashwin~Ramesh Babu, Mohammad~Zaki Zadeh, Debapriya Banerjee, and Fillia Makedon.
\newblock A survey on contrastive self-supervised learning.
\newblock \emph{Technologies}, 9\penalty0 (1):\penalty0 2, 2020.

\bibitem[Jing and Tian(2020)]{jing2020self}
Longlong Jing and Yingli Tian.
\newblock Self-supervised visual feature learning with deep neural networks: A survey.
\newblock \emph{IEEE transactions on pattern analysis and machine intelligence}, 43\penalty0 (11):\penalty0 4037--4058, 2020.

\bibitem[Krause et~al.(2013)Krause, Deng, Stark, and Fei-Fei]{krause2013collecting}
Jonathan Krause, Jia Deng, Michael Stark, and Li~Fei-Fei.
\newblock Collecting a large-scale dataset of fine-grained cars.
\newblock 2013.

\bibitem[Krizhevsky et~al.(2009)Krizhevsky, Hinton, et~al.]{krizhevsky2009learning}
Alex Krizhevsky, Geoffrey Hinton, et~al.
\newblock Learning multiple layers of features from tiny images.
\newblock 2009.

\bibitem[Le-Khac et~al.(2020)Le-Khac, Healy, and Smeaton]{le2020contrastive}
Phuc~H Le-Khac, Graham Healy, and Alan~F Smeaton.
\newblock Contrastive representation learning: A framework and review.
\newblock \emph{IEEE Access}, 8:\penalty0 193907--193934, 2020.

\bibitem[Li et~al.(2020)Li, Zhou, Xiong, and Hoi]{li2020prototypical}
Junnan Li, Pan Zhou, Caiming Xiong, and Steven~CH Hoi.
\newblock Prototypical contrastive learning of unsupervised representations.
\newblock \emph{arXiv preprint arXiv:2005.04966}, 2020.

\bibitem[Lin et~al.(2014)Lin, Maire, Belongie, Hays, Perona, Ramanan, Doll{\'a}r, and Zitnick]{lin2014microsoft}
Tsung-Yi Lin, Michael Maire, Serge Belongie, James Hays, Pietro Perona, Deva Ramanan, Piotr Doll{\'a}r, and C~Lawrence Zitnick.
\newblock Microsoft coco: Common objects in context.
\newblock In \emph{Computer Vision--ECCV 2014: 13th European Conference, Zurich, Switzerland, September 6-12, 2014, Proceedings, Part V 13}, pages 740--755. Springer, 2014.

\bibitem[Lin et~al.(2017)Lin, Doll{\'a}r, Girshick, He, Hariharan, and Belongie]{lin2017feature}
Tsung-Yi Lin, Piotr Doll{\'a}r, Ross Girshick, Kaiming He, Bharath Hariharan, and Serge Belongie.
\newblock Feature pyramid networks for object detection.
\newblock In \emph{Proceedings of the IEEE conference on computer vision and pattern recognition}, pages 2117--2125, 2017.

\bibitem[Liu et~al.(2021)Liu, Zhang, Hou, Mian, Wang, Zhang, and Tang]{liu2021self}
Xiao Liu, Fanjin Zhang, Zhenyu Hou, Li~Mian, Zhaoyu Wang, Jing Zhang, and Jie Tang.
\newblock Self-supervised learning: Generative or contrastive.
\newblock \emph{IEEE Transactions on Knowledge and Data Engineering}, 2021.

\bibitem[Loshchilov and Hutter(2016)]{loshchilov2016sgdr}
Ilya Loshchilov and Frank Hutter.
\newblock Sgdr: Stochastic gradient descent with warm restarts.
\newblock \emph{arXiv preprint arXiv:1608.03983}, 2016.

\bibitem[Maji et~al.(2013)Maji, Rahtu, Kannala, Blaschko, and Vedaldi]{maji2013fine}
Subhransu Maji, Esa Rahtu, Juho Kannala, Matthew Blaschko, and Andrea Vedaldi.
\newblock Fine-grained visual classification of aircraft.
\newblock \emph{arXiv preprint arXiv:1306.5151}, 2013.

\bibitem[Misra and Maaten(2020)]{misra2020self}
Ishan Misra and Laurens van~der Maaten.
\newblock Self-supervised learning of pretext-invariant representations.
\newblock In \emph{Proceedings of the IEEE/CVF Conference on Computer Vision and Pattern Recognition}, pages 6707--6717, 2020.

\bibitem[Nilsback and Zisserman(2008)]{nilsback2008automated}
Maria-Elena Nilsback and Andrew Zisserman.
\newblock Automated flower classification over a large number of classes.
\newblock In \emph{2008 Sixth Indian Conference on Computer Vision, Graphics \& Image Processing}, pages 722--729. IEEE, 2008.

\bibitem[Noroozi and Favaro(2016)]{noroozi2016unsupervised}
Mehdi Noroozi and Paolo Favaro.
\newblock Unsupervised learning of visual representations by solving jigsaw puzzles.
\newblock In \emph{European conference on computer vision}, pages 69--84. Springer, 2016.

\bibitem[Oh~Song et~al.(2016)Oh~Song, Xiang, Jegelka, and Savarese]{oh2016deep}
Hyun Oh~Song, Yu~Xiang, Stefanie Jegelka, and Silvio Savarese.
\newblock Deep metric learning via lifted structured feature embedding.
\newblock In \emph{Proceedings of the IEEE conference on computer vision and pattern recognition}, pages 4004--4012, 2016.

\bibitem[Oord et~al.(2018)Oord, Li, and Vinyals]{oord2018representation}
Aaron van~den Oord, Yazhe Li, and Oriol Vinyals.
\newblock Representation learning with contrastive predictive coding.
\newblock \emph{arXiv preprint arXiv:1807.03748}, 2018.

\bibitem[Parkhi et~al.(2012)Parkhi, Vedaldi, Zisserman, and Jawahar]{parkhi2012cats}
Omkar~M Parkhi, Andrea Vedaldi, Andrew Zisserman, and CV~Jawahar.
\newblock Cats and dogs.
\newblock In \emph{2012 IEEE conference on computer vision and pattern recognition}, pages 3498--3505. IEEE, 2012.

\bibitem[Paszke et~al.(2019)Paszke, Gross, Massa, Lerer, Bradbury, Chanan, Killeen, Lin, Gimelshein, Antiga, et~al.]{paszke2019pytorch}
Adam Paszke, Sam Gross, Francisco Massa, Adam Lerer, James Bradbury, Gregory Chanan, Trevor Killeen, Zeming Lin, Natalia Gimelshein, Luca Antiga, et~al.
\newblock Pytorch: An imperative style, high-performance deep learning library.
\newblock \emph{Advances in neural information processing systems}, 32, 2019.

\bibitem[Ren et~al.(2015)Ren, He, Girshick, and Sun]{ren2015faster}
Shaoqing Ren, Kaiming He, Ross Girshick, and Jian Sun.
\newblock Faster r-cnn: Towards real-time object detection with region proposal networks.
\newblock \emph{Advances in neural information processing systems}, 28, 2015.

\bibitem[Schroff et~al.(2015)Schroff, Kalenichenko, and Philbin]{schroff2015facenet}
Florian Schroff, Dmitry Kalenichenko, and James Philbin.
\newblock Facenet: A unified embedding for face recognition and clustering.
\newblock In \emph{Proceedings of the IEEE conference on computer vision and pattern recognition}, pages 815--823, 2015.

\bibitem[Sohn(2016)]{sohn2016improved}
Kihyuk Sohn.
\newblock Improved deep metric learning with multi-class n-pair loss objective.
\newblock \emph{Advances in neural information processing systems}, 29, 2016.

\bibitem[Tarvainen and Valpola(2017)]{tarvainen2017mean}
Antti Tarvainen and Harri Valpola.
\newblock Mean teachers are better role models: Weight-averaged consistency targets improve semi-supervised deep learning results.
\newblock \emph{Advances in neural information processing systems}, 30, 2017.

\bibitem[Tian et~al.(2020)Tian, Krishnan, and Isola]{tian2020contrastive}
Yonglong Tian, Dilip Krishnan, and Phillip Isola.
\newblock Contrastive multiview coding.
\newblock In \emph{European conference on computer vision}, pages 776--794. Springer, 2020.

\bibitem[Wang and Isola(2020)]{wang2020understanding}
Tongzhou Wang and Phillip Isola.
\newblock Understanding contrastive representation learning through alignment and uniformity on the hypersphere.
\newblock In \emph{International Conference on Machine Learning}, pages 9929--9939. PMLR, 2020.

\bibitem[Wang et~al.(2022)Wang, Fan, Tian, Kihara, and Chen]{wang2022importance}
Xiao Wang, Haoqi Fan, Yuandong Tian, Daisuke Kihara, and Xinlei Chen.
\newblock On the importance of asymmetry for siamese representation learning.
\newblock In \emph{Proceedings of the IEEE/CVF Conference on Computer Vision and Pattern Recognition}, pages 16570--16579, 2022.

\bibitem[Wu et~al.(2019)Wu, Kirillov, Massa, Lo, and Girshick]{wu2019detectron2}
Yuxin Wu, Alexander Kirillov, Francisco Massa, Wan-Yen Lo, and Ross Girshick.
\newblock Detectron2.
\newblock \url{https://github.com/facebookresearch/detectron2}, 2019.

\bibitem[Wu et~al.(2018)Wu, Xiong, Yu, and Lin]{wu2018unsupervised}
Zhirong Wu, Yuanjun Xiong, Stella~X Yu, and Dahua Lin.
\newblock Unsupervised feature learning via non-parametric instance discrimination.
\newblock In \emph{Proceedings of the IEEE conference on computer vision and pattern recognition}, pages 3733--3742, 2018.

\bibitem[Xiao et~al.(2010)Xiao, Hays, Ehinger, Oliva, and Torralba]{xiao2010sun}
Jianxiong Xiao, James Hays, Krista~A Ehinger, Aude Oliva, and Antonio Torralba.
\newblock Sun database: Large-scale scene recognition from abbey to zoo.
\newblock In \emph{2010 IEEE computer society conference on computer vision and pattern recognition}, pages 3485--3492. IEEE, 2010.

\bibitem[Xiao et~al.(2018)Xiao, Liu, Zhou, Jiang, and Sun]{xiao2018unified}
Tete Xiao, Yingcheng Liu, Bolei Zhou, Yuning Jiang, and Jian Sun.
\newblock Unified perceptual parsing for scene understanding.
\newblock In \emph{Proceedings of the European conference on computer vision (ECCV)}, pages 418--434, 2018.

\bibitem[Ye et~al.(2019)Ye, Zhang, Yuen, and Chang]{ye2019unsupervised}
Mang Ye, Xu~Zhang, Pong~C Yuen, and Shih-Fu Chang.
\newblock Unsupervised embedding learning via invariant and spreading instance feature.
\newblock In \emph{Proceedings of the IEEE/CVF Conference on Computer Vision and Pattern Recognition}, pages 6210--6219, 2019.

\bibitem[You et~al.(2017)You, Gitman, and Ginsburg]{you2017large}
Yang You, Igor Gitman, and Boris Ginsburg.
\newblock Large batch training of convolutional networks.
\newblock \emph{arXiv preprint arXiv:1708.03888}, 2017.

\bibitem[Zbontar et~al.(2021)Zbontar, Jing, Misra, LeCun, and Deny]{zbontar2021barlow}
Jure Zbontar, Li~Jing, Ishan Misra, Yann LeCun, and St{\'e}phane Deny.
\newblock Barlow twins: Self-supervised learning via redundancy reduction.
\newblock In \emph{International Conference on Machine Learning}, pages 12310--12320. PMLR, 2021.

\bibitem[Zhang et~al.(2016)Zhang, Isola, and Efros]{zhang2016colorful}
Richard Zhang, Phillip Isola, and Alexei~A Efros.
\newblock Colorful image colorization.
\newblock In \emph{European conference on computer vision}, pages 649--666. Springer, 2016.

\bibitem[Zhao et~al.(2017)Zhao, Shi, Qi, Wang, and Jia]{zhao2017pyramid}
Hengshuang Zhao, Jianping Shi, Xiaojuan Qi, Xiaogang Wang, and Jiaya Jia.
\newblock Pyramid scene parsing network.
\newblock In \emph{Proceedings of the IEEE conference on computer vision and pattern recognition}, pages 2881--2890, 2017.

\bibitem[Zhou et~al.(2017)Zhou, Zhao, Puig, Fidler, Barriuso, and Torralba]{zhou2017scene}
Bolei Zhou, Hang Zhao, Xavier Puig, Sanja Fidler, Adela Barriuso, and Antonio Torralba.
\newblock Scene parsing through ade20k dataset.
\newblock In \emph{Proceedings of the IEEE Conference on Computer Vision and Pattern Recognition}, 2017.

\bibitem[Zhou et~al.(2018)Zhou, Zhao, Puig, Xiao, Fidler, Barriuso, and Torralba]{zhou2018semantic}
Bolei Zhou, Hang Zhao, Xavier Puig, Tete Xiao, Sanja Fidler, Adela Barriuso, and Antonio Torralba.
\newblock Semantic understanding of scenes through the ade20k dataset.
\newblock \emph{International Journal on Computer Vision}, 2018.

\bibitem[Zhou et~al.(2019)Zhou, Zhao, Puig, Xiao, Fidler, Barriuso, and Torralba]{zhou2019semantic}
Bolei Zhou, Hang Zhao, Xavier Puig, Tete Xiao, Sanja Fidler, Adela Barriuso, and Antonio Torralba.
\newblock Semantic understanding of scenes through the ade20k dataset.
\newblock \emph{International Journal of Computer Vision}, 127:\penalty0 302--321, 2019.

\end{thebibliography}

\end{document}